%% file: main.tex
\documentclass[10pt,twocolumn,letterpaper]{article}

\usepackage[pagenumbers]{cvpr} 

\usepackage[utf8]{inputenc}
\usepackage[dvipsnames]{xcolor}
\definecolor{cvprblue}{rgb}{0.21,0.49,0.74}
\usepackage{hyperref}
\hypersetup{pagebackref,breaklinks,colorlinks,citecolor=cvprblue}
\usepackage{multicol}
\usepackage{graphicx}
\usepackage{tabularx}
\usepackage{tikz}
\usepackage{color}
\usetikzlibrary{spy}
\usepackage{svg}
\usepackage[margin=1in]{geometry} 
\usepackage{multirow}
\usepackage{array}
\def\paperID{6000} 
\def\confName{CVPR}
\def\confYear{2024}

\newcommand{\bs}{\mathbf{s}}
\newcommand{\ba}{\mathbf{a}}
\newcommand{\bz}{\mathbf{z}}
\newcommand{\bx}{\mathbf{x}}
\newcommand{\bc}{\mathbf{c}}

\newcommand{\z}[0]{\mathbf{z}}
\newcommand{\zt}[0]{{\z}_t}

\newcommand{\x}[0]{\mathbf{x}}
\newcommand{\expect}[2]
{\mathbb{E}_{#1} \left[ #2 \right] }
\DeclareMathOperator*{\argmax}{arg\,max}

\newcommand\scalemath[2]{\scalebox{#1}{\mbox{\ensuremath{\displaystyle #2}}}}

\title{RL Dreams: Policy Gradient Optimization for Score Distillation based 3D Generation}

\author{Aradhya N. Mathur \thanks{denotes equal contribution by the authors}\\
IIITD\\
{\tt\small aradhyam@iiitd.ac.in}
\and
Phu Pham \footnotemark[1]\\
Purdue University\\
{\tt\small pham84@purdue.edu}
\and
Aniket Bera\\
Purdue University\\
{\tt\small aniketbera@purdue.edu}
\and
Ojaswa Sharma\\
IIITD\\
{\tt\small ojaswa@iiitd.ac.in}
}

\begin{document}
\maketitle
\input{sec/0_abstract}    
\input{sec/1_intro}
\input{sec/2_related_work}
\input{sec/3_2_approach}
\input{sec/results_1}
\input{sec/conclusion}
{
    \small
    \bibliographystyle{ieeenat_fullname}
    \bibliography{main}
}

\input{sec/X_suppl}

\end{document}

%% file: sec/0_abstract.tex
\begin{abstract}

3D generation has rapidly accelerated in the past decade owing to the progress in the field of generative modeling. Score Distillation Sampling (SDS) based rendering has improved 3D asset generation to a great extent. Further, the recent work of Denoising Diffusion Policy Optimization (DDPO)  demonstrates that the diffusion process is compatible with policy gradient methods and has been demonstrated to improve the 2D diffusion models using an aesthetic scoring function. We first show that this aesthetic scorer acts as a strong guide for a variety of SDS-based methods and demonstrates its effectiveness in text-to-3D synthesis.
Further, we leverage the DDPO approach to improve the quality of the 3D rendering obtained from 2D diffusion models. Our approach, DDPO3D, employs the policy gradient method in tandem with aesthetic scoring. 
To the best of our knowledge, this is the first method that extends policy gradient methods to 3D score-based rendering and shows improvement across SDS-based methods such as DreamGaussian, which are currently driving research in text-to-3D synthesis. Our approach is compatible with score distillation-based methods, which would facilitate the integration of diverse reward functions into the generative process. Our project page can be accessed via \href{https://ddpo3d.github.io}{https://ddpo3d.github.io}.

\end{abstract}

%% file: sec/1_intro.tex
\section{Introduction}
\label{sec:intro}

The intersection of computer vision and graphics has experienced significant advancements in recent years, unveiling novel possibilities for advancing our understanding of visual data \cite{mildenhall2020nerf, yu2021pixelnerf, barron2021mip, barron2023zip}. The combination of these two disciplines has given rise to innovative approaches that transcend traditional boundaries, unlocking new domains of exploration. One such frontier lies in bridging the gap between vision and language, a pursuit that holds the potential to transform how we interpret and engage with the visual world.

Recently, the advent of diffusion models \cite{Imagen, StableDiffusion, Dalle3} has played a pivotal role in extending the capabilities of this interdisciplinary field, enabling the seamless translation of textual descriptions into 2D representations. However, as we dive into the realm of three-dimensional (3D) reconstructions, a notable challenge emerges. Despite the development made in 2D translation, the transition from text to 3D remains a formidable task, primarily attributed to the scarcity of comprehensive 3D datasets. This scarcity not only hinders the training of robust models but also underscores the pressing need for innovative methodologies to overcome the inherent challenges in this dimensionally complex domain.

The capability of diffusion models to approximate distributions has accelerated the synthetic data generation process, thus becoming the mainstream generative modeling technique. Diffusion models have demonstrated effective and high-quality generation of image, audio, video, and several other modalities. 
Beyond their initial training objectives, these models have surpassed their limitations and evolved into fundamental frameworks, extending their utility to a range of downstream tasks, including but not limited to image composition, in-painting, and image editing, as demonstrated in notable works like \cite{RePaint, SmartBrush, Imagic, NullInversion}.

Recently, pre-trained large diffusion models have been utilized for 3D shape generation techniques due to the advantages that they offer in terms of the vast knowledge base that they serve as, which can be distilled into neural radiance fields (NeRFs) for rendering \cite{poole2022dreamfusion, raj2023dreambooth3d, wang2023score}. Recent techniques such as score distillation sampling and variational score distillation have demonstrated how effectively these models can be leveraged to generate implicit 3D representations called NeRFs, allowing for fast, direct, high-quality 3D renderings from text prompt input \cite{poole2022dreamfusion, wang2023score, wang2023prolificdreamer} or single image \cite{liu2023zero, melas2023realfusion} and now even multi-image based generation \cite{raj2023dreambooth3d}.

Diffusion models, a recent entrant in generative modeling, deviate from the conventional minimax game theoretic approach employed by GANs \cite{goodfellow2020generative} and utilize log-likelihood objectives in a multi-step optimization process. Building upon this paradigm, DDPO \cite{black2023training} adopts a reinforcement learning framework to treat the diffusion process as a multi-step decision-making challenge, enhancing the image generation capabilities of pre-trained 2D diffusion models. In alignment with these advancements, our research endeavors to enhance the visual quality of 3D rendering techniques, specifically focusing on aesthetic improvements. We propose the integration of a DDPO-based policy gradient approach into the score distillation sampling process, aiming for better results in 3D rendering. This approach not only facilitates the refinement of aesthetic aspects but also allows for the incorporation of non-differentiable rewards, enabling diverse optimization strategies. The main contributions of our work can be summarized below.

\begin{enumerate}
    
    \item We introduce DDPO3D, an adaptable framework designed for seamless integration with any SDS-based method and its derivatives in the field of 3D rendering.
    \item We showcase the efficacy of DDPO3D through both qualitative and quantitative assessments, showcasing improvements in  CLIP scores. These enhancements are observed when our framework is integrated with existing SDS-based rendering methods, including DreamGaussian \cite{tang2023dreamgaussian}, GsGen \cite{chen2023text}, and Dreamfusion \cite{poole2022dreamfusion}.
    \item Leveraging the foundation of DDPO, our approach ensures the flexibility to incorporate non-differentiable reward functions into SDS-based methods, thereby extending the applicability of our framework to a broader range of scenarios in 3D rendering.
\end{enumerate}

%% file: sec/2_related_work.tex
\section{Related Work}
\label{sec:formatting}
Our work primarily relates to the 2D diffusion models and the subsequent derivative methods for generating 3D shapes.

\subsection{NeRFs}
One of the seminal works by Mildenhall et al. \cite{mildenhall2020nerf} introduced novel view synthesis by using shallow MLPs and allowing them to overfit on a set of views conditioned on the respective camera poses. The input to such a network is a 5D point (position $(x,y,z)$ and viewing direction $(\theta, \phi)$ ), and the output is an implicit field consisting of RGB color and density $\sigma$ at any point. For rendering a view after the optimization, a volume raymarching process evaluates the MLP at various ray positions to compose an image. Several variants have evolved focusing on anti-aliasing, unbounded scenes \cite{barron2021mip, barron2023zip}, reflections \cite{Guo_2022_CVPR}, and stylization \cite{huang2022stylizednerf}. NeRFs have been further extended to representations other than MLPs as well, such as voxel-based representations \cite{Kondo2021VaxNeRFRT, yu_and_fridovichkeil2021plenoxels}, tensor-based representations such as Hexplane \cite{cao2023hexplane}, K-planes \cite{fridovich2023k}, and TensorRF \cite{Chen2022ECCV}. Recently NeRFs have been used for generative modeling with diffusion-based rendering as further discussed.

\subsubsection{Gaussian splatting}
Kerbl et al. \cite{kerbl3Dgaussians} present a technique to enable fast high-resolution novel view synthesis using Gaussian Splatting, which has emerged as a faster alternative for NeRF-based approaches. The technique first generates a sparse point cloud from provided multi-view images using Structure-from-Motion (SFM) and uses that for initializing 3D Gaussians that are differentiable and allow for easy projection to 2D splats, resulting in fast $\alpha$-blending for real-time performance.
These methods have also been extended beyond 3D owing to their fast speed and have been used for learning dynamic scene representation. Wu et al. \cite{wu20234d} explore Gaussian splatting for dynamic scene rendering. They propose a pipeline in which they use the centers of the Gaussians and use the time step to query the multi-resolution voxel planes. Luiten et al. \cite{luiten2023dynamic} have further performed 6-DOF tracking along with view synthesis. They allow for Gaussians to be transformed under persistent color, opacity, and size.    
While these methods learn from a set of images and generate novel views, there have been several works in the realm of generative modeling that have also been leveraged for the same.
\subsection{Diffusion Models}
Taking inspiration from non-equilibrium statistical physics, Sohl-Dickstein et al. \cite{sohl2015deep} introduced a novel deep unsupervised learning approach aimed at capturing data distributions. Their method involves an initial disruption of the inherent structure in the data distribution through an iterative forward diffusion process. Subsequently, a deep learning model is trained to reconstruct the original distribution, thus facilitating the acquisition of generative modeling skills throughout the learning process. This methodology has been expanded to various modalities, encompassing images \cite{dhariwal2021diffusion}, videos \cite{ruan2023mm, ho2022imagen}, audio \cite{kong2020diffwave, mittal2021symbolic, schneider2023archisound}, among others. Dhariwal and Nichol \cite{dhariwal2021diffusion} showcased the efficacy of diffusion models, demonstrating significant enhancements in image generation. They achieved superior FID scores on ImageNet $512\times512$ by employing a modified UNet architecture as a generator and introducing \emph{classifier guidance}, allowing the diffusion model to be conditioned on classifier gradients. The extension of this approach to high-resolution image generation without auxiliary classifier assistance was further advanced by Ho et al. \cite{ho2022cascaded}. In the medical domain, Kazerouni et al. \cite{kazerouni2022diffusion} applied diffusion models, with subsequent extensions to conditional generation using text prompts. Moreover, these techniques have found application in image super-resolution \cite{gao2023implicit, li2022srdiff}, inpainting \cite{Rombach_2022_CVPR}, restoration \cite{liu2023improved}, image translation \cite{xia2023diffi2i}, and editing \cite{ruiz2023dreambooth}.
    
\subsubsection{Image Generation from text}
Diffusion models have played a pivotal role in advancing the field of generative AI, contributing to significant milestones. Notably, the seminal work of Rombach et al. \cite{Rombach_2022_CVPR} has revolutionized high-resolution text-to-image synthesis by employing diffusion within the latent space. Their innovative approach involves a two-step process, wherein an autoencoder is initially trained on a large dataset, followed by training of a diffusion model on the latent vectors obtained from the encoder.

The evolution of prompt-based image generation owes much to diffusion models. Ramesh et al. \cite{ramesh2022hierarchical} showcase the integration of contrastive models like CLIP \cite{radford2021learning} in text-conditioned image generation. The proposed two-stage model generates an image embedding from CLIP based on a text caption, followed by a decoder generating an image from the embedding. The shared embedding space of CLIP facilitates language-guided image manipulation. Experimental results demonstrate the superior performance of diffusion models compared to autoregressive models, particularly on DALL-E datasets \cite{pmlr-v139-ramesh21a}. Additionally, Gu et al. \cite{gu2022vector} showcase high-quality image generation using Vector Quantised-Variational AutoEncoder (VQVAE), offering an improved trade-off between quality and speed. Kang et al. \cite{kang2023gigagan} scale up GANs for extensive image datasets like LAION \cite{schuhmann2022laion}, delivering advantages such as high-speed generation, interpretable latent spaces, and high-resolution $4k$ generation.
    


\subsubsection{Diffusion+NeRF Models for 3D Rendering}
Numerous diffusion models have undergone training on extensive image datasets, such as LAION \cite{schuhmann2022laion}, containing over a billion meticulously curated training samples sourced from the web. These prompt-based models exhibit remarkable generalization capabilities. Leveraging the strengths of prompt-based diffusion models, researchers employ them to train Neural Radiance Fields (NeRF). Poole et al. \cite{poole2022dreamfusion} introduced Score Distillation Sampling that involves utilizing a pre-trained diffusion model as a frozen critic to guide the NeRF in image generation. To incorporate pose information, prompts are enriched with pose-centric phrases and keywords, enhancing the NeRF model's ability to distill accurate pose-dependent details.

Addressing limitations like blurring and the Janus problem, Wang et al. \cite{wang2023score} acknowledge the distribution mismatch issue. They propose a Perturb and Average Scoring (PAAS) mechanism to mitigate this problem, computing scores on noisy images by perturbing them with noise and averaging the scores obtained from each perturbation.

Instead of relying on Score Distillation Sampling (SDS), Liu et al. \cite{liu2023zero} advocate fine-tuning a Stable Diffusion model on an input image and target camera pose. Training on the Objaverse dataset \cite{deitke2023objaverse}, they engage in single-image conditioned multi-view generation. Methods like DiffRF \cite{muller2023diffrf} explore the use of 3D UNets to learn diffusion-based NeRF-like rendering. Render Diffusion \cite{anciukevivcius2023renderdiffusion} endeavors to conduct 3D reconstruction, inpainting, and generation solely with monocular 2D supervision. The method employs diffusion to learn to render novel views conditioned on camera views directly. Similarly, Daniel et al. \cite{watson2022novel} introduce a simple diffusion-based novel view rendering method for directly generating images corresponding to novel viewpoints. Extensive results on the ShapeNet NMR dataset \cite{chang2015shapenet} validate their approach. Several extensions of Gaussian Splatting to SDS-based differentiable rendering \cite{chen2023text}, \cite{tang2023dreamgaussian} accelerate text-to-3D and image-to-3D processes significantly.

Our approach provides an extension for DDPO to be used in conjunction with the aforementioned diffusion-based 3D generation, demonstrating higher-quality results.

%% file: sec/3_2_approach.tex
\begin{figure*}
    \centering
    \includegraphics[width=.95\textwidth]{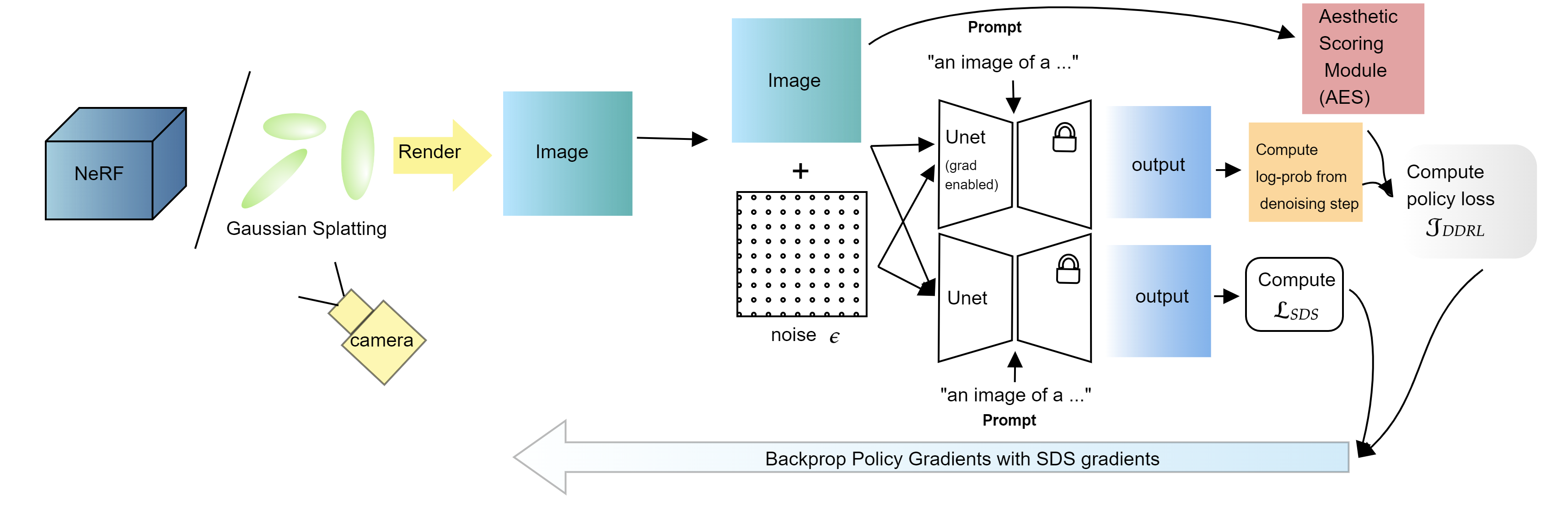}
    \caption{We demonstrate our approach in the above figure. We start with the SDS-based gradient calculation using a frozen UNet model calculated by performing the denoising operation on the image generated using the NeRF/Gaussian Splatting technique. Then, using the rendered image, we calculate the aesthetic score and perform another step of noising-denoising using the UNet as a policy network. We use the generated latents as the action and compute the policy gradient, which is then used to update the NeRF/Gaussian splat parameters to maximize the aesthetic score. }
    \label{fig:enter-label}
\end{figure*}

\section{Approach}
We observe that the aesthetic scoring function is vital in improving the quality of the diffusion models trained using DDPO. We, therefore, first experiment with the impact of the aesthetic scoring function referred to as AES that takes in an input image $x$ and provides an aesthetic score, $AES(x):\mathbb{R}^{c \times h \times w } \mapsto \mathbb{R}$. This is first discussed by Schuhmann et al.~\cite{schuhmann2022laion}, who implemented the scorer as a shallow MLP that was trained on annotated data for aesthetic score prediction using the CLIP embeddings of the images and later used in DDPO as a means to improve the diffusion model. 
We incorporate this term with diffusion loss to improve the aesthetic quality of the image rendered via the SDS method. Thus, the $AES(\cdot)$ model acts as a guide for the aesthetic quality and allows for a better generation quality. We optimize the typical NeRF parameters $\Theta$ in an SDS setup using the following objective
\vspace*{-5pt}
\begin{equation}
    \Theta* = \argmax_\Theta AES(x) ; x \sim p(x).
\end{equation}
\vspace*{-5pt}
Thus, the complete training loss for a rendered sample $x$ is, 
\begin{equation}
    \nonumber
    \mathcal{L}_{total}  = \mathcal{L}_{SDS} - \lambda AES(x).
\end{equation}
where $\mathcal{L}_{SDS}$ is the SDS loss term, which we will further explain. 
However, this also requires investigation as to how to incorporate other rewards in the SDS generation. We derive inspiration from DDPO and realize that the same property of viewing the diffusion process from a reinforcement learning standpoint could offer advantages even for 3D generation. The seminal work by Black et al. \cite{black2023training} demonstrates effective ways for fine-tuning and training diffusion models. They view the diffusion process as an MDP and treat noise prediction by the network as the action taken by a policy network.
\par Thus, we aim to leverage the reinforcement learning framework for tuning the Score Distillation-based approaches for additional reward functions that would yield higher-quality results. We use the method of Black et al. \cite{black2023training} for policy gradient estimation to get the Monte Carlo estimate of our policy loss $\mathcal{J}_\text{DDRL}$ as $\nabla_\theta\mathcal{J}_\text{DDRL}$ which we refer in our case as $\mathcal{J}_\text{DDRL3D}$ and its gradient $\nabla_\theta\mathcal{J}_\text{DDRL3D}$. 

The pre-trained diffusion model takes the context $c \sim p(c)$ 
and induces a sample distribution $p_\theta(x_0|c)$, from the underlying context distribution for the training data $x_0 \sim p_\theta(x_0|c)$. This denoising diffusion RL (DDRL) objective is used to maximize the reward signal $r$ defined on the samples and the context $c$ for the MDP
\vspace*{-5pt}
\begin{equation*}
    \mathcal{J}_\text{DDRL3D}(\theta) = \; \expect{
        \bc \sim p(\bc),~\bx_0 \sim p_\theta(\bx_0 \mid \bc)
    }{
        r(\bx_0, \bc).
    }
\end{equation*}
 The denoising diffusion process is considered as an MDP whereby the state is defined as a tuple of the context $c$, time $t$ i.e. $\bs_t  \triangleq (\bc, t, \bx_t)$ and the latent $x_t$ which are taken by the diffusion model which is represented as policy $\pi$ and defined as $\pi(\ba_t \mid \bs_t) \triangleq p_\theta(\bx_{t-1} \mid \bx_t, \bc)$ and performs the action $ \ba_t  \triangleq \bx_{t-1}$ for performing the noising operation. The state transition is given by $P$ where  $P(\bs_{t+1} \mid \bs_t, \ba_t)  \triangleq \big( \delta_\bc, \delta_{t-1}, \delta_{\bx_{t-1}} \big)$ where $\delta_y$ represents the Dirac delta distribution with non-zero density solely at $y$.
  The reward $ R(\bs_t, \ba_t)$ is considered as $0$ for all timesteps except at the final denoising step $t=0$ where it is given by $ r(\bx_0, \bc)$.
Black and colleagues \cite{black2023training} employ a training regimen based on REINFORCE \citep{williams1992simple, mohamed2020monte} for fine-tuning, utilizing the gradients derived from:
\begin{align}
    \nabla_{\theta} \mathcal{J}_\text{DDRL}&=\expect{}{\; \sum_{t=0}^{T}\nabla_\theta \log p_\theta(\bx_{t-1}\mid \bx_t,\bc)\;r(\bx_0,\bc)}. \label{eq:score_pg}\end{align}
We use the same REINFORCE-based training regimen as given in equation \ref{eq:score_pg}
and extend this policy gradient update to the generated image. 
\par
We consider a setting similar to DreamFusion \cite{poole2022dreamfusion} that proposes 3D generation by considering differentiable image parameterization (DIP) where the generator $g$, which is a DIP parameterized by $\Theta$ produces an image $\x = g(\Theta)$ which is then optimized using the gradients calculated by the denoising process from the latents $z_t$ obtained from the image $\x$ rendered at a timestep $t$ conditioned on the prompt $y$
\begin{align}
\scalemath{0.95}{
\nabla_{\Theta}\mathcal{L}_{\text{SDS}}(\phi, \x=g(\Theta))\triangleq \mathbb{E}_{t, \epsilon}\left[w(t)\left(\hat\epsilon_\phi(\zt; y, t)  - \epsilon\right){\partial \mathbf{x} \over \partial \Theta}\right]\label{eq:sdsgrad}}
\end{align}
where $\hat\epsilon_\phi$ is the noise predicted from a network parameterized by $\phi$; $\epsilon$ is the target noise and $w(t)$ is the time-dependent weight.
This approach, termed Score Distillation Sampling (SDS), involves employing the denoising network, UNet, from the diffusion model as a critic for the generator. To accomplish this, we compute the gradients with respect to the generator parameters $\Theta$, acquiring the gradient term from the latents $\bz$ derived from the rendered image $\bx$ as in eq. \ref{eq:ddrl}.
\vspace{-3pt}
\begin{multline}
    \nabla_\Theta \mathcal{J}_\text{DDRL3D}(\phi, \x_{t-1}=g(\Theta)) \\
    = \expect{}{\; \sum_{t=0}^{T} \nabla_\Theta \log p_\theta(\bz_{t-1} \mid \bz_t, \bc) \; r(\bz_{t-1},\bc)}. \label{eq:ddrl}
\end{multline}

This gradient term is used to drive the generator $g$ according to the additional reward functions used in conjunction with the generation.
Instead of fine-tuning the Stable Diffusion backbone, we propagate the gradients back to the neural renderer to update the parameters w.r.t. the additional rewards. We refer to this method as \emph{DDPO3D} since it allows for the differentiable image parameterization to be updated using the additional reward-based updates. This can be intuitively thought of as state representation learning from a fixed policy since we do not update the Stable Diffusion model parameters. It can be explained as an attempt to find a state where the policy works successfully. This allows integration with Gaussian Splatting-based generation methods that, at each step, sample a random camera pose $p$ around the object center and render the RGB image $I^p_\text{RGB}$ and transparency $I^p_\text{A}$ of the current view.
For the Gaussian splatting, the gradients are computed as
\begin{align}
    \nabla_{\Theta} \mathcal{L}_\text{SDS(splatting)} = \mathbb{E}_{t, p, \mathbf{\epsilon}}\left[(\epsilon_\phi(I^p_\text{RGB}; t, e) - \epsilon) \frac {\partial I^p_\text{RGB}} {\partial {\Theta}}\right],\label{eq:sdssplat}
\end{align}
where $e$ is the CLIP embedding of the input text description. Thus helping us find a sweet spot between speed and quality of generation. We combine the DDPO with the above gradient updates to yield the following update equations

\begin{itemize}
\item For SDS (using eq. \ref{eq:sdsgrad} and \ref{eq:ddrl})
\begin{align}
    \nabla_\Theta \mathcal{L}_{\text{total(SDS)}} = \nabla_\Theta \mathcal{L}_{\text{SDS}} + \nabla_\Theta \mathcal{L}_{\text{DDRL3D}}, \text{and}
\end{align}
\item For splatting (using eq. \ref{eq:sdssplat} and \ref{eq:ddrl})
\begin{align}
    \nabla_\Theta \mathcal{L}_{\text{total(splatting)}} = \nabla_\Theta \mathcal{L}_{\text{SDS(splatting)}} + \nabla_\Theta \mathcal{L}_{\text{DDRL3D}}.
\end{align}
\end{itemize}

Our approach is compatible with both DreamFusion-based NeRF rendering and DreamGaussian-based Gaussian splatting, offering significant flexibility and enhancing the overall generation process.
Since these methods rely on Stable Diffusion-based generation, we can integrate our proposed method relatively easily. However, since most of these methods rely on single-step updates during optimization, we currently restrict our reward computation to one-step MDP, that is, the stage at which we start the noising process. We take the rendered image during the generation, perform the noising step in score distillation, and then do the denoising using the Stable Diffusion UNet. Through this, we compute the log probability similar to DDPO corresponding to the denoising step. However, instead of updating the parameters of UNet, we backpropagate the gradients back to the NeRF/Gaussian Splatting parameters and update them.

The policy gradient term provides an additional direction for the update to move in a direction that performs optimization for the reward obtained by the additional scoring function. In our experiments, we find that \emph{aesthetic score} is highly supportive for 3D generation tasks and helps provide good feedback for the gradient updates.  

Since our technique relies on policy gradient computed from Stable Diffusion, we can also use it with other variants, such as the recent 3D-SDS-based methods, such as GSGen \cite{chen2023text}, shown in the next section, thus showing the versatility of our proposed approach. 

\begin{table*}[!htp]
\caption{The following figure shows the results we obtained using DDPO-based optimization combined with Gaussian Splatting for text-to-3D generation. Our method further enhances the details and has better variation in the generated structures and lower artifacts. Using our method, we also observe better CLIP scores, where a larger CLIP Score suggests higher CLIP similarity between the rendered image and prompt. }
\label{table:comparison}
\scalebox{0.85}{
\begin{tabular}{ccccc}
Method & "a campfire"  & "a small saguaro cactus & "a tulip" \\
       &               &  planted in a clay pot" & \\

  Shap-E \cite{jun2023shap} & 
 \begin{tikzpicture}[spy using outlines={rectangle,magnification=2,size=1.5cm}]
	       \node {\includegraphics[width=0.15\linewidth, trim={1cm 0cm 1cm 1cm }, clip]{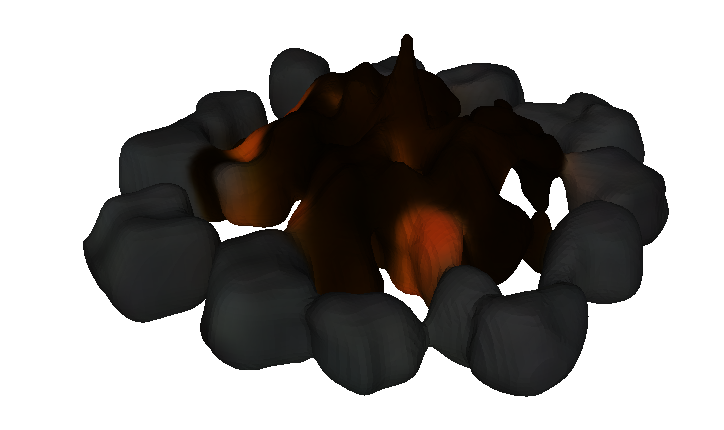}};			
				\node[] at (1, 0) {};
				\spy[color=green] on (0.0,0.2) in node [right] at (1.5, 1);
			\end{tikzpicture}  
   &
   \begin{tikzpicture}[spy using outlines={rectangle,magnification=2,size=1.5cm}]
                \node {\includegraphics[width=0.1\linewidth]{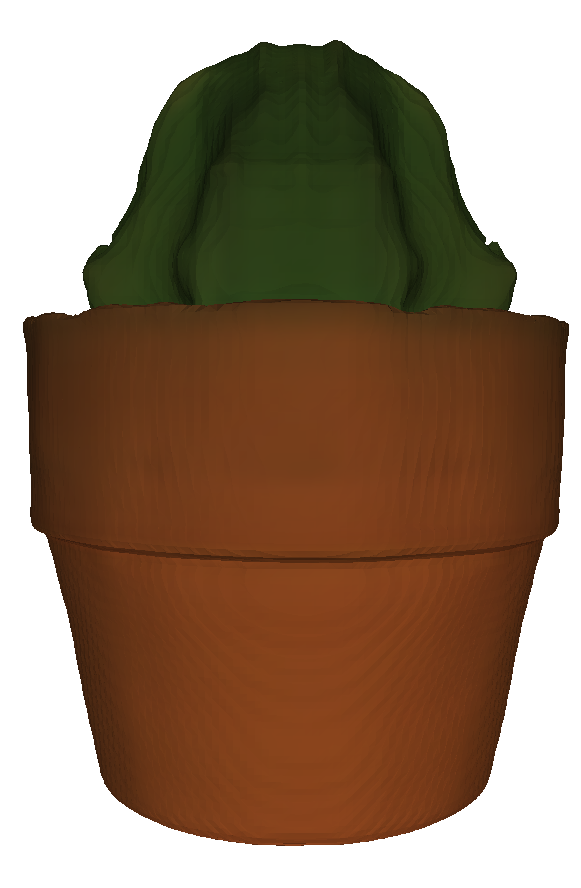}};
				\node[] at (1, 0) {};
				\spy[color=green] on (0.0,0.7) in node [right] at (1.2, 1);
			\end{tikzpicture}
   & 
   \begin{tikzpicture}[spy using outlines={rectangle,magnification=2,size=1.5cm}]
                \node {\includegraphics[width=0.1\linewidth]{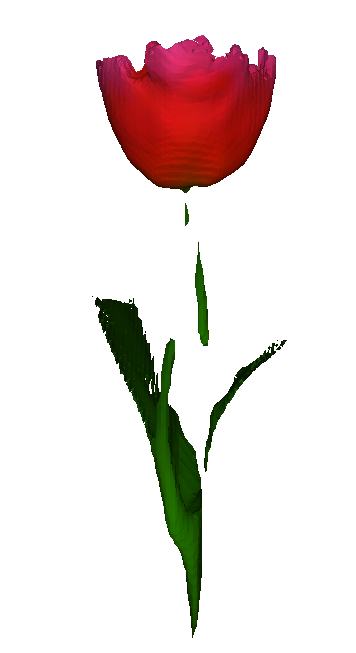}}; 
                
				\node[] at (1, 0) {};
				\spy[color=green] on (0.0,1.) in node [right] at (1., 1);
			\end{tikzpicture} \\

            
    DreamFusion \cite{poole2022dreamfusion} &
    \begin{tikzpicture}[spy using outlines={rectangle,magnification=2,size=1.5cm}]
                \node {\includegraphics[width=0.2\linewidth]{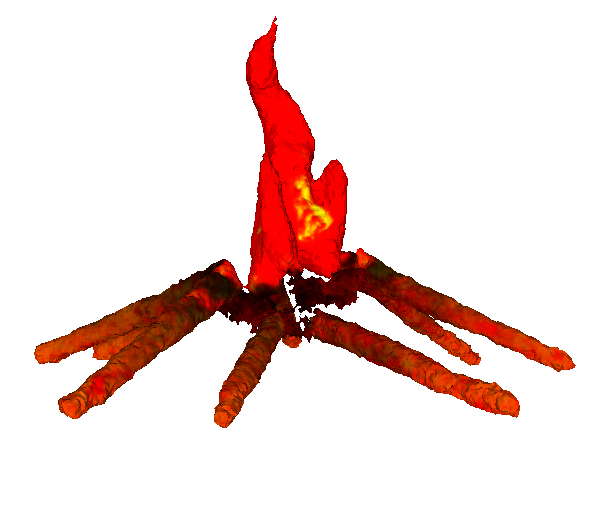}};
				\node[] at (1, 0) {};
				\spy[color=green] on (0.0,0.2) in node [right] at (1, 1);
			\end{tikzpicture} 
   &
    \begin{tikzpicture}[spy using outlines={rectangle,magnification=2,size=1.5cm}]
                \node {\includegraphics[width=0.18\linewidth]{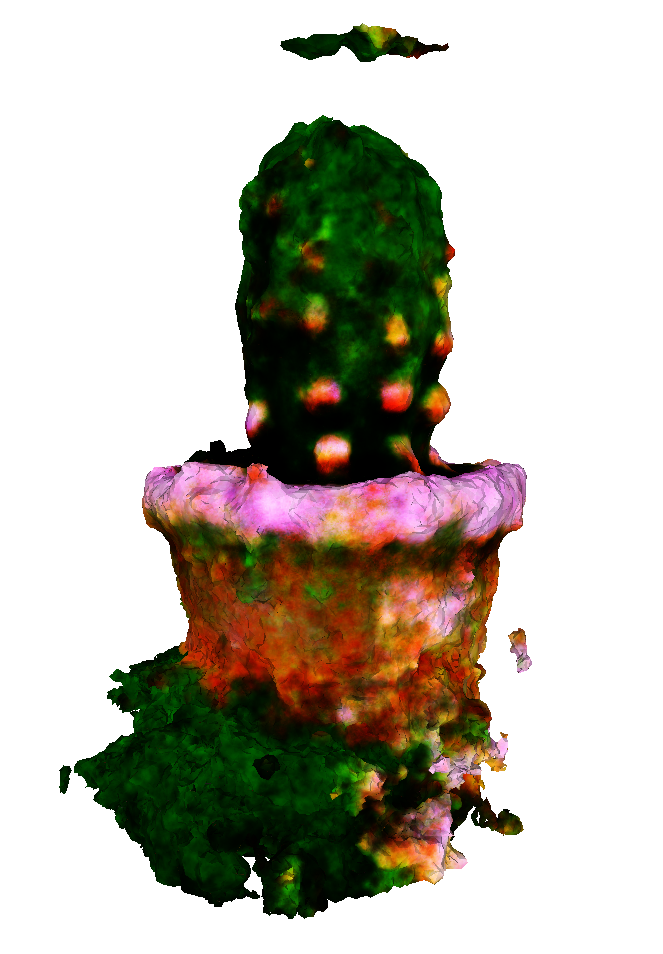}};
				\node[] at (1, 0) {};
				\spy[color=green] on (0.0,0.5) in node [right] at (1, 1);
			\end{tikzpicture}
   
   &
   \begin{tikzpicture}[spy using outlines={rectangle,magnification=2,size=1.5cm}]
                \node {\includegraphics[width=0.12\linewidth]{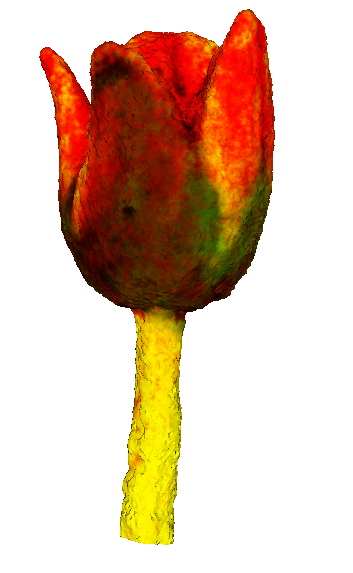}};
				\node[] at (1, 0) {};
				\spy[color=green] on (0.0,0.9) in node [right] at (1, 1);
			\end{tikzpicture}
   \\ 
   
   Dreamgaussian \cite{tang2023dreamgaussian}

   &
     \begin{tikzpicture}[spy using outlines={rectangle,magnification=2,size=1.5cm}]
				\node {\includegraphics[width=0.2\linewidth]{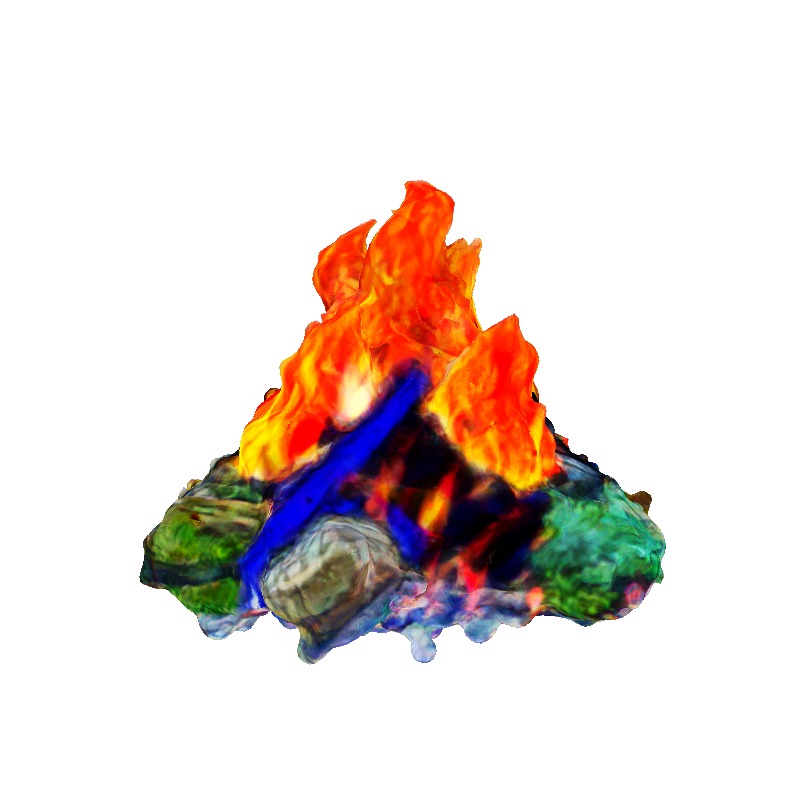}};
				\node[] at (1, 0) {};
				\spy[color=green] on (0.0,0.2) in node [right] at (1, 1);
			\end{tikzpicture}    

    &

   \begin{tikzpicture}[spy using outlines={rectangle,magnification=2,size=1.5cm}]
				\node {\includegraphics[width=0.2\linewidth]{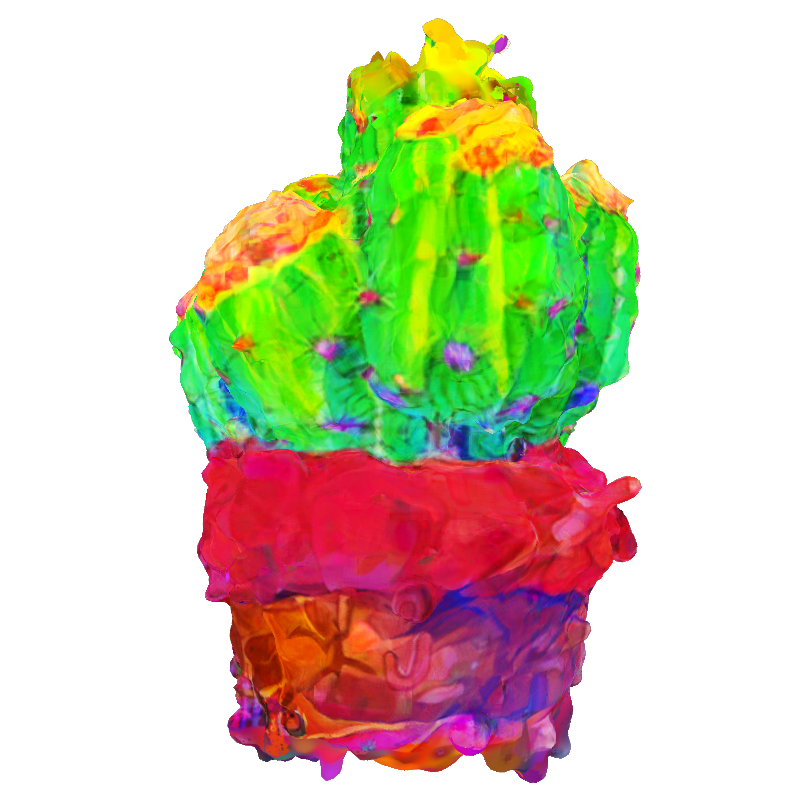}};
				\node[] at (1, 0) {};
				\spy[color=green] on (0.0,0.5) in node [right] at (1, 1);
			\end{tikzpicture}
   & 
   \begin{tikzpicture}[spy using outlines={rectangle,magnification=2,size=1.5cm}]
				\node {\includegraphics[width=0.2\linewidth]{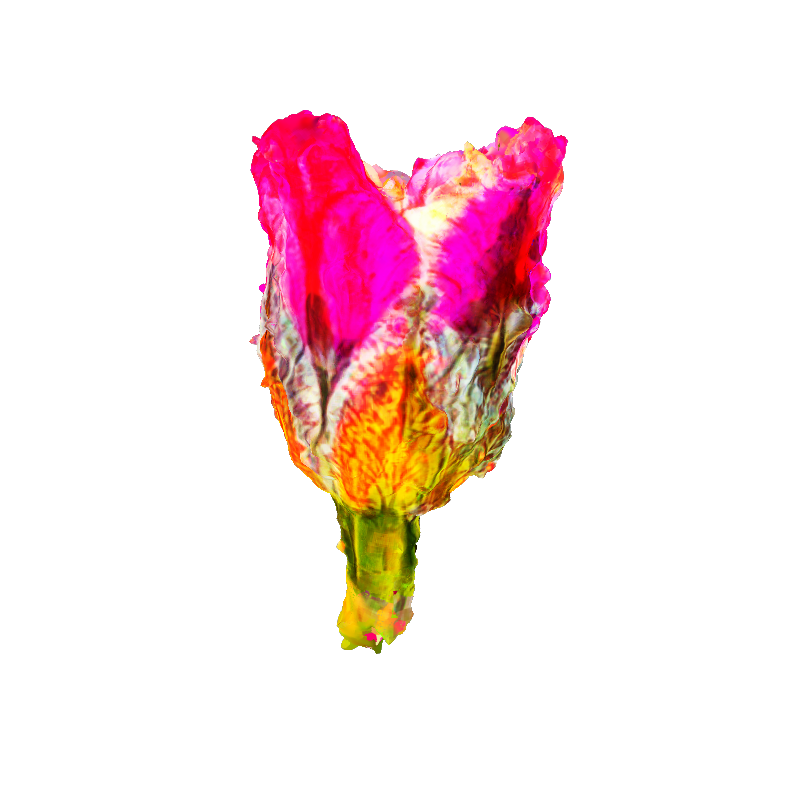}};
				\node[] at (1, 0) {};
				\spy[color=green] on (0.0,0.5) in node [right] at (1, 1);
			\end{tikzpicture}

   \\
   
   Ours (AES guidance)  & 
   \begin{tikzpicture}[spy using outlines={rectangle,magnification=2,size=1.5cm}]
				\node {\includegraphics[width=0.2\linewidth]{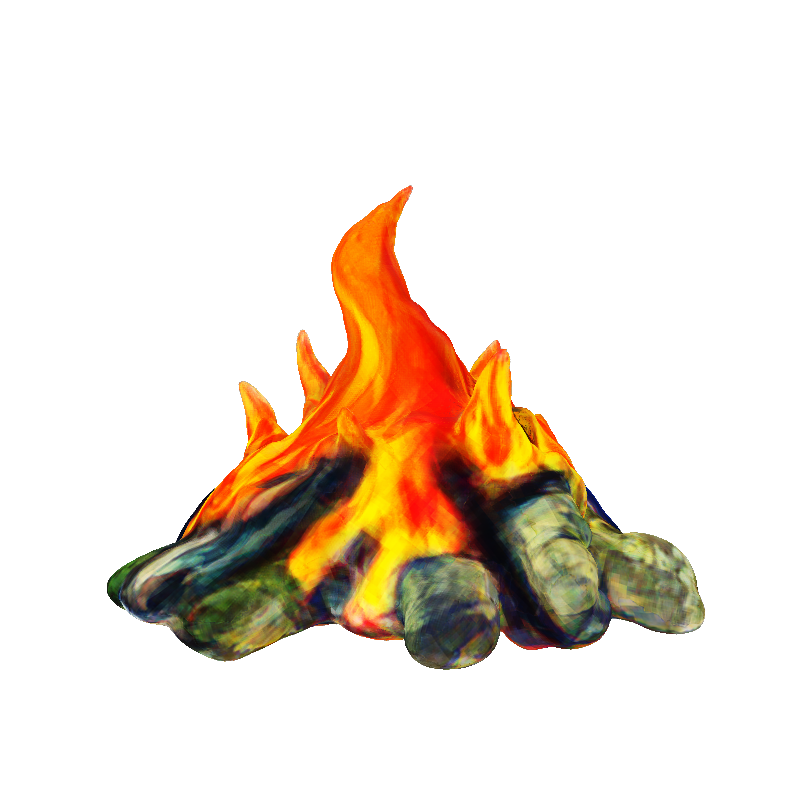}};
				\node[] at (1, 0) {};
				\spy[color=green] on (0.0,0.2) in node [right] at (1, 1);
			\end{tikzpicture}
    & 
    \begin{tikzpicture}[spy using outlines={rectangle,magnification=2,size=1.5cm}]
                \node {\includegraphics[width=0.13\linewidth]{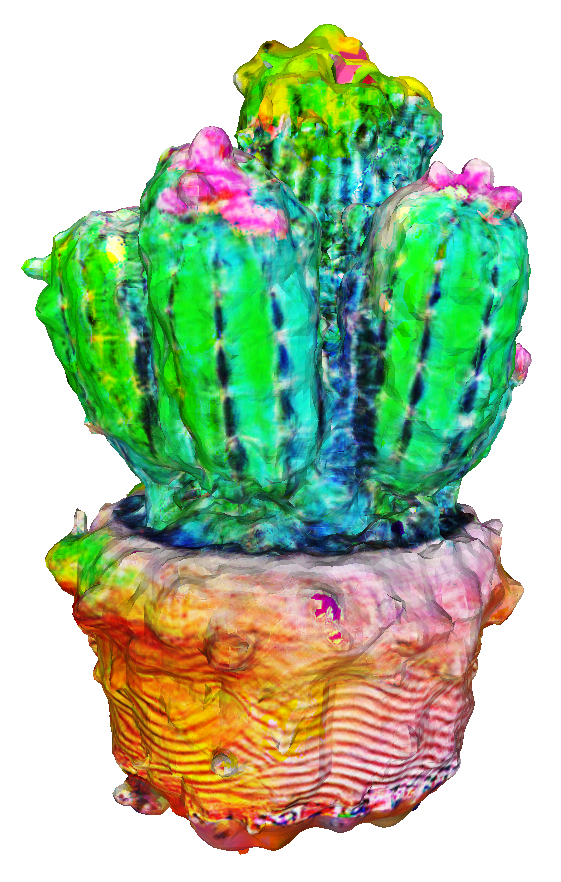}}; 
				\node[] at (1, 0) {};
				\spy[color=green] on (0.2,0.5) in node [right] at (1, 1);
			\end{tikzpicture}   
    &
    \begin{tikzpicture}[spy using outlines={rectangle,magnification=2,size=1.5cm}]
                \node {\includegraphics[width=0.2\linewidth]{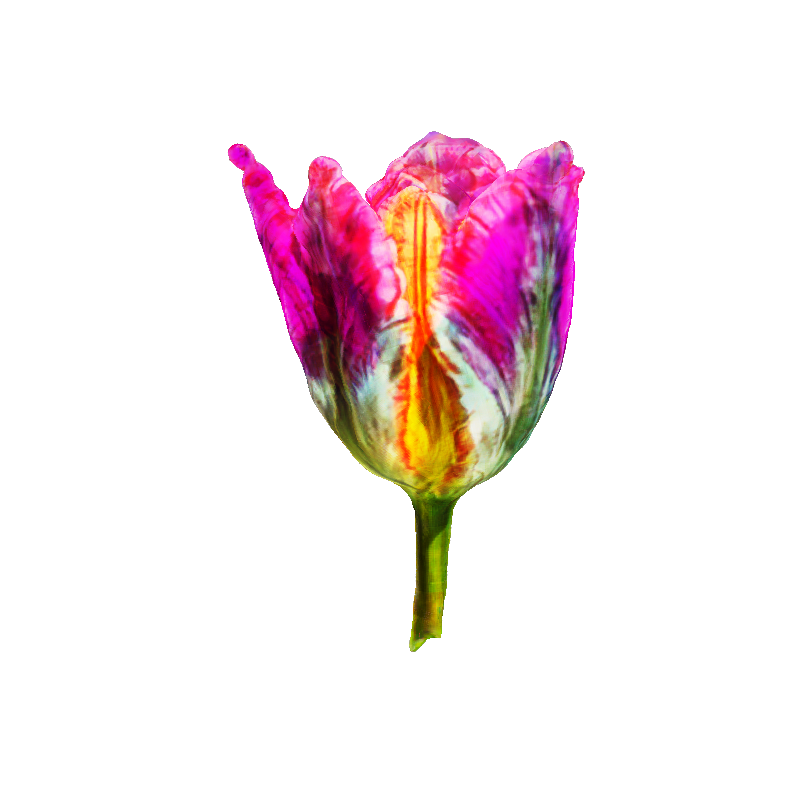}};
				\node[] at (1, 0) {};
				\spy[color=green] on (0.0,0.5) in node [right] at (1, 1);
			\end{tikzpicture}
   \\
  
    Ours (DDPO + AES) &
   \begin{tikzpicture}[spy using outlines={rectangle,magnification=2,size=1.5cm}]
				\node {\includegraphics[width=0.16\linewidth]{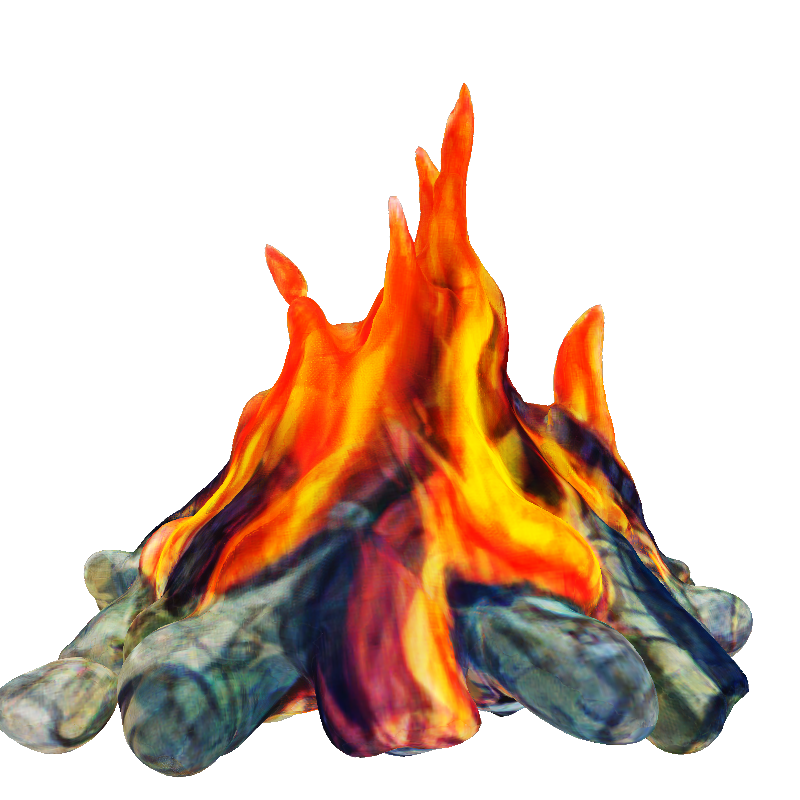}};
				\node[] at (1, 0) {};
				\spy[color=green] on (0.0,0.2) in node [right] at (1, 1);
			\end{tikzpicture}
   & 
    \begin{tikzpicture}[spy using outlines={rectangle,magnification=2,size=1.5cm}]
                \node {\includegraphics[width=0.2\linewidth]{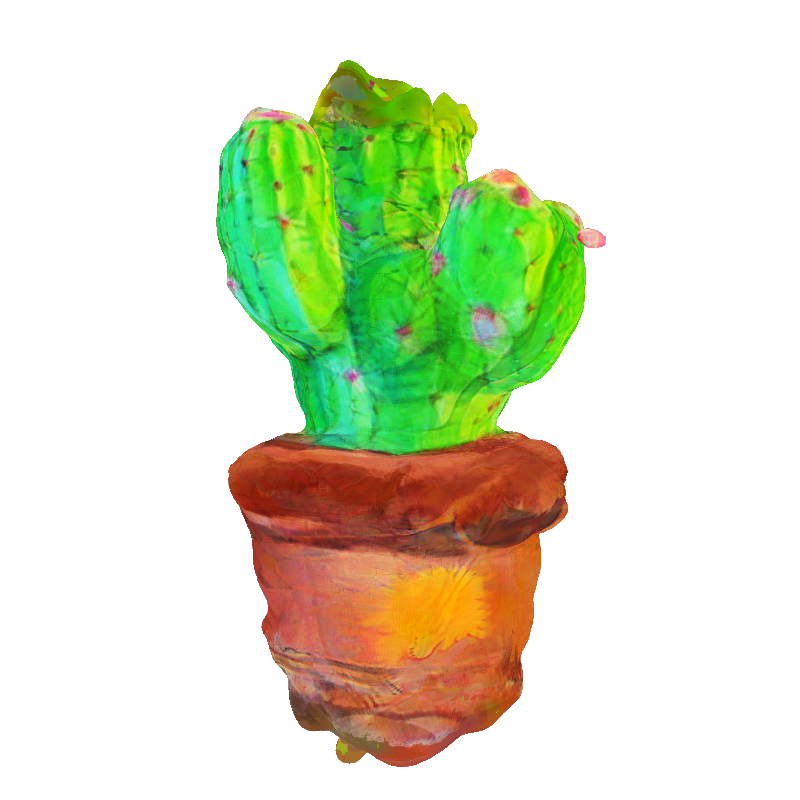}};
				\node[] at (1, 0) {};
				\spy[color=green] on (0.0,0.5) in node [right] at (1, 1);
			\end{tikzpicture}
   
   &
   \begin{tikzpicture}[spy using outlines={rectangle,magnification=2,size=1.5cm}]
				\node {\includegraphics[width=0.15\linewidth]{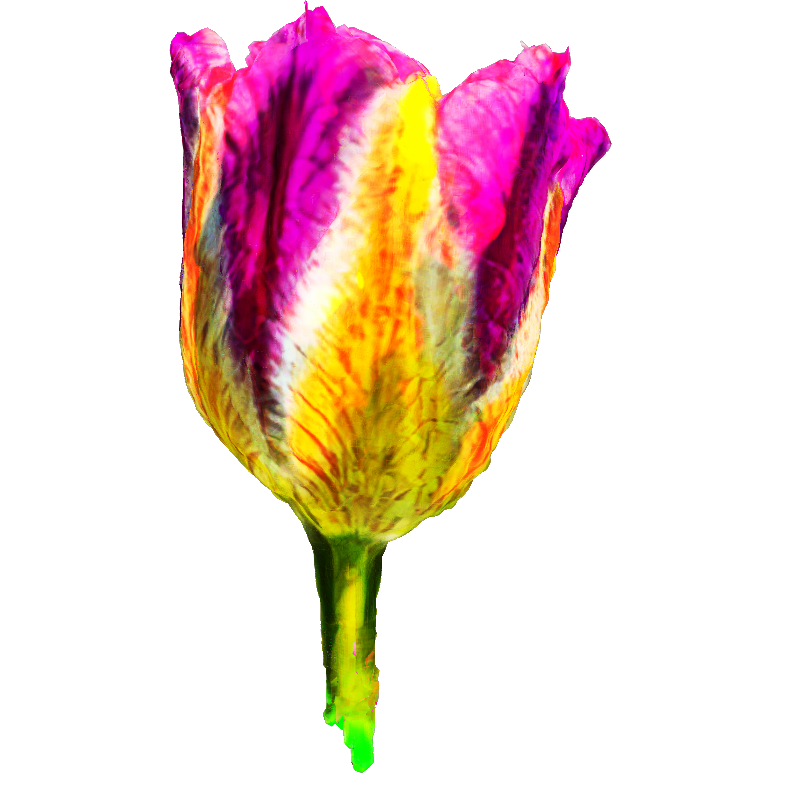}};
				\node[] at (1, 0) {};
				\spy[color=green] on (0.0,0.5) in node [right] at (1, 1);
			\end{tikzpicture}
   
   \\
\end{tabular}}

\end{table*}

\begin{table*}
\centering
\caption{CLIP scores and aesthetic scores for the models shown in \ref{table:comparison}. For each model, we generate 20 images around the object from the resulting mesh and compute the average scores, with higher scores indicating better performance.}
\label{table:metrics_of_table1}
\begin{tabular}{|l|c|c|c|c|c|c|} 
\hline
\multirow{2}{*}{Method} & \multicolumn{2}{l|}{\textit{"A campfire"}} & \multicolumn{2}{l|}{\begin{tabular}[c]{@{}l@{}}\textit{"A small saguaro cactus}\\~\textit{planted in a clay pot"}\end{tabular}} & \multicolumn{2}{l|}{\textit{"A photo of a tulip"}}  \\ 
\cline{2-7}
                        & Aesthetic & CLIP (\%)           & Aesthetic & CLIP (\%)                                                                                       & Aesthetic & CLIP (\%)                     \\ 
\hline
Shap-E                  & 4.51      & 28.09               & 4.17      & 23.94                                                                                           & 4.45      & 23.28                         \\ 
\hline
DreamFusion             & 4.59      & 29.02               & 3.9       & 26.72                                                                                           & 4.42      & 27.05                         \\ 
\hline
Dreamgaussian           & 4.88      & 30.69               & 5.02      & 27.42                                                                                           & 4.91      & 29.51                         \\ 
\hline
Ours (AES)              & 4.85      & 30.46               & 4.98      & 27.25                                                                                           & \textbf{5.21}      & \textbf{32.61}                         \\ 
\hline
Ours (DDPO + AES)       & \textbf{5.16}      & \textbf{31.18}               & \textbf{5.14}      & \textbf{31.29                                             }                                              & 5.07      & 31.58                         \\
\hline
\end{tabular}
\end{table*}

%% file: sec/results_1.tex
\section{Experiments, Results, and Discussion}


We start by testing the AES term with several methods to understand its impact on SDS-based methods. In order to evaluate our method, we use CLIP score to check the semantic relatedness by measuring the cosine similarity between the CLIP prompt embedding and the rendered image embedding. 
During our experimentation, we observed that image-to-3D settings offer lower modifications with respect to the AES score due to a tighter adherence to the image provided, while text-to-3D is a better setting for our approach since the generation can be guided a lot more flexibly due to the lack of ground truth in such a synthesis. 
Therefore, we perform experiments primarily in a text-to-3D setting with DreamGaussian due to its fast generation speed. However, we also demonstrate the effectiveness of our method with other methods, such as DreamFusion and GSGen, which also use SDS-based text-to-3D synthesis. 

In the case of DreamFusion, we see noticeable improvement when combined with aesthetic scorer; not only do we observe the improvement in the textures but also an improvement in the geometry of the generated objects, as discussed in the next section.

With the usage of DDPO3D, we again notice a substantial improvement in text-to-3D setting with DreamGaussian demonstrated in Table \ref{table:comparison}.

\subsection{Impact of AES Term}

Our findings emphasize the vital role of the aesthetic scorer in enhancing the visual quality of generated assets. Additionally, this scorer proves versatile, enabling its integration with policy gradients and facilitating improvements over the baseline DreamGaussian model.

\begin{figure}[!htp]
    \centering
    ``a DSLR photo of a burger''
    \begin{tabular}{c c}
         DreamFusion~\cite{poole2022dreamfusion} & \textbf{\small Ours}+DreamFusion \\
         \includegraphics[trim={0cm 0.5cm 0cm 0cm},clip,width=0.45\linewidth]{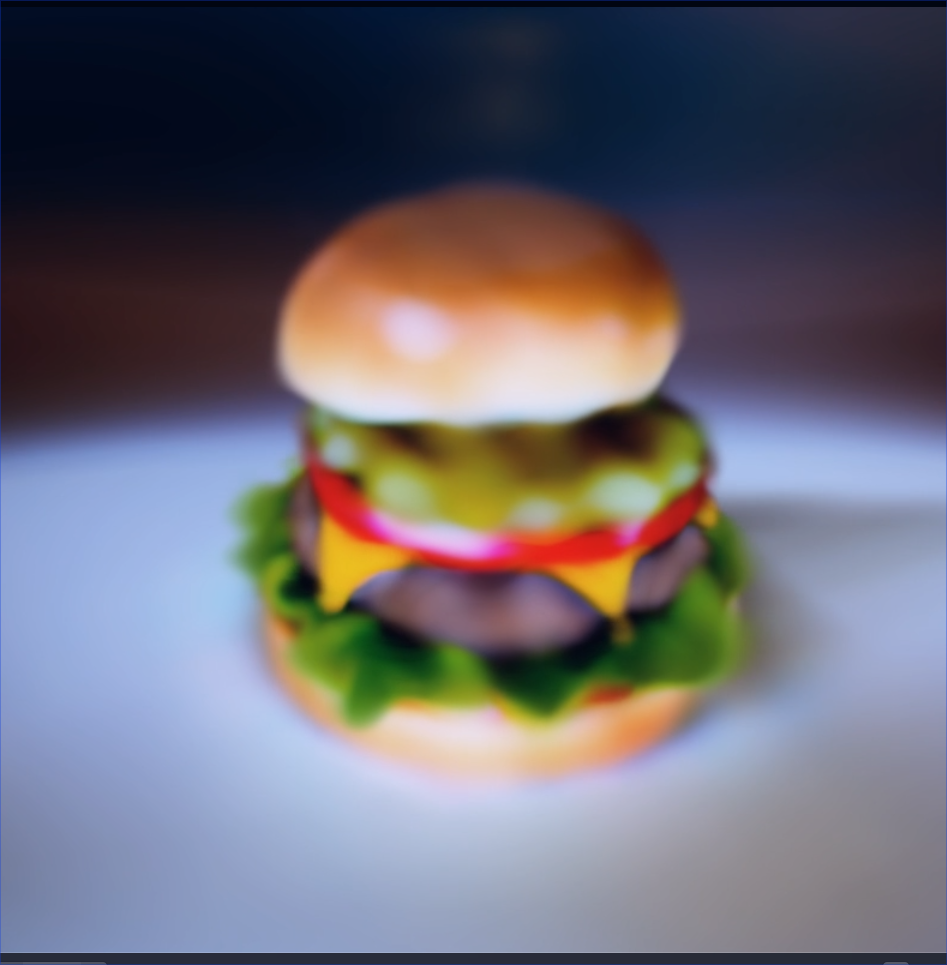}
         &\includegraphics[width=0.45\linewidth]{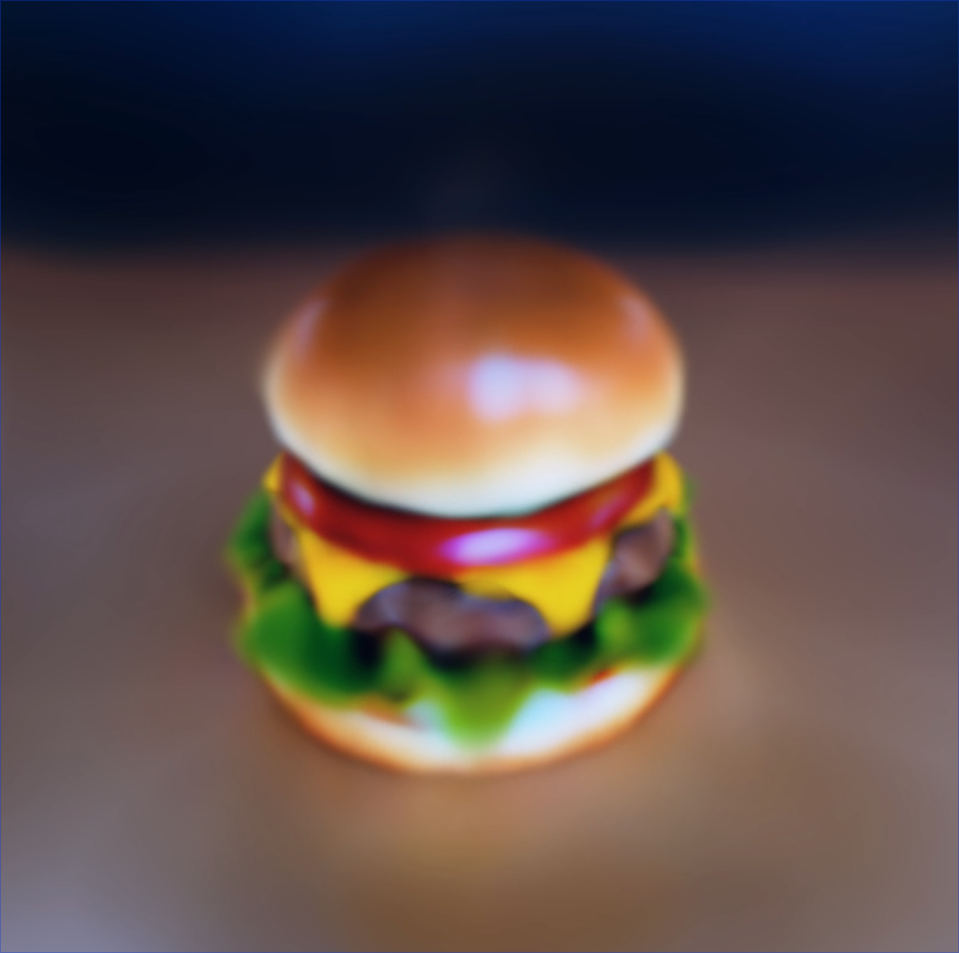}  \\
         CLIP Score: $\uparrow$ : $32.01$ & CLIP Score: $\uparrow$ : $\mathbf{34.72}$ \\
         
         \includegraphics[trim={0cm 0cm 0cm 5cm},clip,width=0.45\linewidth]{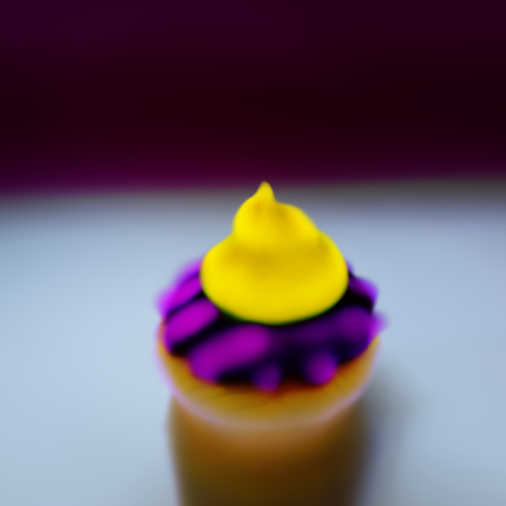}
         &
         \includegraphics[trim={0cm 0cm 2cm 6.2cm},clip,width=0.45\linewidth]{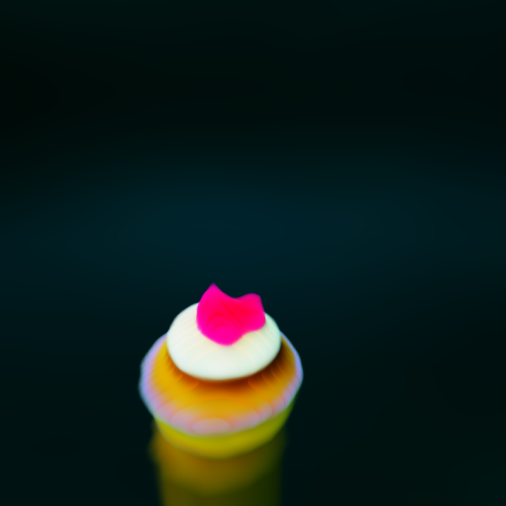}  \\
         CLIP Score: $\uparrow$ : $29.57$ & CLIP Score: $\uparrow$ : $\mathbf{31.19}$
    \end{tabular}
    \caption{Images of burger generated in the same number of iterations using AES (right) and without AES (left). 
    and similarly, the second row shows the cupcake generated in a similar setting without (left) and with (right) AES. 
     The burger image generated by our method seems more symmetric and sharper.}
    \label{fig:dreamfusion-ddpo1}
\end{figure}

\begin{figure}[!htp]
    \centering
    ``a DSLR photo of a car made out of cheese''
    \begin{tabular}{c c}
         GSGen \cite{chen2023text} & \textbf{\small Ours}+GSGen \\
         \includegraphics[width=0.35\linewidth]{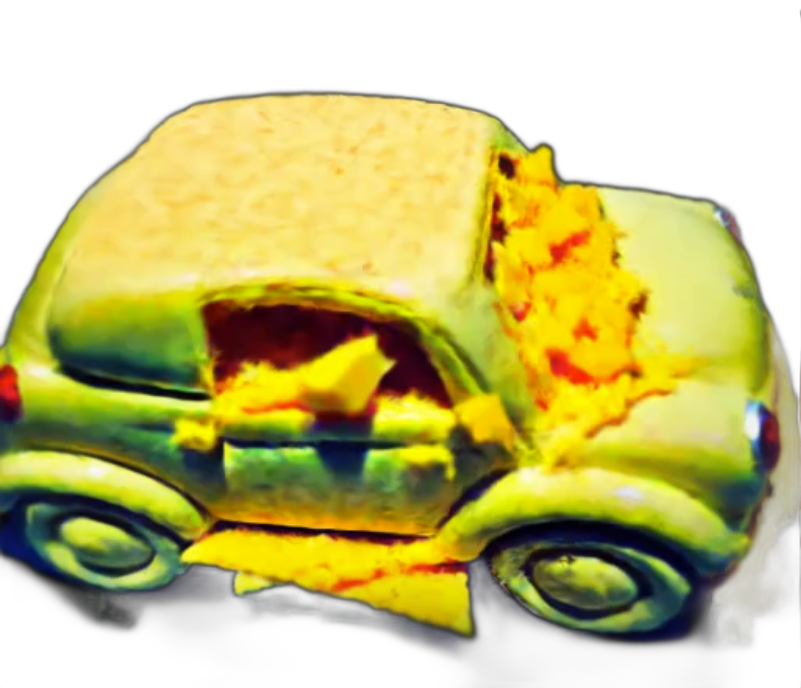}
         
         &\includegraphics[trim={0cm 0cm 0cm 0.5cm},clip,width=0.35\linewidth]{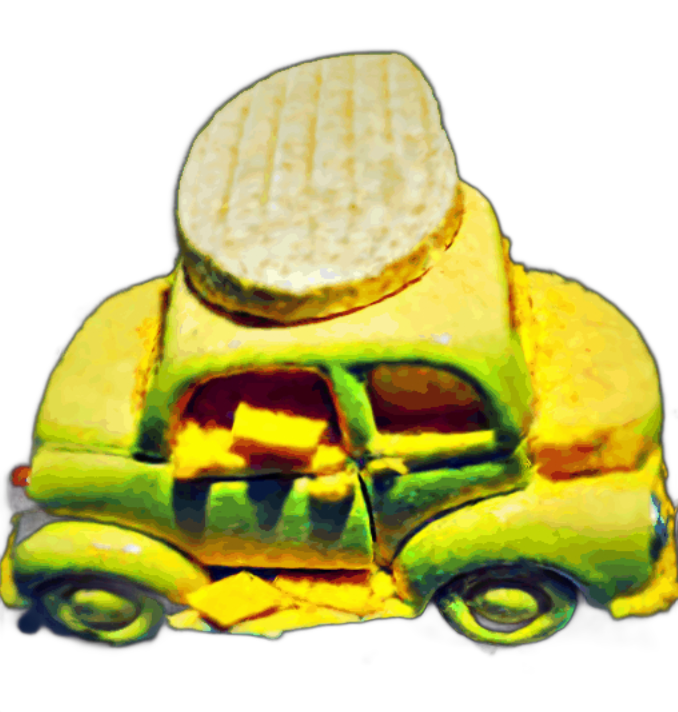}  \\
        
    CLIP Score: $\uparrow$ : 32.53 & CLIP Score: $\uparrow$ : $\mathbf{35.16}$
    \end{tabular}
    \caption{Assets generated in the same number of iterations using AES (right) based optimization and without AES (left) using GSGen \cite{chen2023text}. 
    }
    \label{fig:gsgen-ddpo1}
\end{figure}

In Figure \ref{fig:dreamfusion-ddpo1}, we can visually see the difference in the quality of the generation. The contents of the burger in the AES-based generation look more symmetric and organized as compared to the original generation. In both cupcake and burger, we achieved higher clip scores of $31.19$ \&  $34.72$ respectively, as compared to the original scores of  $29.57$ \& $32.01$.
Similarly, we achieve better metrics with GSGen as demonstrated in Figure \ref{fig:gsgen-ddpo1}, where our method achieved a score of $35.1584$ as compared to the original score of $32.53$.
Further, we observe that the AES-based optimization consistently yields better results along with improved metrics in most cases. 

\begin{figure}[!htp]
    \centering

    \scalebox{0.65}{
    \begin{tabular}{c}
    
    DreamGaussian \cite{tang2023dreamgaussian} \\
    \begin{tikzpicture}[spy using outlines={rectangle,magnification=2,size=2cm}]
                    \node {\includegraphics[width=0.5\linewidth]{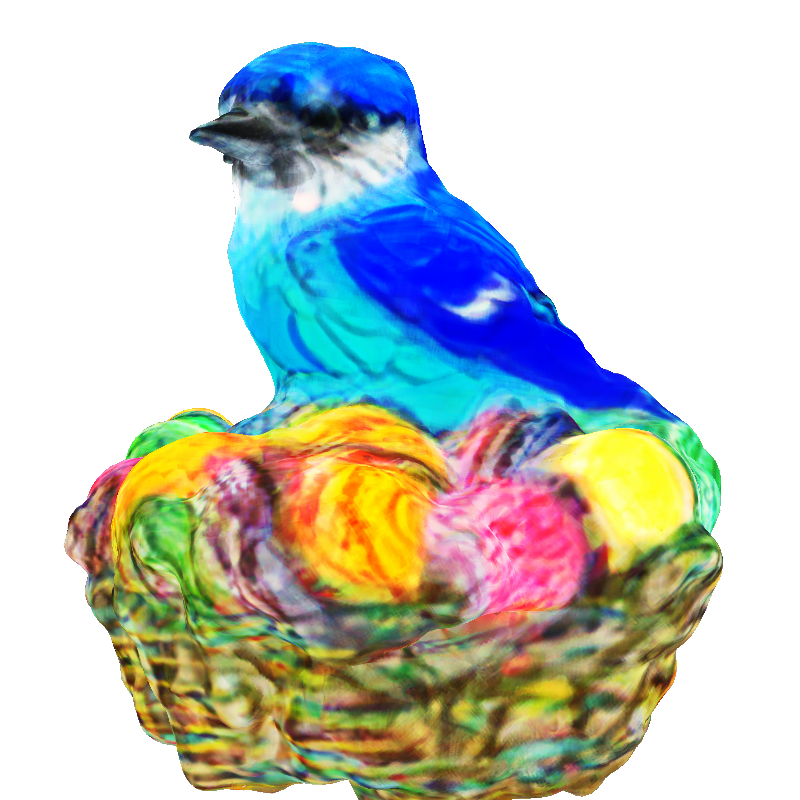}};
				\node[] at (1, 0) {};
				\spy[color=green] on (0.7,-0.) in node [left] at (4, 1);
                \node[] at (1, 0) {};
				\spy[color=green] on (-0.5,1.4) in node [right] at (-4, 1);
			\end{tikzpicture} \\
    \textbf{Ours} + DreamGaussian \\ 
    \begin{tikzpicture}[spy using outlines={rectangle,magnification=2,size=2cm}]
				\node {\includegraphics[width=0.5\linewidth]{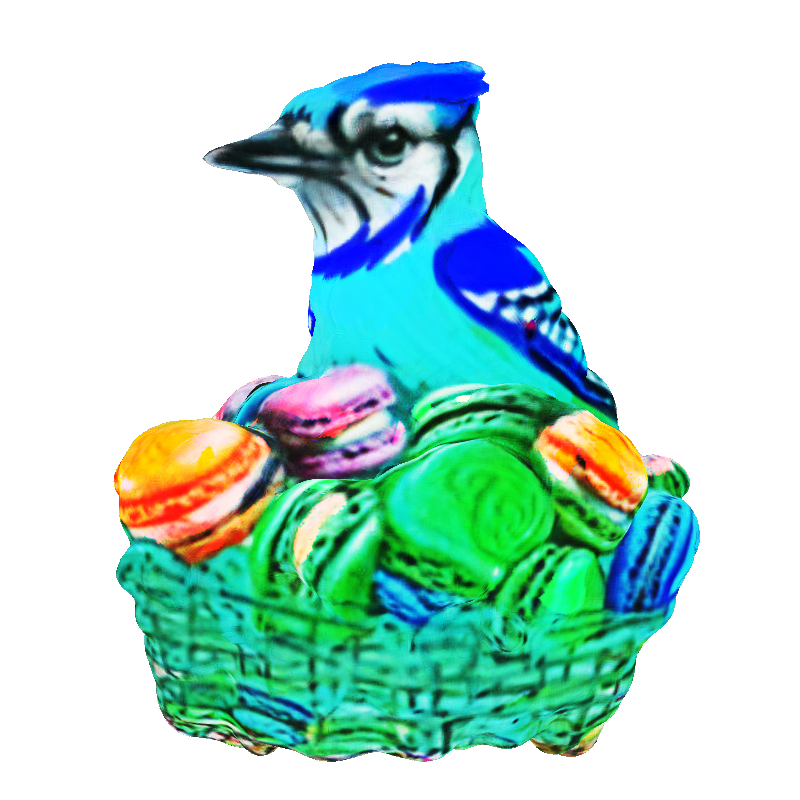}};
				\node[] at (1, 0) {};
				\spy[color=green] on (0.7,-0.) in node [left] at (4, 1);
                \node[] at (1, 0) {};
				\spy[color=green] on (-0.4,1.4) in node [right] at (-4, 1);
			\end{tikzpicture} \\


   
   \end{tabular}}
    \caption{A zoom-in view of the differences in generation using our method (row 2) and the original DreamGaussian (row 1), for the prompt ``A blue jay standing on a large basket of rainbow macarons''. }
    \label{fig:ddpo_stage_1_only}
\end{figure}
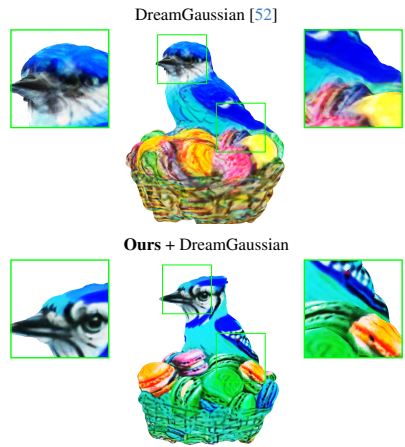
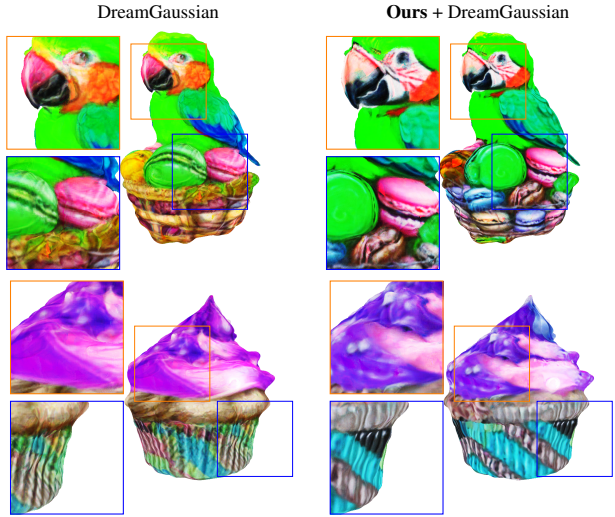
\begin{figure}[!htp] 
\centering
  \scalebox{0.5}{
    \begin{tabular}{c c}
    \Large
    DreamGaussian & \Large \textbf{Ours} + DreamGaussian \\
    
    \begin{tikzpicture}[spy using outlines={rectangle,magnification=1.5,size=3cm}]
                \node {\includegraphics[width=0.75\linewidth]{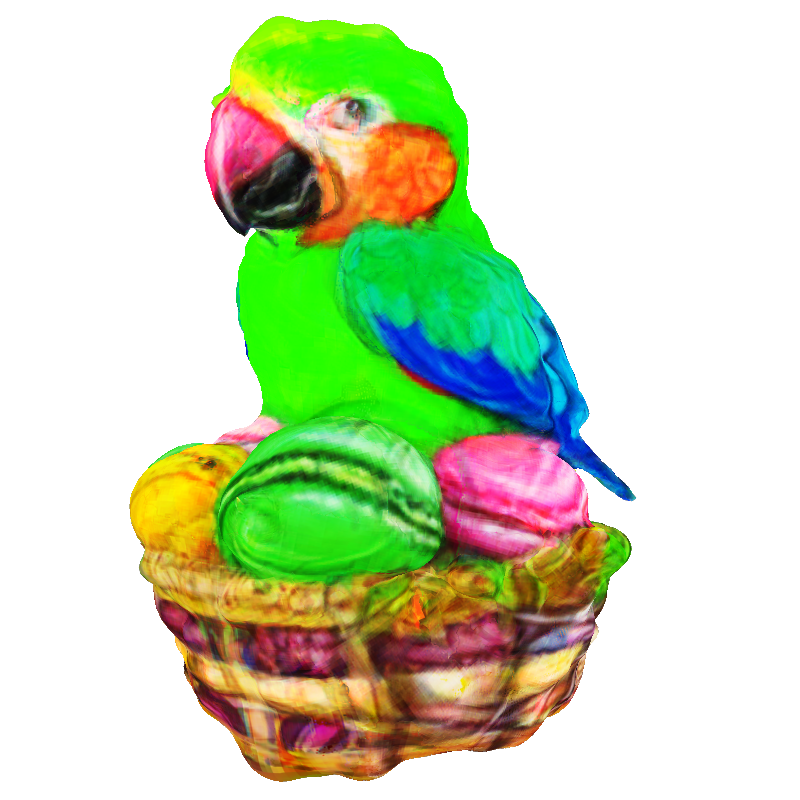}};
				\node[] at (1, 0) {};
				\spy[color=blue] on (0.4,-0.9) in node [left] at (-2.,-2.);
                \node[] at (1, 0) {};
				\spy[color=orange] on (-0.7,1.5) in node [right] at (-5,1.2);
			\end{tikzpicture} 
   &
     
    \begin{tikzpicture}[spy using outlines={rectangle,magnification=1.5,size=3cm}]
                \node {\includegraphics[width=0.75\linewidth]{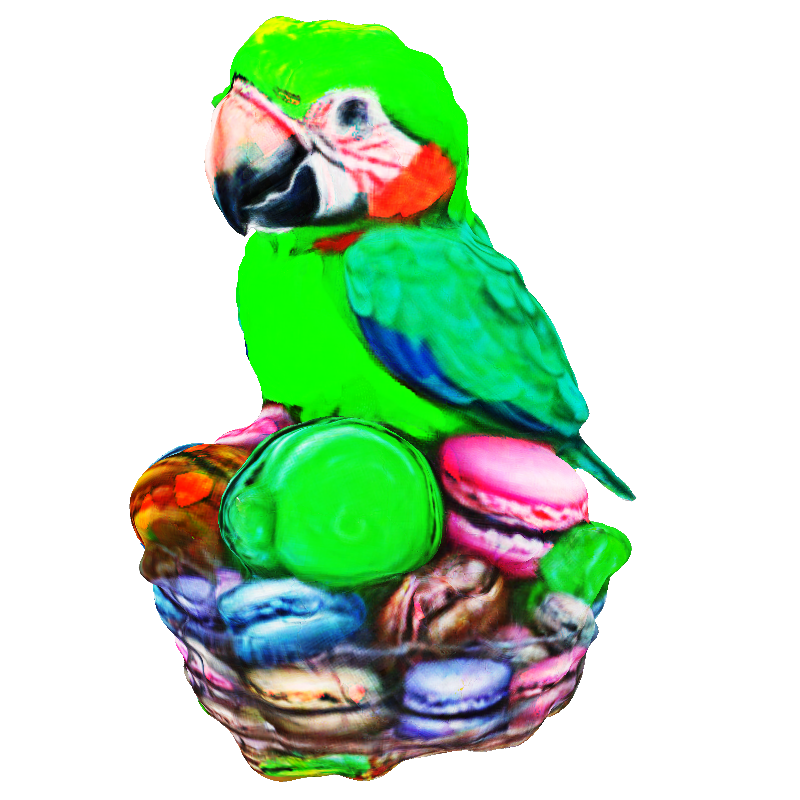}};
				\node[] at (1, 0) {};
				\spy[color=blue] on (0.4,-0.9) in node [left] at (-2., -2.);
                \node[] at (1, 0) {};
				\spy[color=orange] on (-0.7,1.5) in node [right] at (-5, 1.2);
			\end{tikzpicture} \\
        
    \begin{tikzpicture}[spy using outlines={rectangle,magnification=1.5,size=3cm}]
                \node {\includegraphics[width=0.7\linewidth]{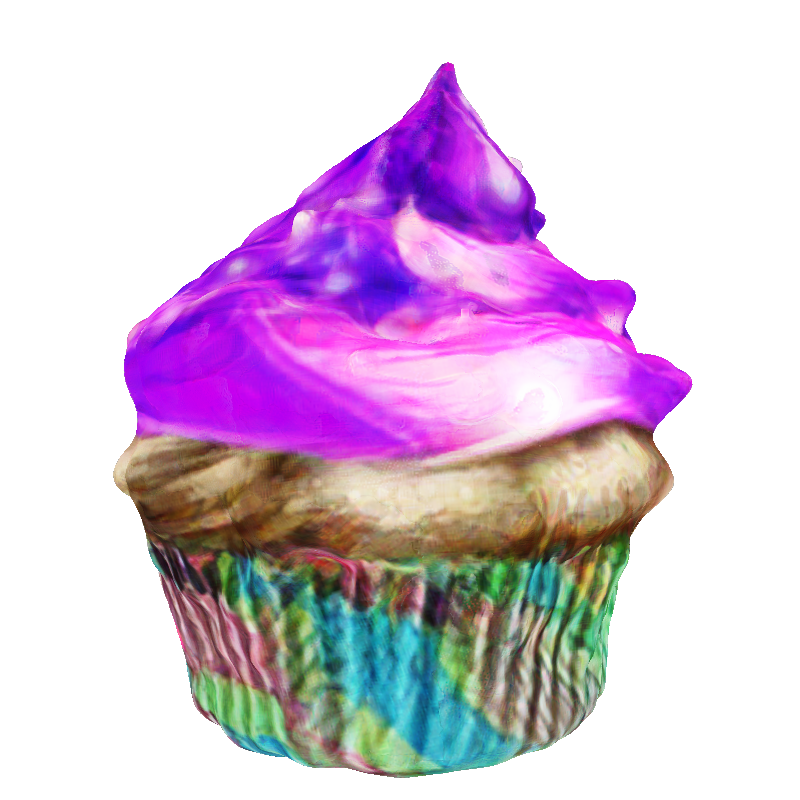}};
				\node[] at (1, 0) {};
				\spy[color=blue] on (1.5,-1.5) in node [left] at (-2., -2.);
                \node[] at (1, 0) {};
				\spy[color=orange] on (-0.7,0.5) in node [right] at (-5, 1.2);
			\end{tikzpicture} 
   &
    \begin{tikzpicture}[spy using outlines={rectangle,magnification=1.5,size=3cm}]
                \node {\includegraphics[width=0.7\linewidth]{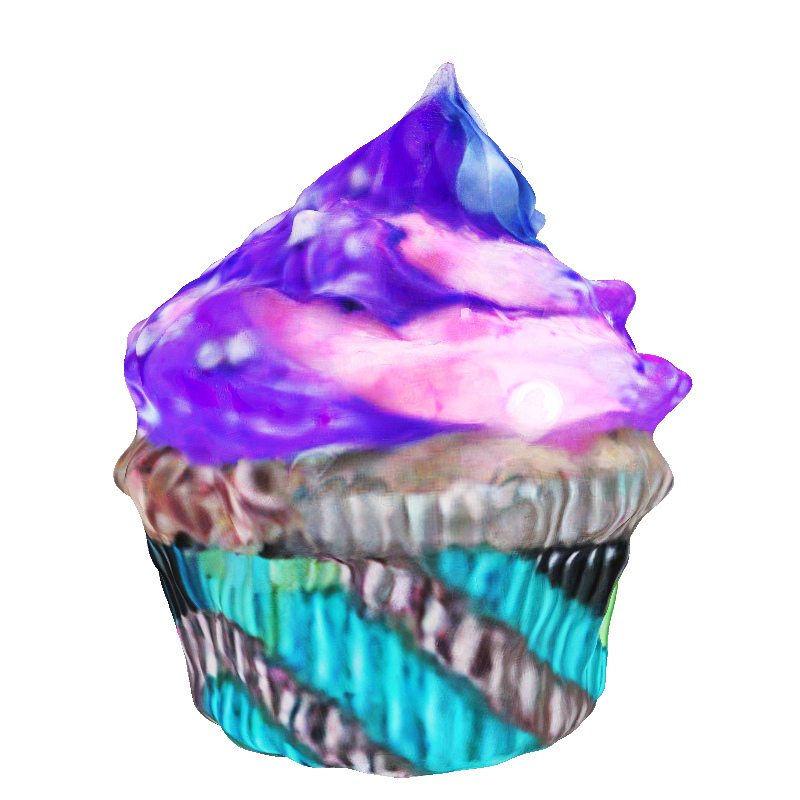}};
				\node[] at (1, 0) {};
				\spy[color=blue] on (1.5,-1.5) in node [left] at (-2., -2.);
                \node[] at (1, 0) {};
				\spy[color=orange] on (-0.7,0.5) in node [right] at (-5, 1.2);
			\end{tikzpicture} \\
   \end{tabular}}
    \caption{A zoom-in view of the differences in generation using our method and the original DreamGaussian in only stage two of the optimization process, for the prompts ``a parrot sitting on a basket of macarons'' and ``a photo of a cupcake''.}
    \label{fig:ddpo_stage2_only}
\end{figure}

We notice a similar improvement in the case of DreamGaussian as well as evident from Figure \ref{fig:ddpo_stage_1_only} and Table \ref{table:comparison}. Our method consistently outperforms others in terms of both the aesthetic scores and CLIP scores as shown in Table \ref{table:metrics_of_table1} with a considerable difference in scores. 

We tested our method with text to 3D setting in DreamGaussian, whereby the policy gradient updates are used to update the Gaussian splats. The text-to-3D renderings of Gaussian splats are not at par with image-to-3D renderings and tend to suffer from artifacts. We observe that DDPO-based optimization not only reduces these artifacts but also improves upon the structure of the generated mesh. 

In order to observe how the AES term impacts the generation process, we increase the AES term weight to further steer the optimization towards a more aesthetically pleasing result. 
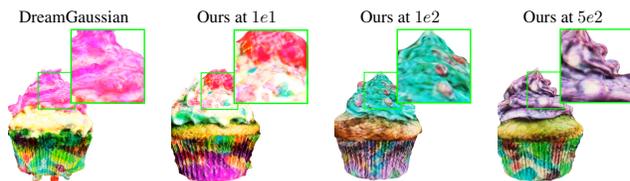
\begin{figure}[!htp]
    \scalebox{0.65}{
    \begin{tabular}{ccccc}
         DreamGaussian & Ours at $1e1$ & Ours at $1e2$ & Ours at $5e2$  \\
          \begin{tikzpicture}[spy using outlines={rectangle,magnification=2,size=1.5cm}]
				\node { \includegraphics[trim={6cm 0cm 4cm 0cm},clip,width=0.25\linewidth]{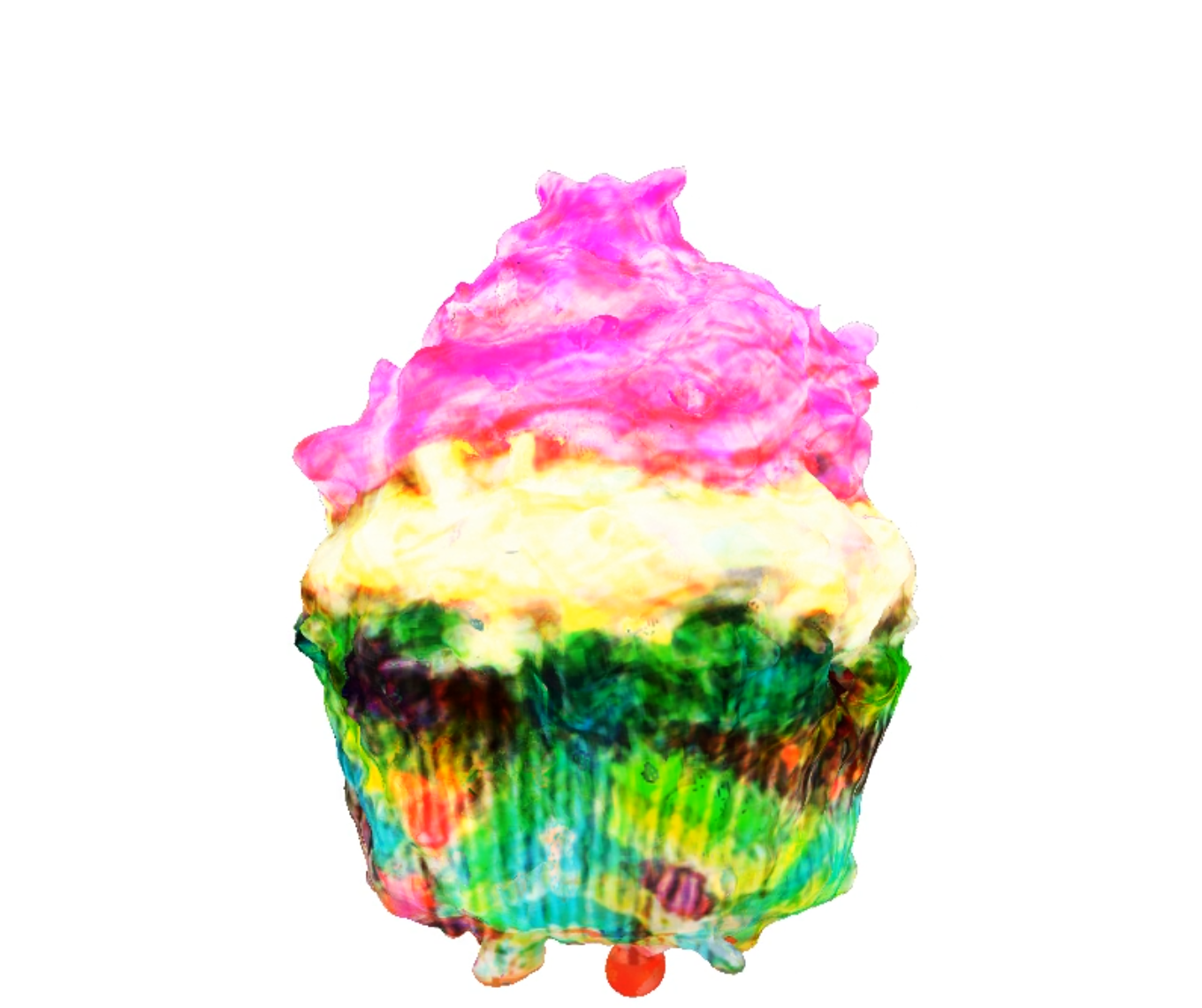}};
				\node[] at (1, 0) {};
				\spy[color=green] on (0.0,0.5) in node [right] at (0.3, 1);
			\end{tikzpicture}
        &
        \begin{tikzpicture}[spy using outlines={rectangle,magnification=2,size=1.5cm}]
				\node { \includegraphics[trim={6cm 0cm 4cm 0cm},clip,width=0.25\linewidth]{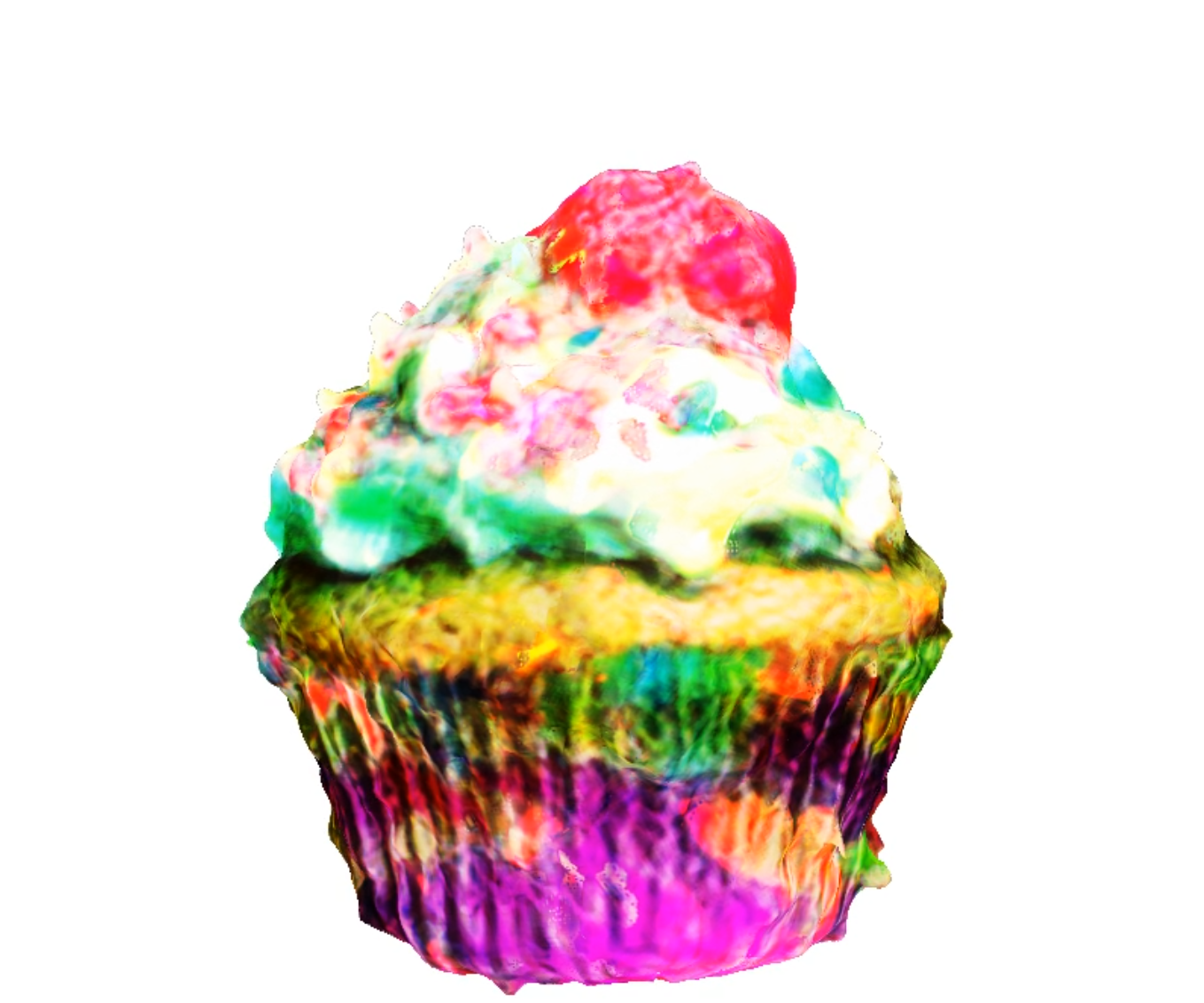}};
				\node[] at (1, 0) {};
				\spy[color=green] on (0.0,0.5) in node [right] at (0.3, 1);
			\end{tikzpicture}
         &
        \begin{tikzpicture}[spy using outlines={rectangle,magnification=2,size=1.5cm}]
				\node { \includegraphics[trim={6cm 0cm 4cm 0cm},clip,width=0.25\linewidth]{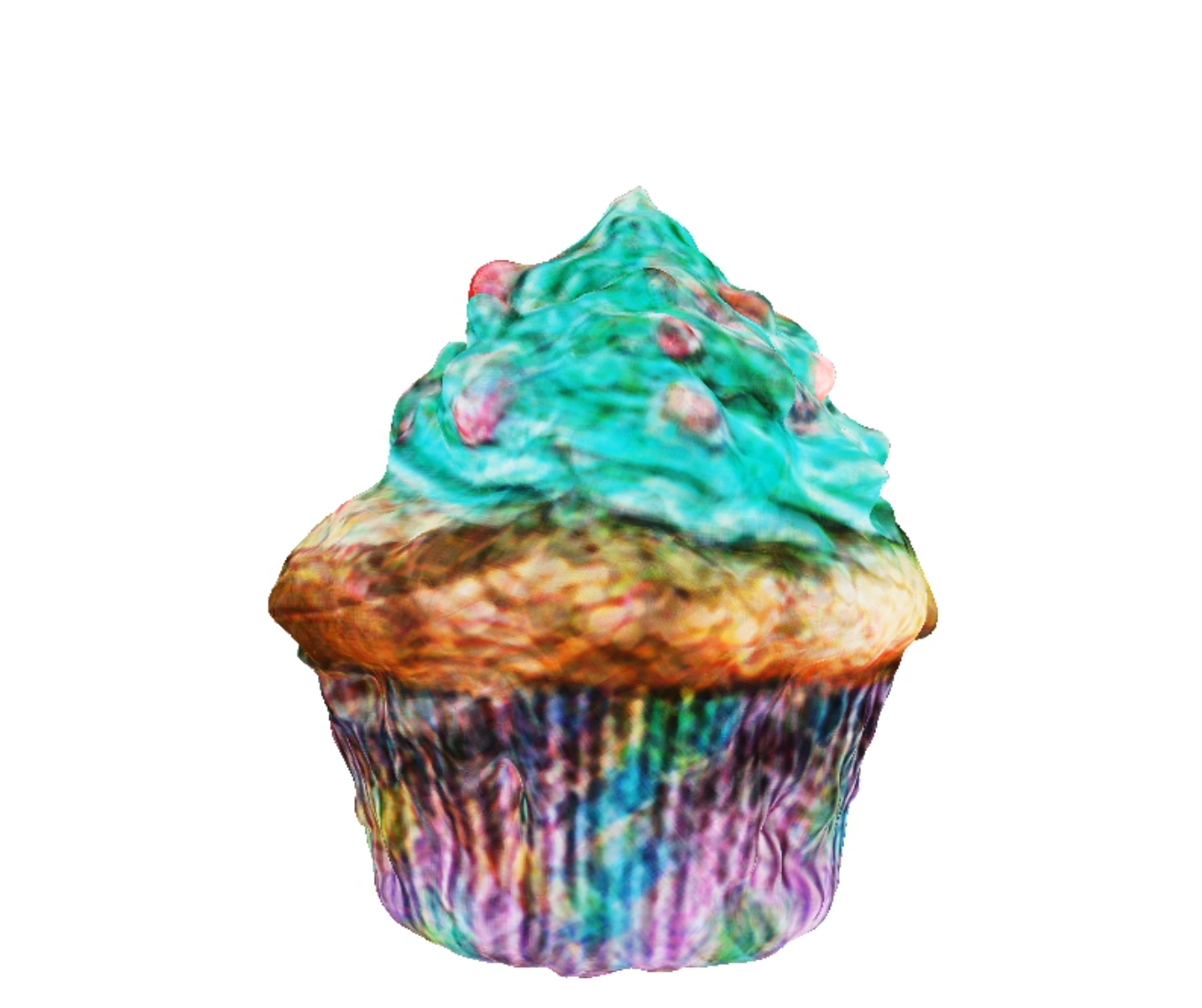}};
				\node[] at (1, 0) {};
				\spy[color=green] on (0.0,0.5) in node [right] at (0.3, 1);
			\end{tikzpicture}
         &
        
        \begin{tikzpicture}[spy using outlines={rectangle,magnification=2,size=1.5cm}]
				\node {\includegraphics[trim={6cm 0cm 4cm 0cm},clip,width=0.25\linewidth]{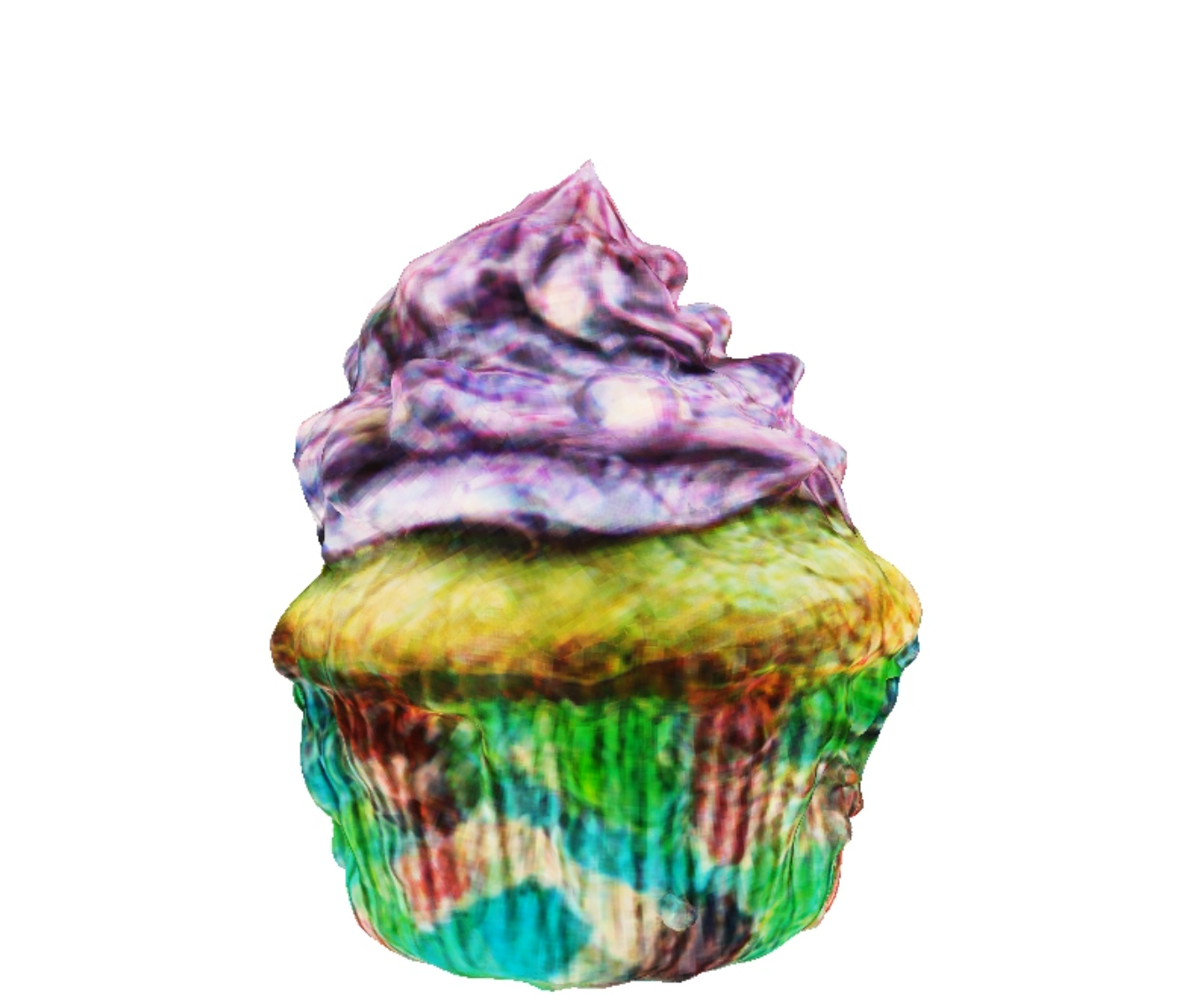}};
				\node[] at (1, 0) {};
				\spy[color=green] on (0.0,0.5) in node [right] at (0.3, 1);
			\end{tikzpicture}
        
    \end{tabular}}
    
    \caption{Changes in the appearance with increasing weightage of the AES term.}
    \label{fig:aes_weight_increase}
\end{figure}

DDPO3D employs SDS in two stages: during the training of the Gaussian splat and the final stage of UV refinement. Our term assumes distinct roles in each scenario. In the initial stage, noticeable geometric improvements are evident, as illustrated in \ref{fig:ddpo_stage_1_only}. Additionally, a significant enhancement in visual appearance is observed when our term is applied. To achieve this, we initially train the first stage following the original DreamGaussian method and introduce our term solely in the second stage. Remarkably, with $100-200$ iterations in stage two, our method surpasses existing approaches in terms of visual quality, as depicted in Figure \ref{fig:ddpo_stage2_only}.

\subsection{Limitations \& Future Scope}

We recognize a visible trade-off between runtime and quality arising from the inclusion of the additional policy gradient term. This is attributed to the increased computation of the reward function, a focus area for future improvements. Similar to many regularization terms, the RL term requires fine-tuning dependent on the magnitude of the SDS terms. Effective results were obtained in the range of $1e1$ to $5e2$, particularly when used in conjunction with AES only. The impact of weighting on generation is illustrated in Figure \ref{fig:aes_weight_increase}, where enhanced details are initially achieved but may diminish after a certain threshold (in this case, $5e2$).

\begin{figure}[!htp]
    \centering
    \begin{tabular}{c c}
         \includegraphics[width=0.23\linewidth]{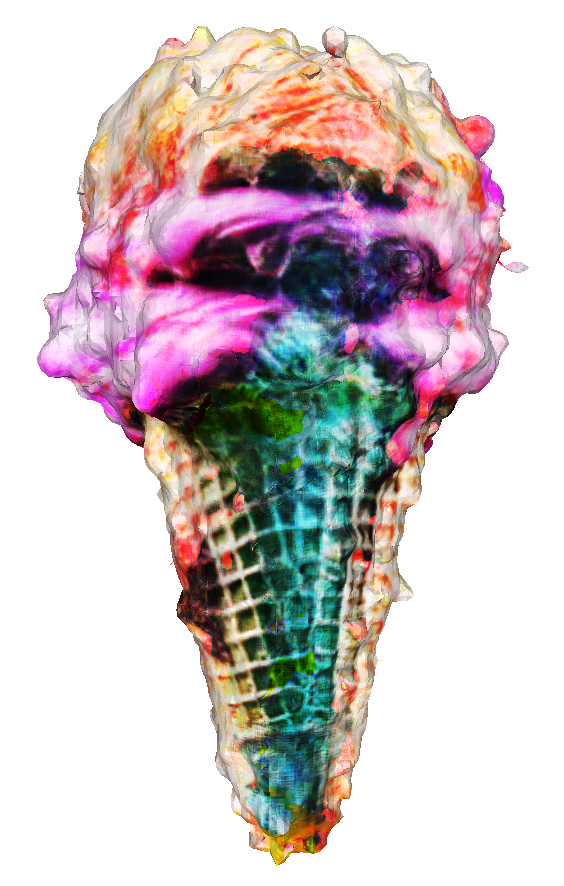}&
         \includegraphics[width=0.2\linewidth]{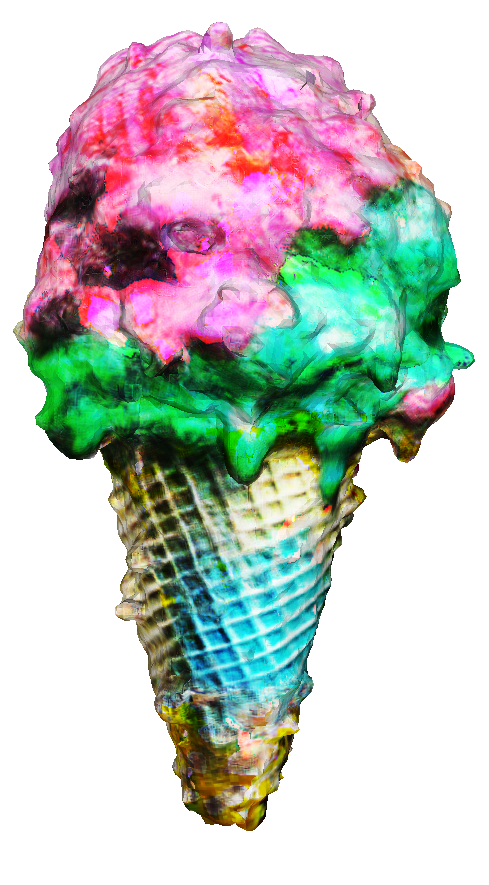}
    \end{tabular}
    \caption{We observe color bleeding in some cases and some high-frequency artifacts.}
    \label{fig:limitations-coloration}
\end{figure}

A notable issue observed is over-colorization and color bleeding with an increased weight of the AES term, as depicted in Figure \ref{fig:limitations-coloration}. However, despite challenges, AES-based guidance proves more stable compared to DDPO, which relies on policy gradients and can be trickier to stabilize.


Another limitation is the lack of a good evaluation strategy. We observe that for many text-to-3D tasks, good evaluation methods are not available due to the lack of ground truth to be compared against. Therefore, we rely on the CLIP score and aesthetic score as a measure for semantic and quality quantification. While aesthetic scoring gives us an idea of how well would the rendered image be perceived by humans in general, the CLIP score talks about semantic relatedness. Despite this, we notice some ambiguity that arises from CLIP scoring. For the views rendered in Figure \ref{fig:ddpo_stage_1_only}, we notice that the DreamGaussian result attains a higher value of CLIP score of $37.55$ while our AES-only based achieves a score of $36.96$, which is difficult to explain despite being clear differences in the bird structures, requiring further investigation. 
Though our method improves upon several other methods, we do not notice any improvement in the Janus problem, which we attribute to the lack of 3D awareness in the scoring mechanism. 

%% file: sec/conclusion.tex
\section{Conclusion}
We currently primarily focus on Score distillation-based techniques. However, we hypothesize that this can further be extended to recent variational score distillation techniques such as ProlificDreamer \cite{wang2023prolificdreamer}, which also rely on Stable Diffusion guidance. Our method demonstrates how policy gradient methods can be extended to score distillation-based techniques, allowing for better metrics and visual quality of results. Since it can be used with non-differentiable rewards, we look forward to the vision and graphics community using it for new reward-based 3D asset generation. We also show how a simple aesthetic score-based optimization can widely enhance the visual properties of the rendered object, not only impacting the textures but guiding changes in the geometry as well. Our aesthetic score-based technique shows improvement with many SDS-based techniques out-of-the-box and can readily be used to steer the generation to a better result.

%% file: sec/X_suppl.tex
\clearpage
\appendix
\section*{Appendix}
\section{Additional Results}
\begin{table}[!htbp]
\centering
\scalebox{0.82}{
\begin{tabular}{ccccc}
\multicolumn{5}{c}{\textit{``A parrot sitting on a basket of macarons."}}  \\
 \includegraphics[width=0.2\textwidth]{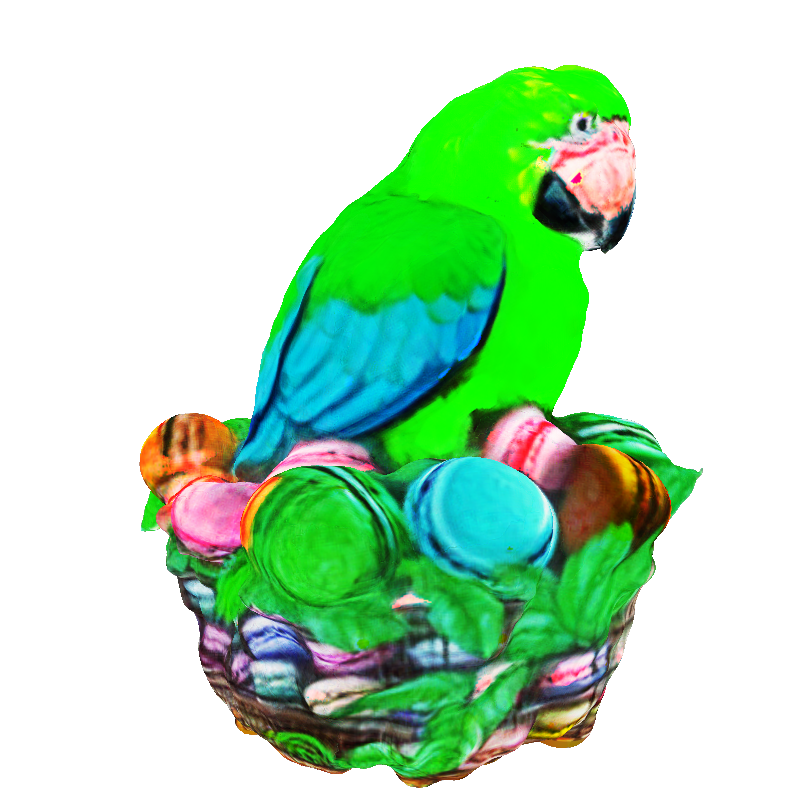} & 
 \includegraphics[width=0.2\textwidth]{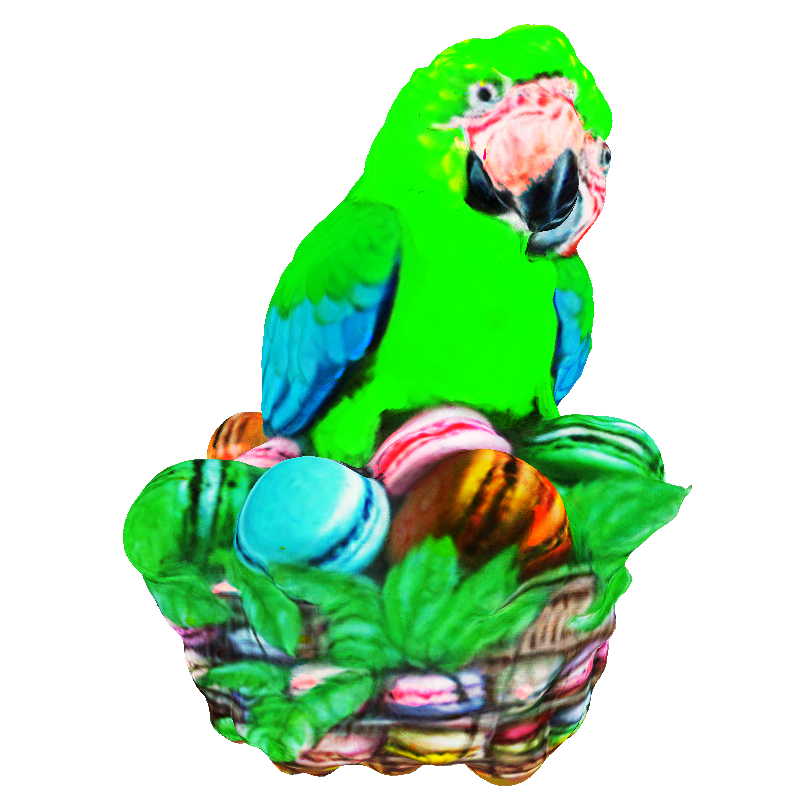} & 
 \includegraphics[width=0.2\textwidth]{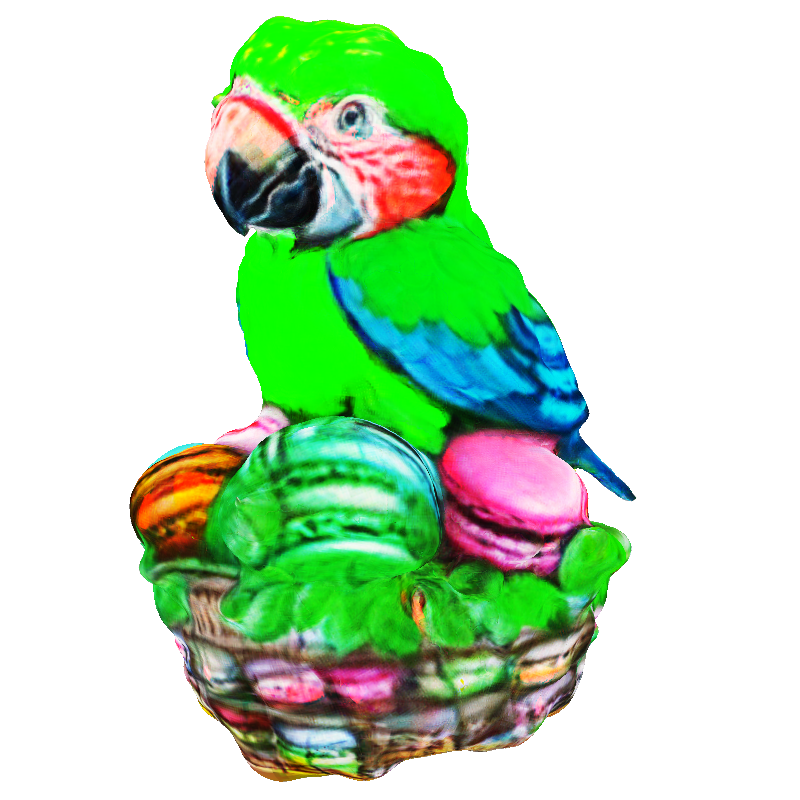} & 
 \includegraphics[width=0.2\textwidth]{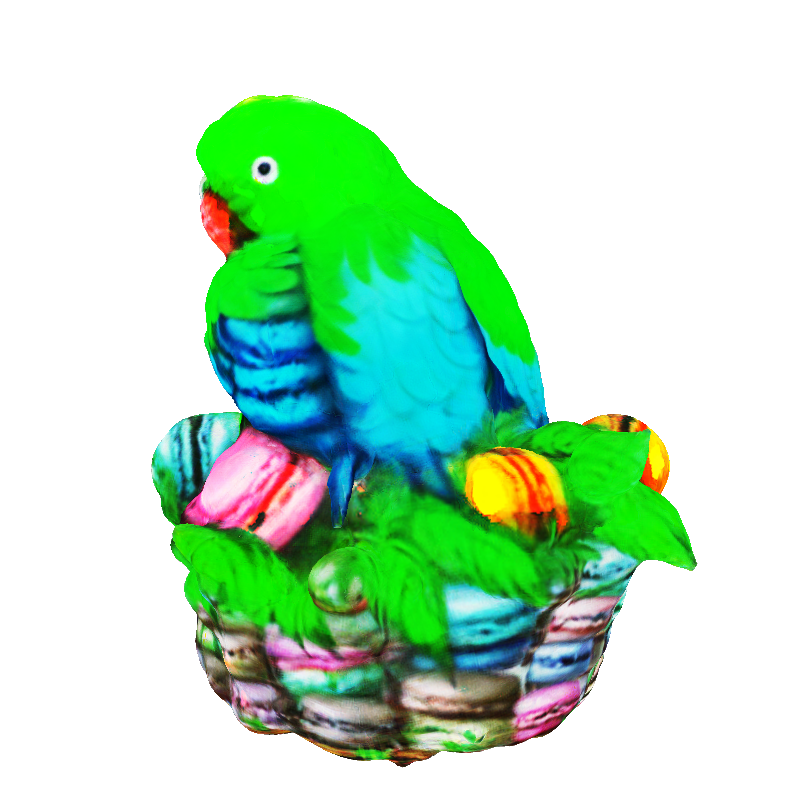} & 
 \includegraphics[width=0.23\textwidth]{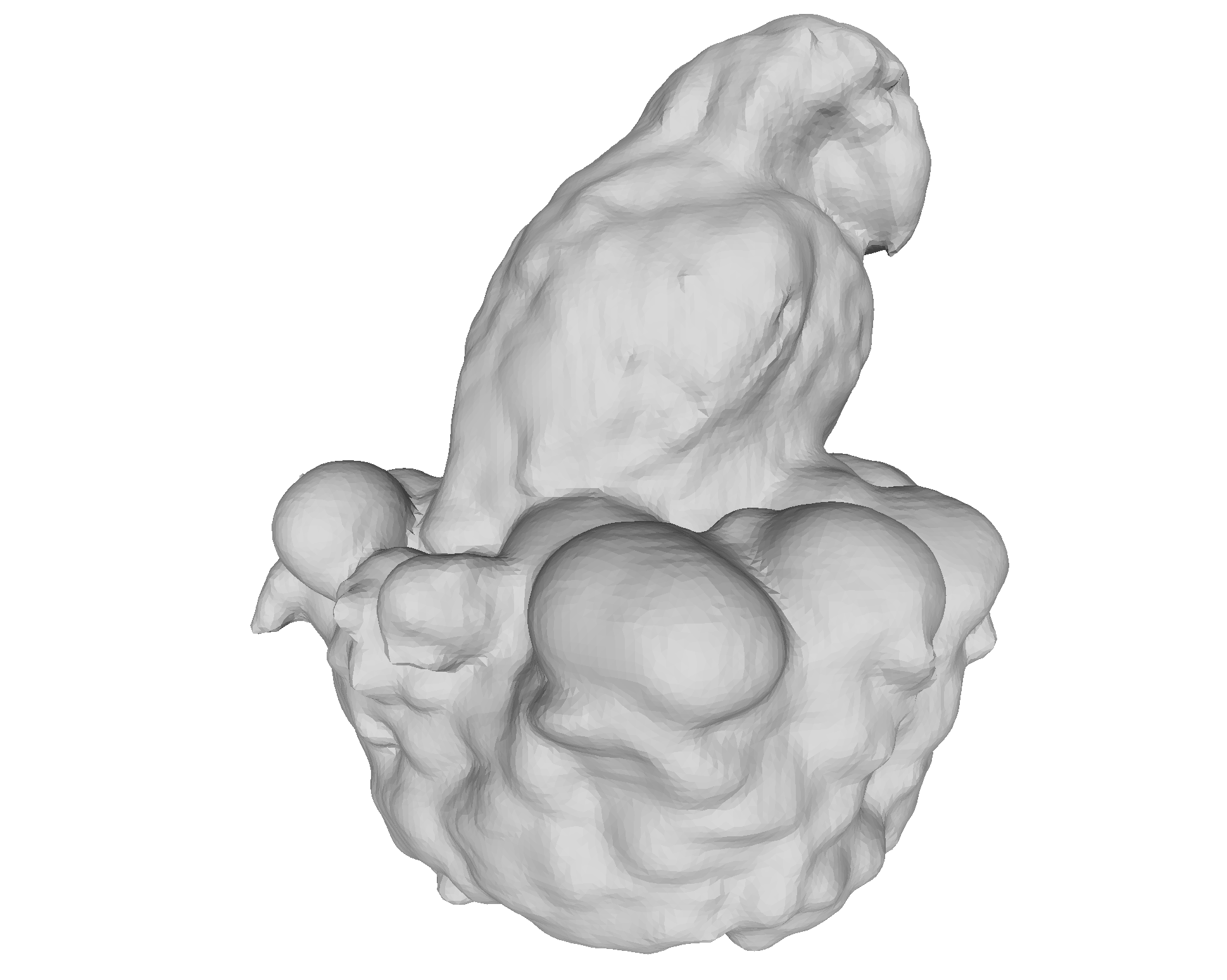} \\
\multicolumn{5}{c}{\textit{``An imperial state crown of England"}}  \\
 \includegraphics[width=0.2\textwidth]{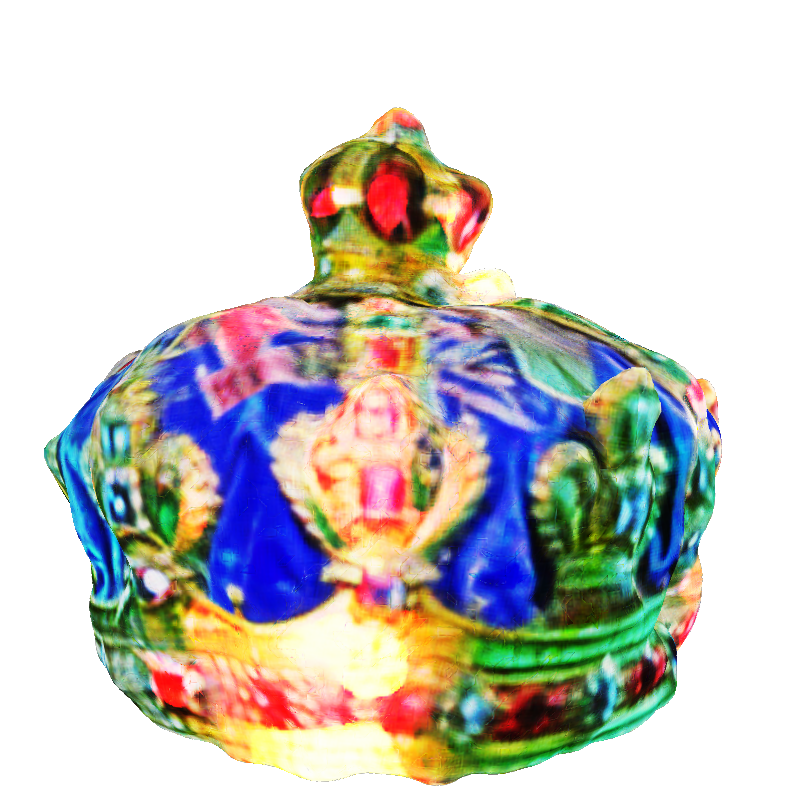} & 
 \includegraphics[width=0.2\textwidth]{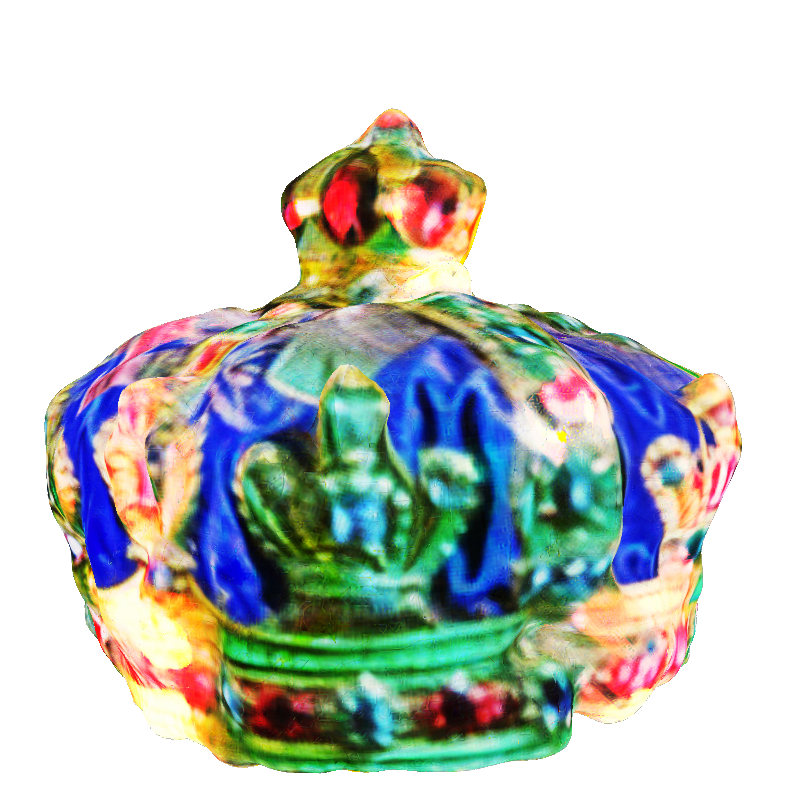} & 
 \includegraphics[width=0.2\textwidth]{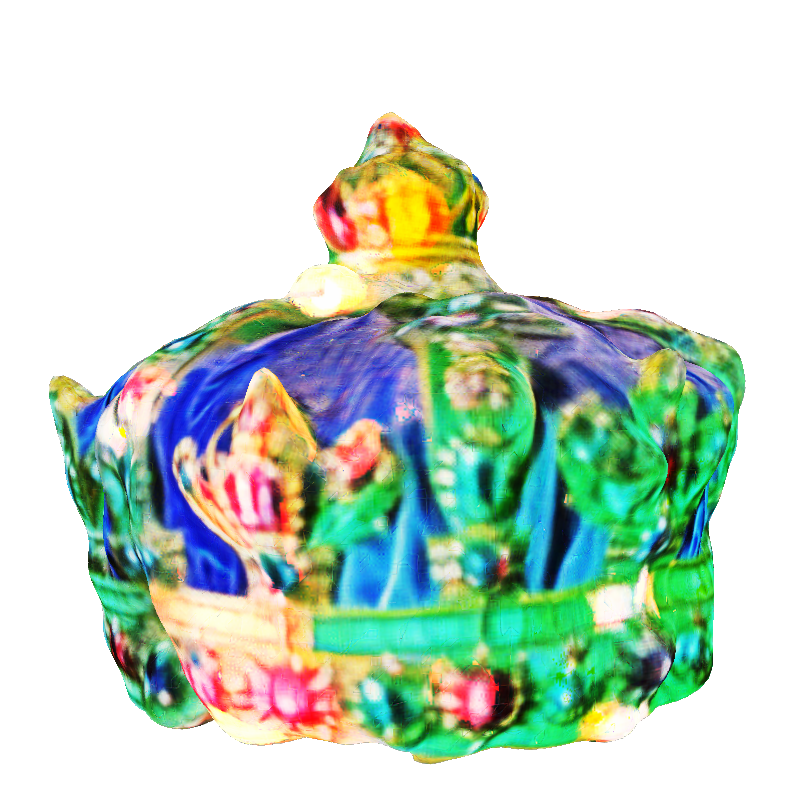} & 
 \includegraphics[width=0.2\textwidth]{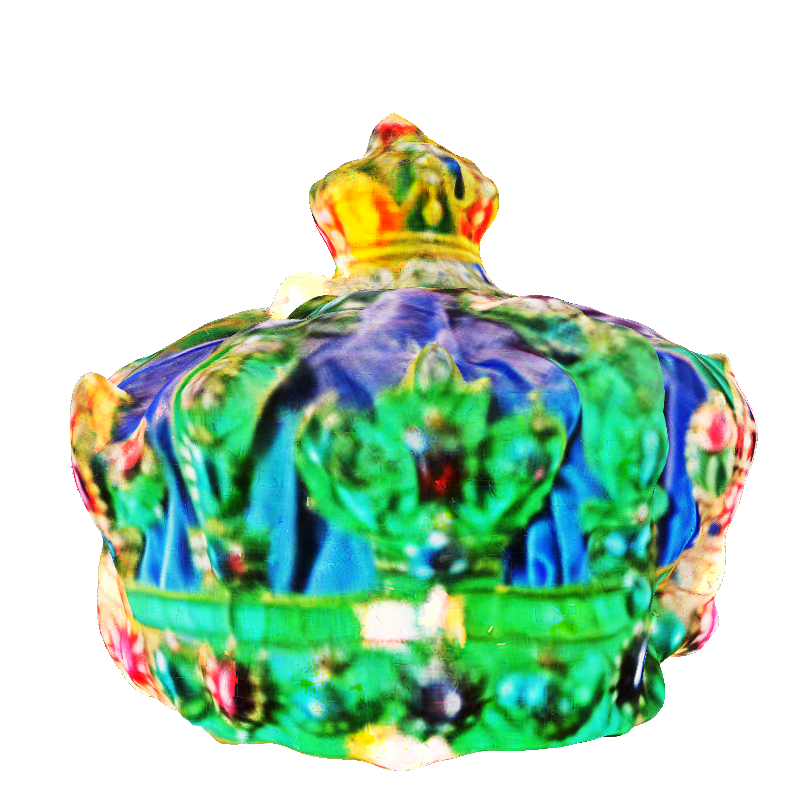} & 
 \includegraphics[width=0.23\textwidth]{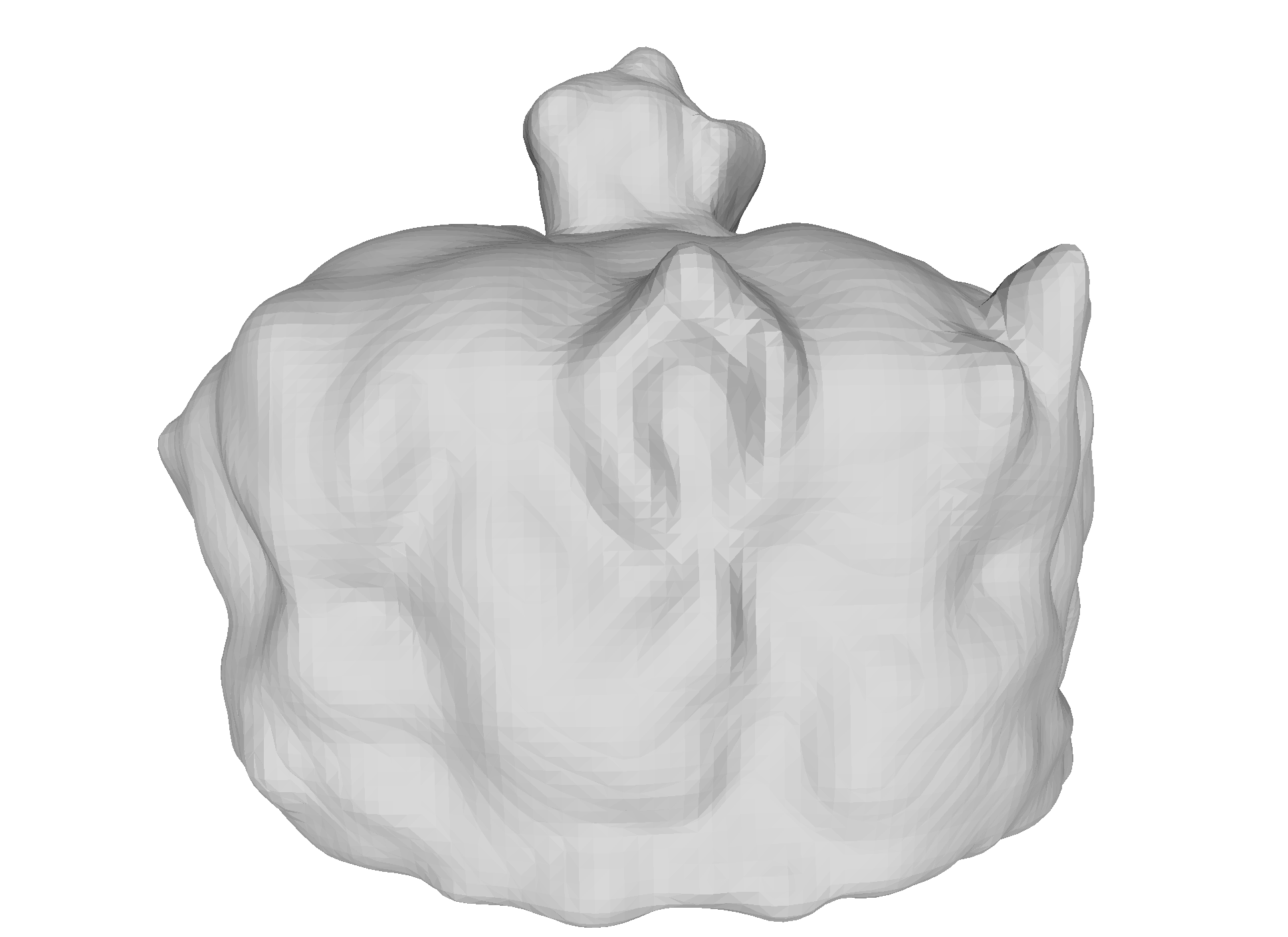} \\
\multicolumn{5}{c}{\textit{``An icecream sundae"}}  \\
 \includegraphics[width=0.2\textwidth]{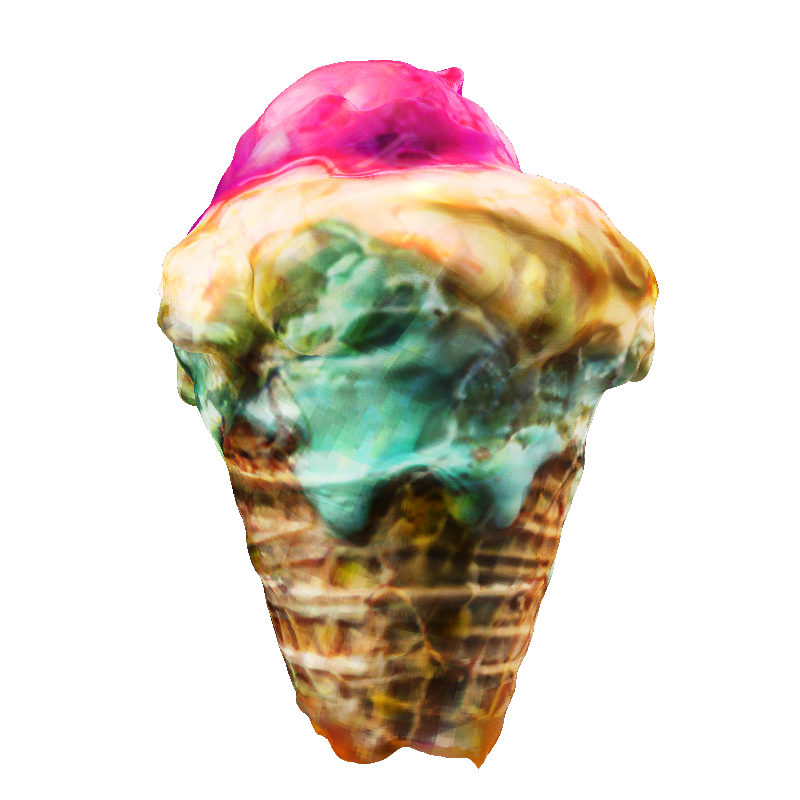} & 
 \includegraphics[width=0.2\textwidth]{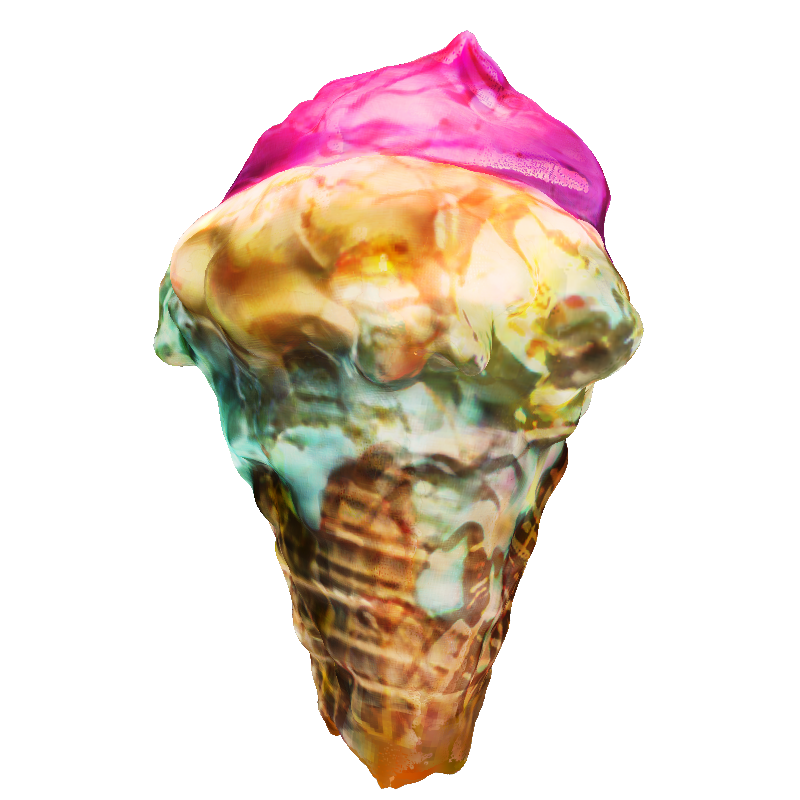} & 
 \includegraphics[width=0.2\textwidth]{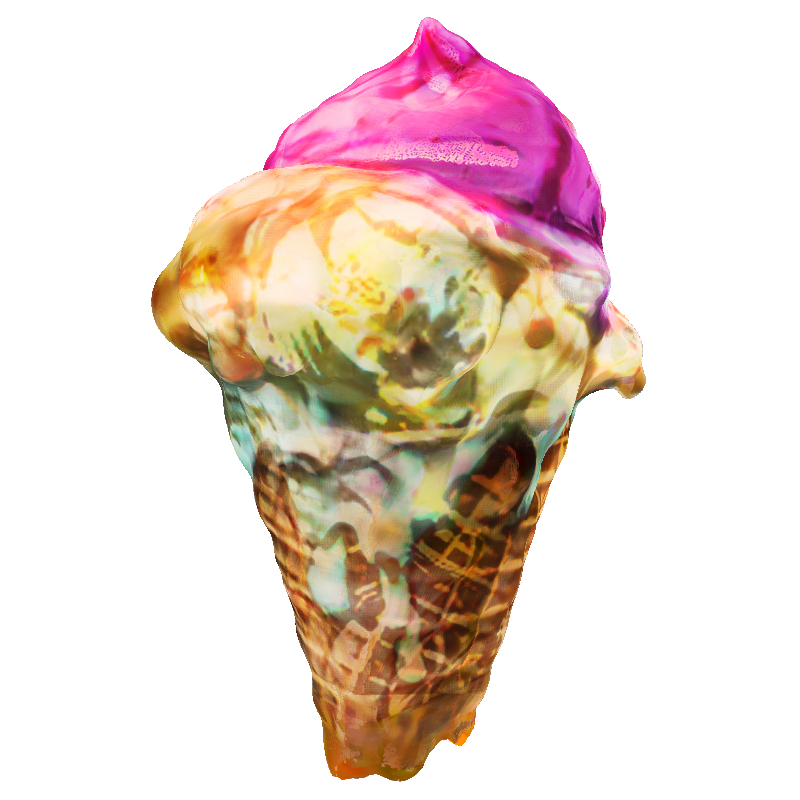} & 
 \includegraphics[width=0.2\textwidth]{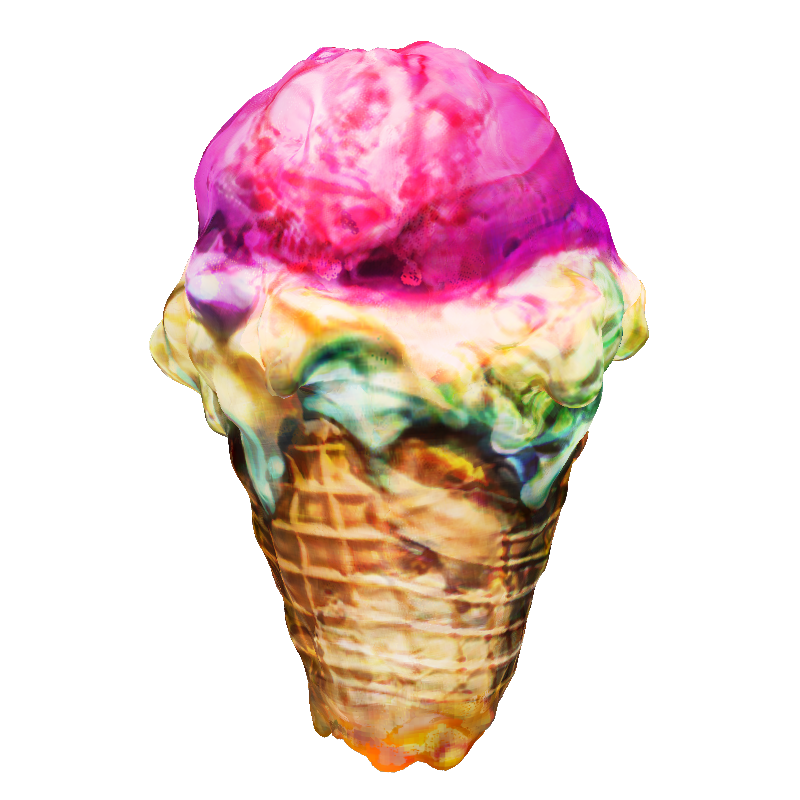} & 
 \includegraphics[width=0.2\textwidth]{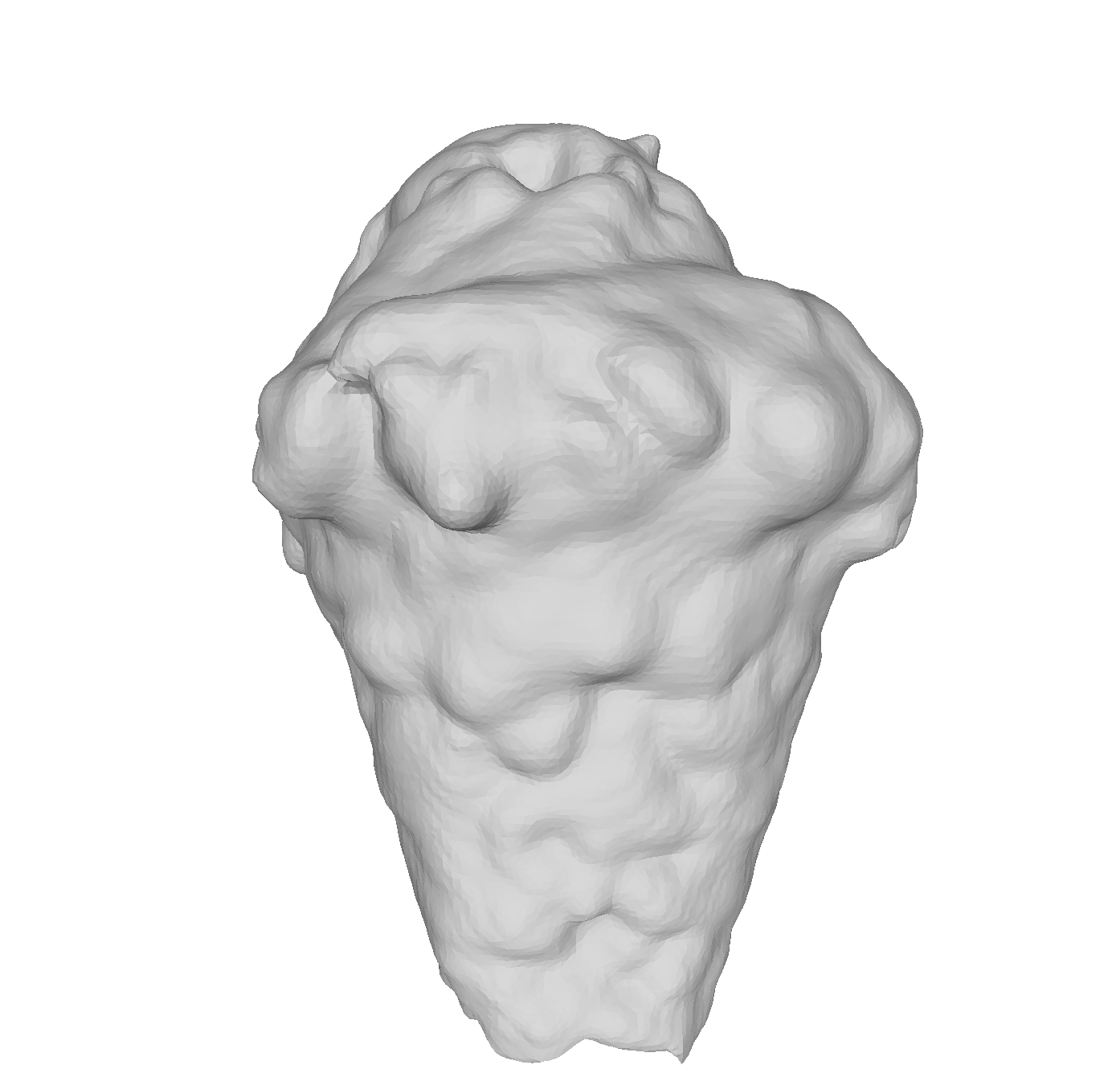} \\
\multicolumn{5}{c}{\textit{``A photo of a fire hydrant, highly detailed"}}  \\
 \includegraphics[width=0.2\textwidth]{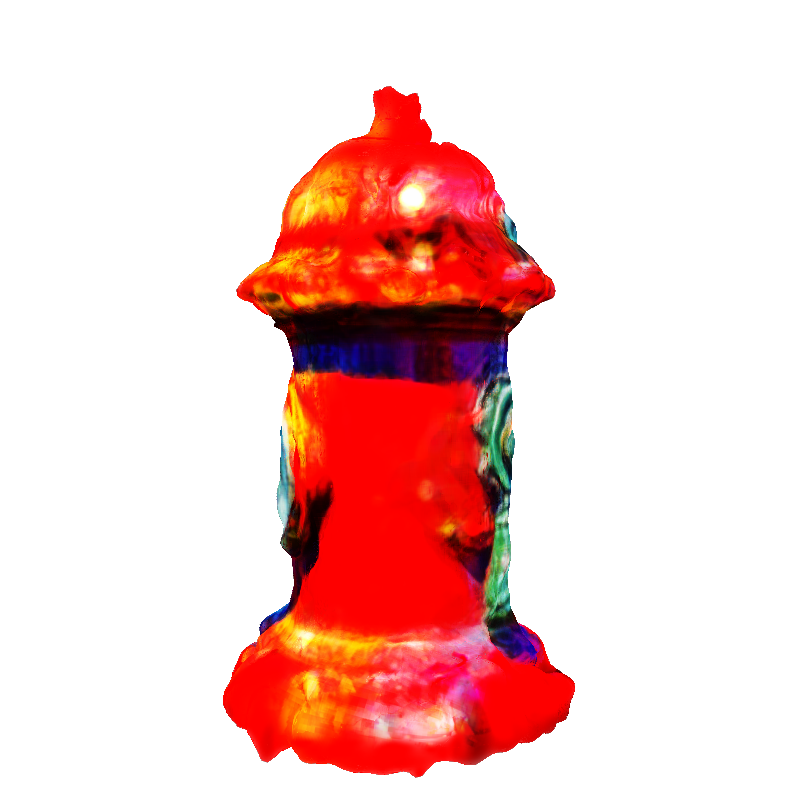} & 
 \includegraphics[width=0.2\textwidth]{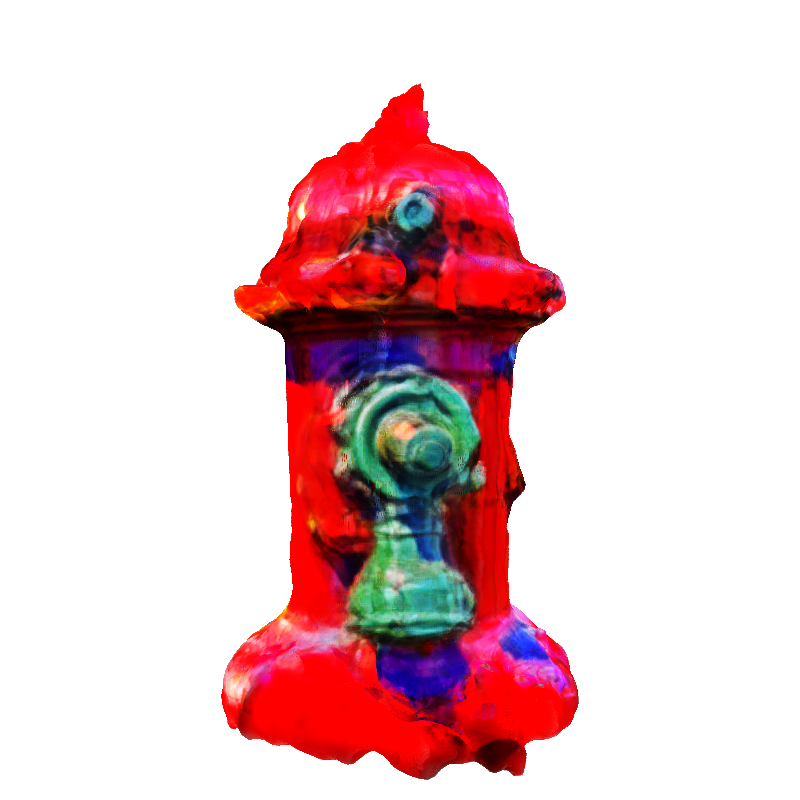} & 
 \includegraphics[width=0.2\textwidth]{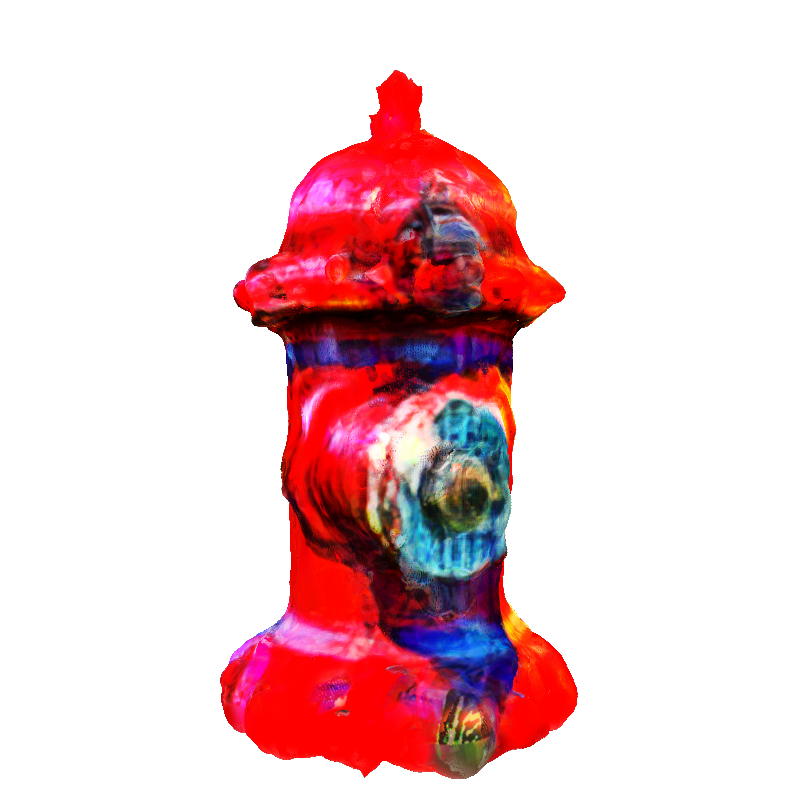} & 
 \includegraphics[width=0.2\textwidth]{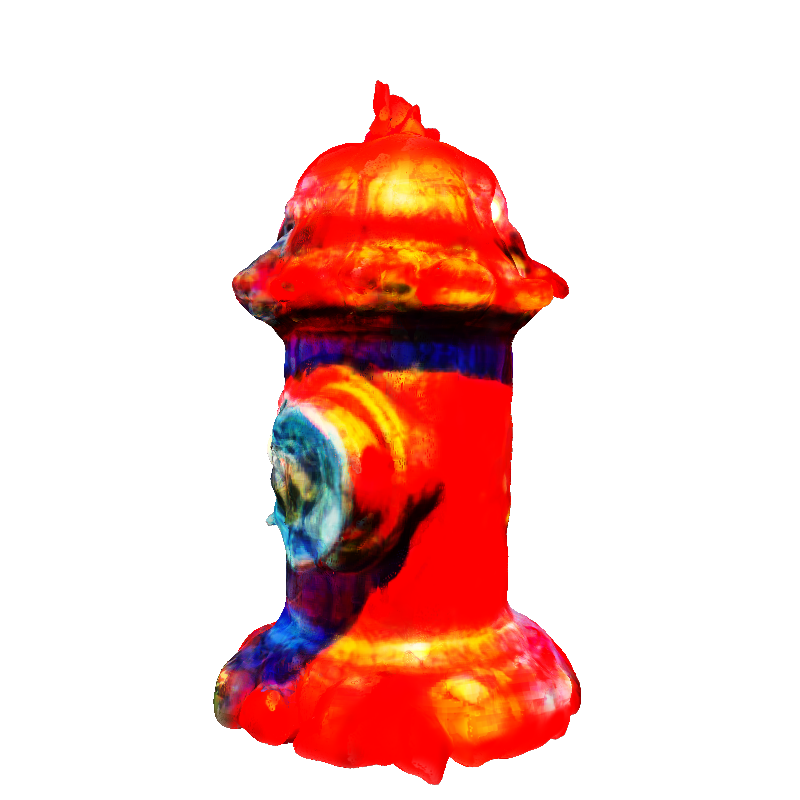} & 
 \includegraphics[width=0.22\textwidth]{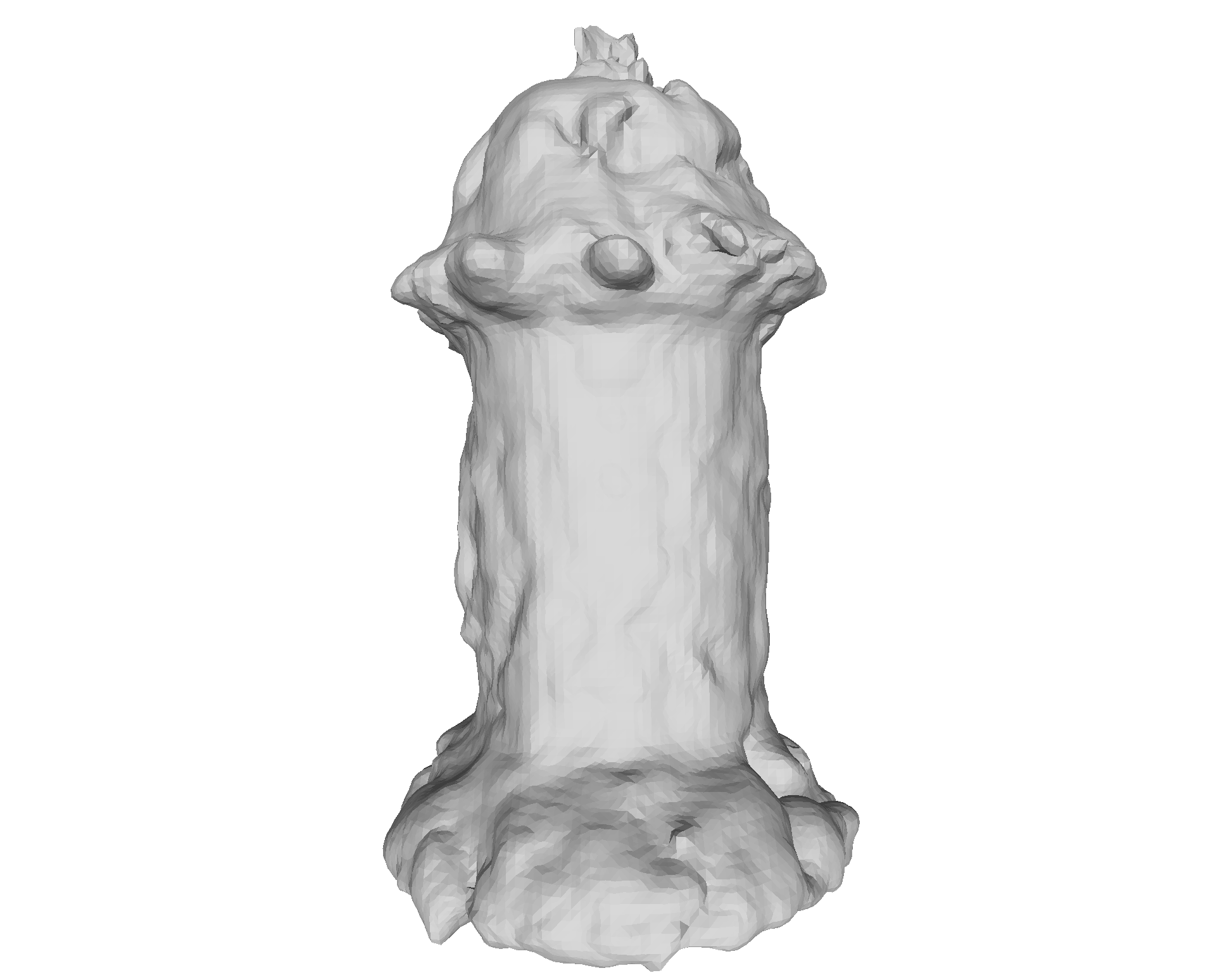} \\
\multicolumn{5}{c}{\textit{``A bagel filled with cream cheese and lox."}}  \\
 \includegraphics[width=0.2\textwidth]{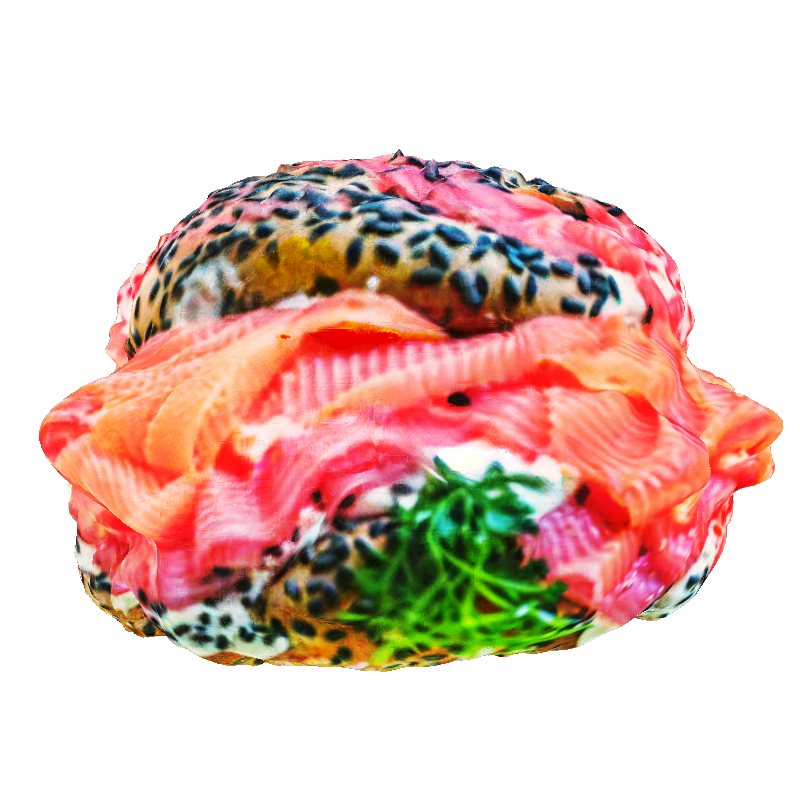} & 
 \includegraphics[width=0.2\textwidth]{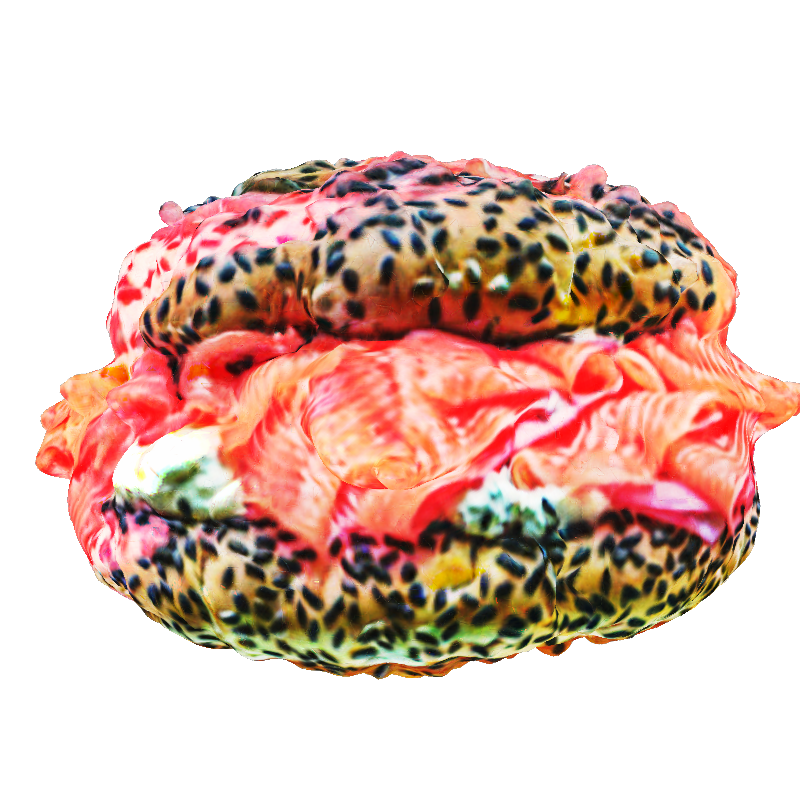} & 
 \includegraphics[width=0.2\textwidth]{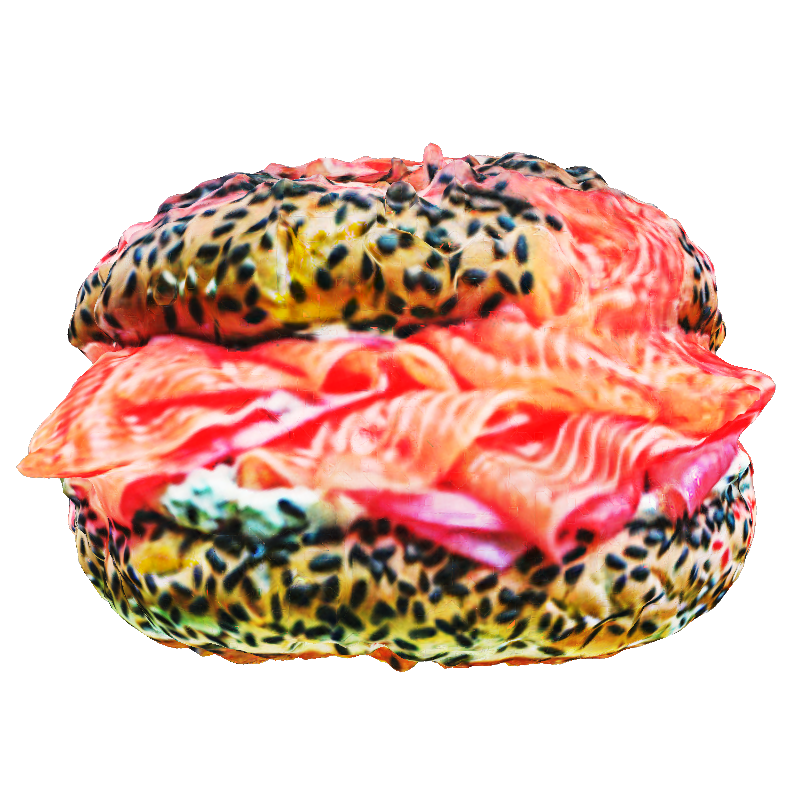} & 
 \includegraphics[width=0.2\textwidth]{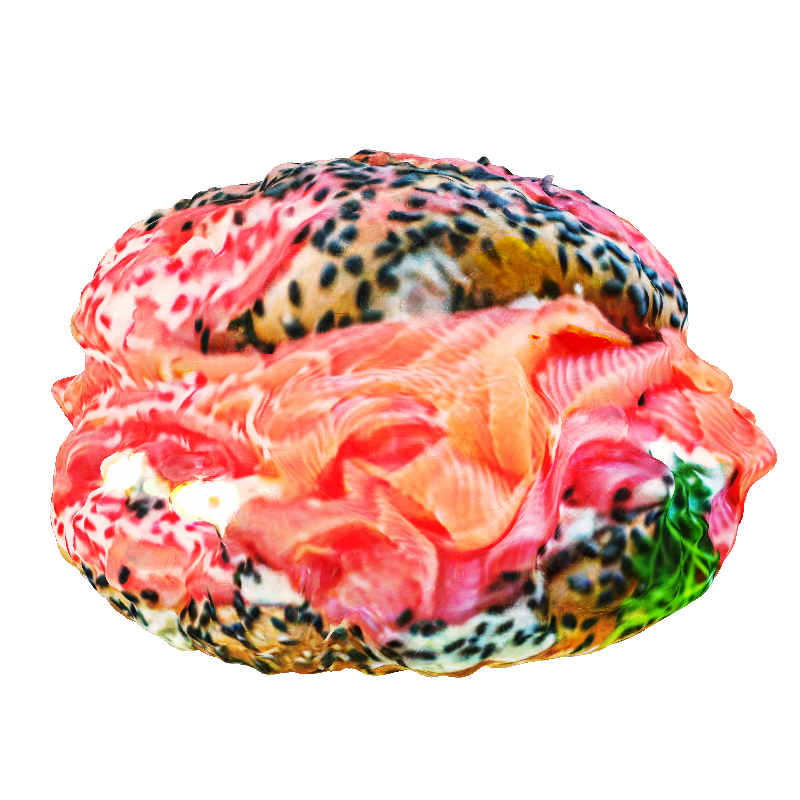} & 
 \includegraphics[width=0.22\textwidth]{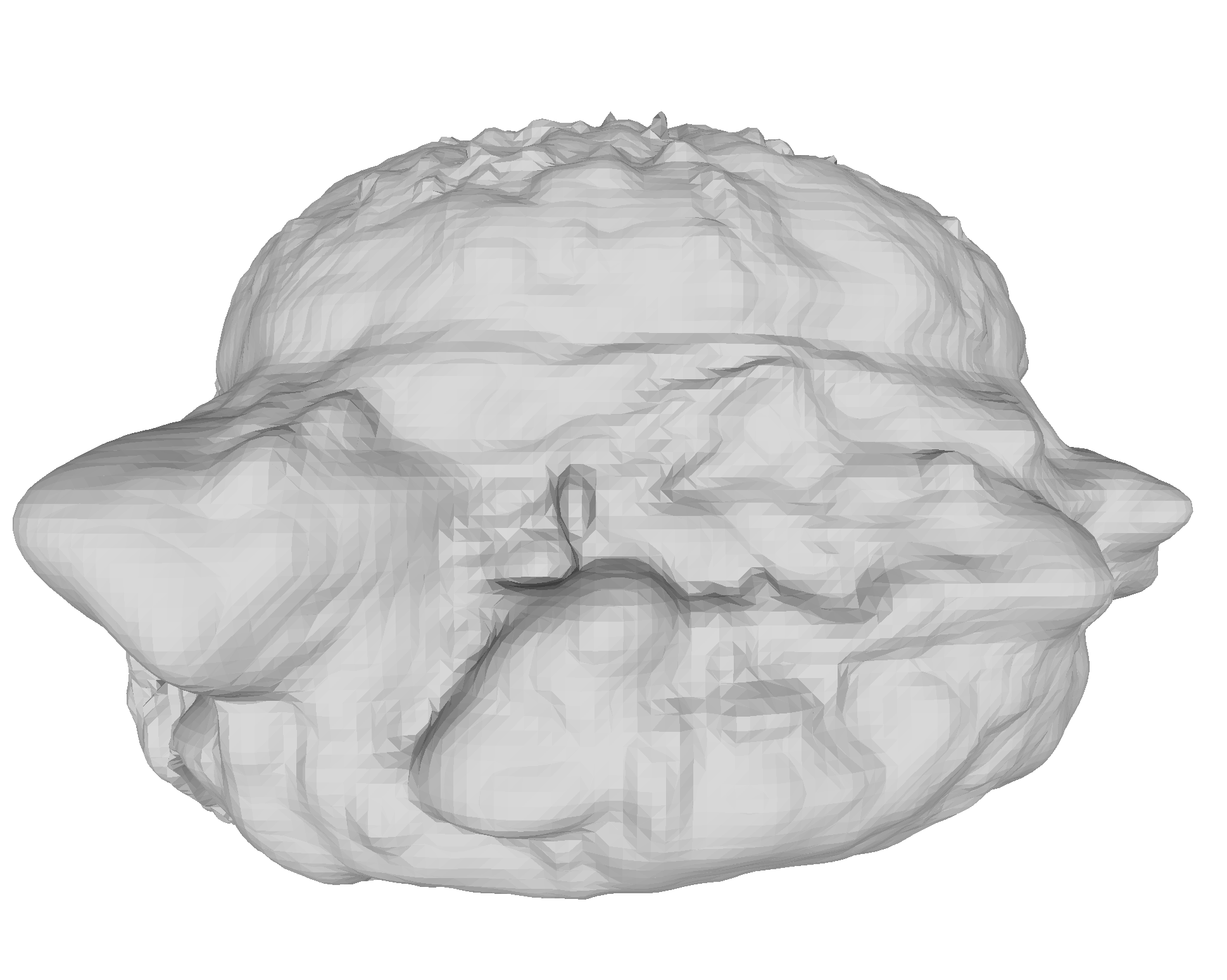} \\
\multicolumn{5}{c}{\textit{``A fresh cinnamon roll covered in glaze, high resolution."}}  \\
 \includegraphics[width=0.2\textwidth]{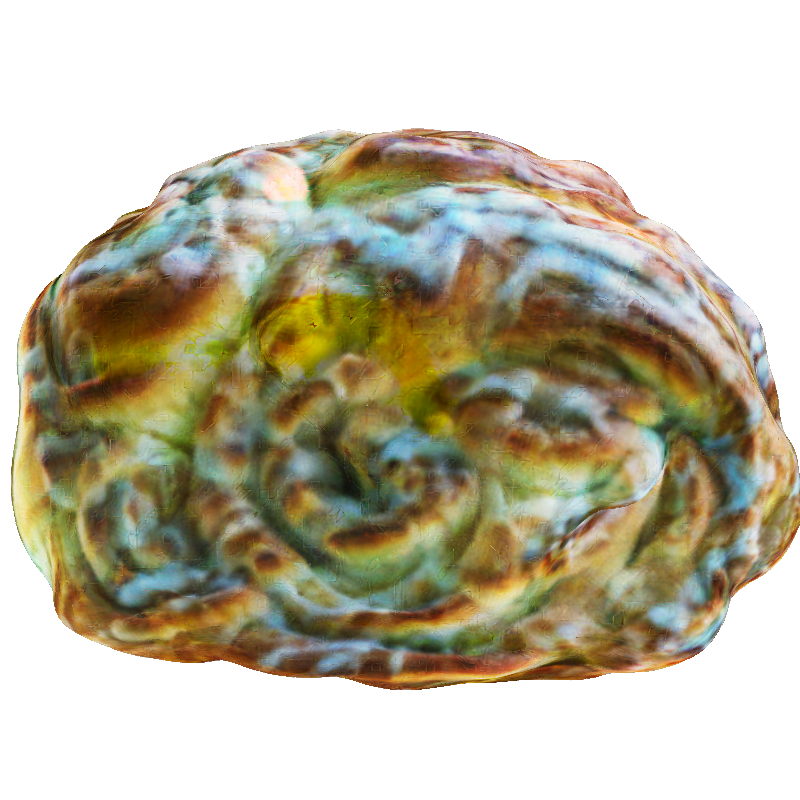} & 
 \includegraphics[width=0.2\textwidth]{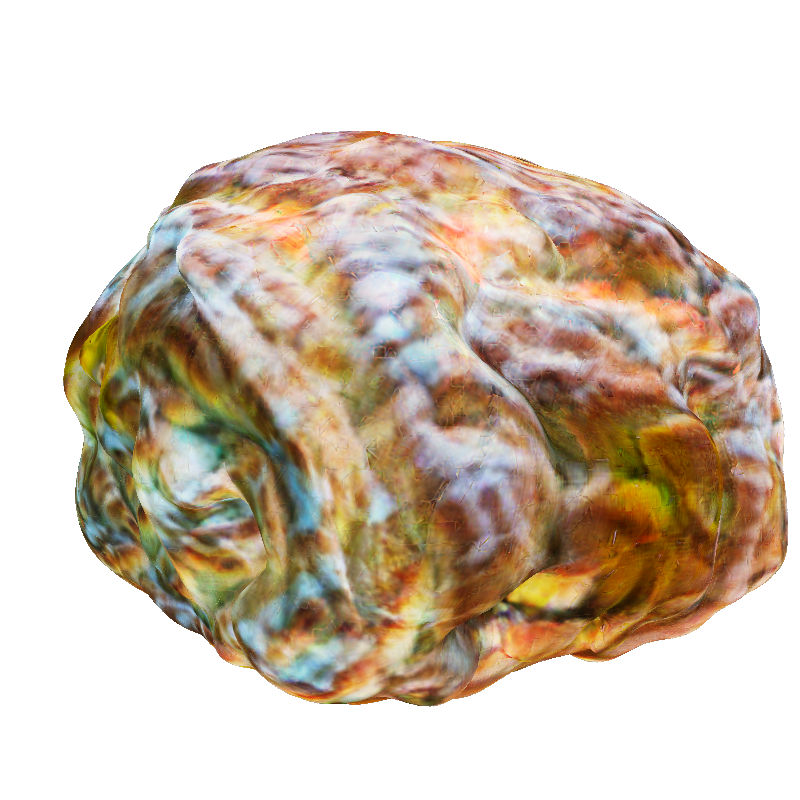} & 
 \includegraphics[width=0.2\textwidth]{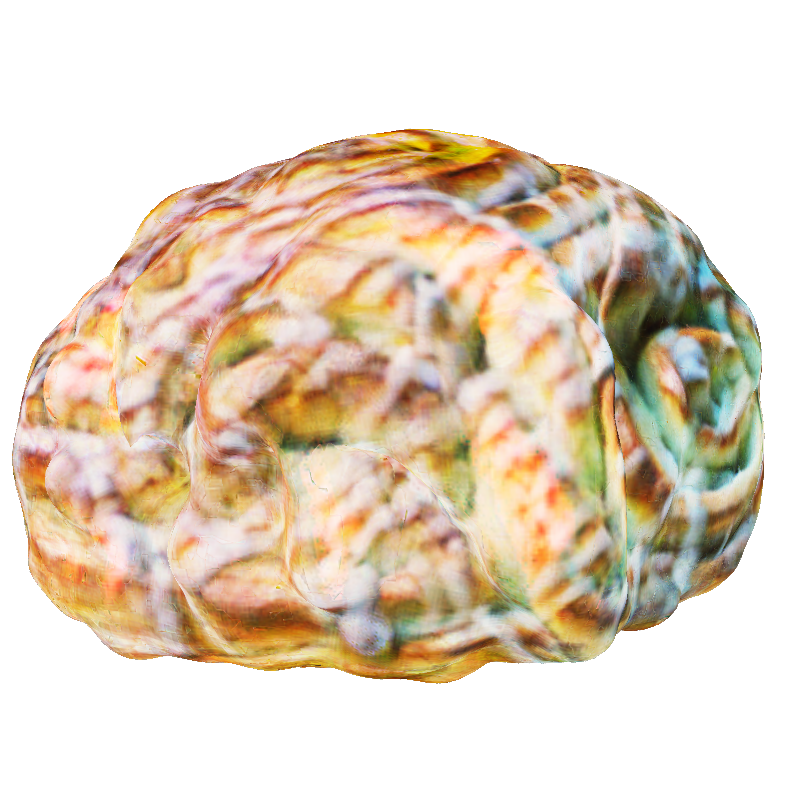} & 
 \includegraphics[width=0.2\textwidth]{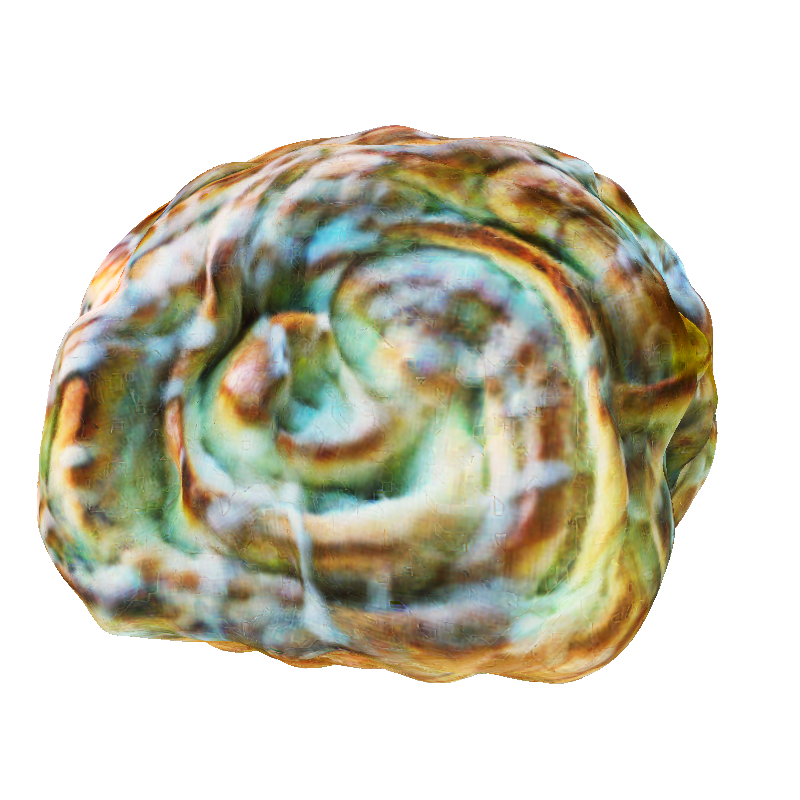} & 
 \includegraphics[width=0.22\textwidth]{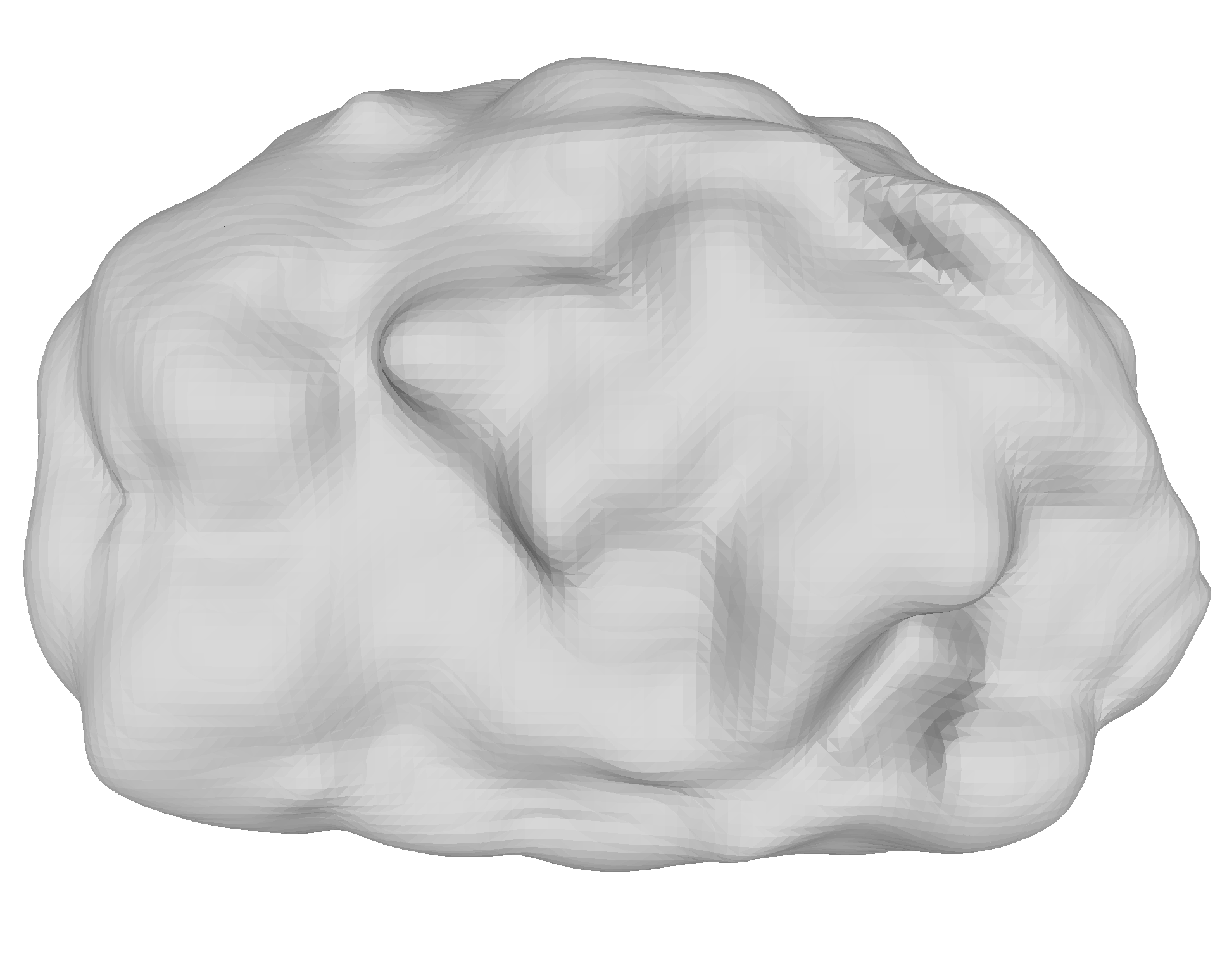} \\

\end{tabular}}
\end{table}

\clearpage
\begin{table*}[!htbp]
\centering
\scalebox{0.85}{
\begin{tabular}{ccccc}
\multicolumn{5}{c}{\textit{``A delicious hamburger."}}  \\
 \includegraphics[width=0.2\textwidth]{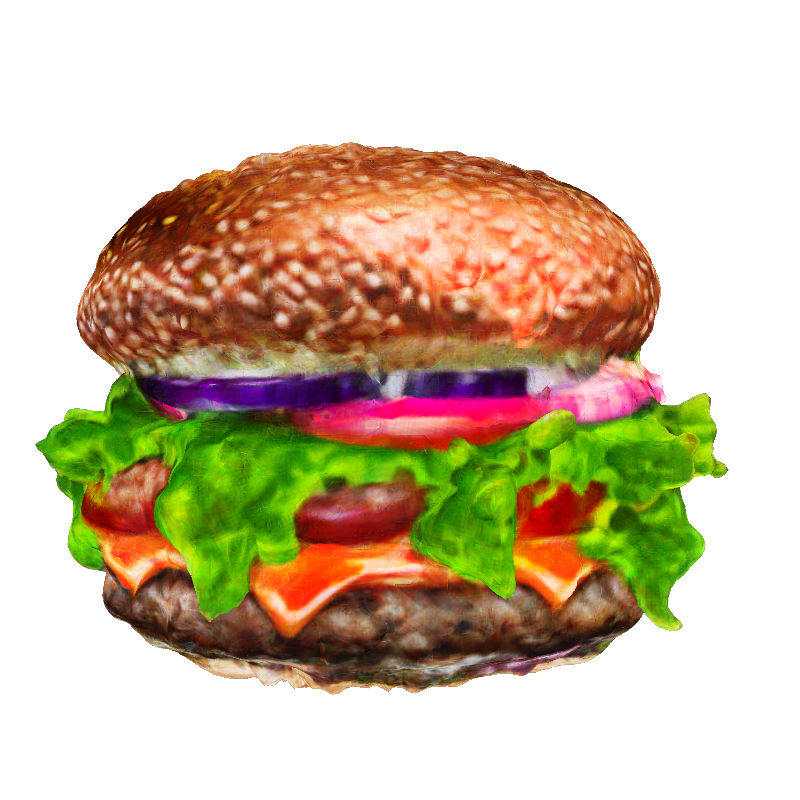} & 
 \includegraphics[width=0.2\textwidth]{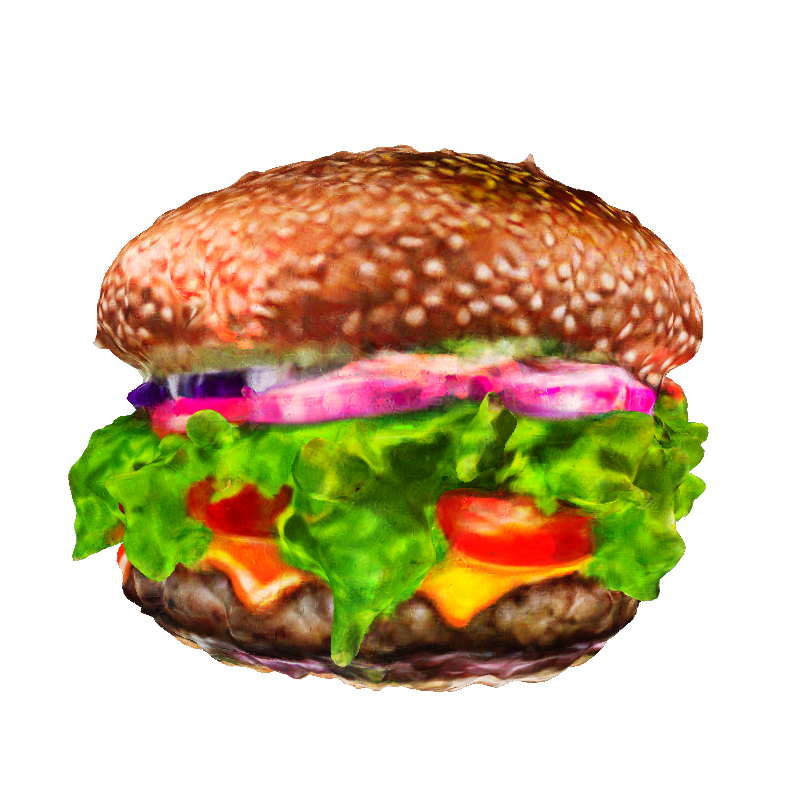} & 
 \includegraphics[width=0.2\textwidth]{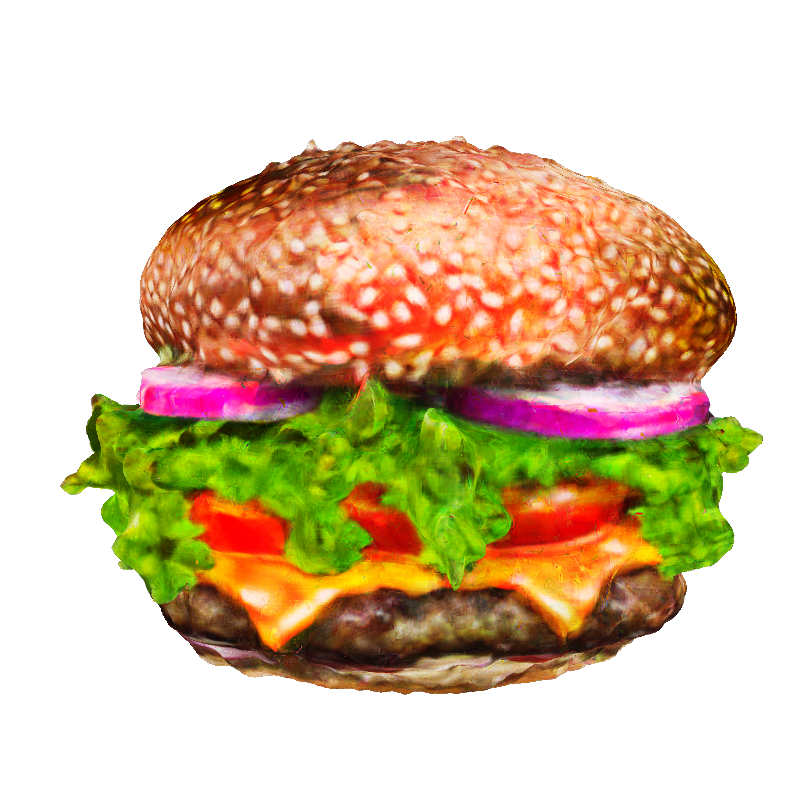} & 
 \includegraphics[width=0.2\textwidth]{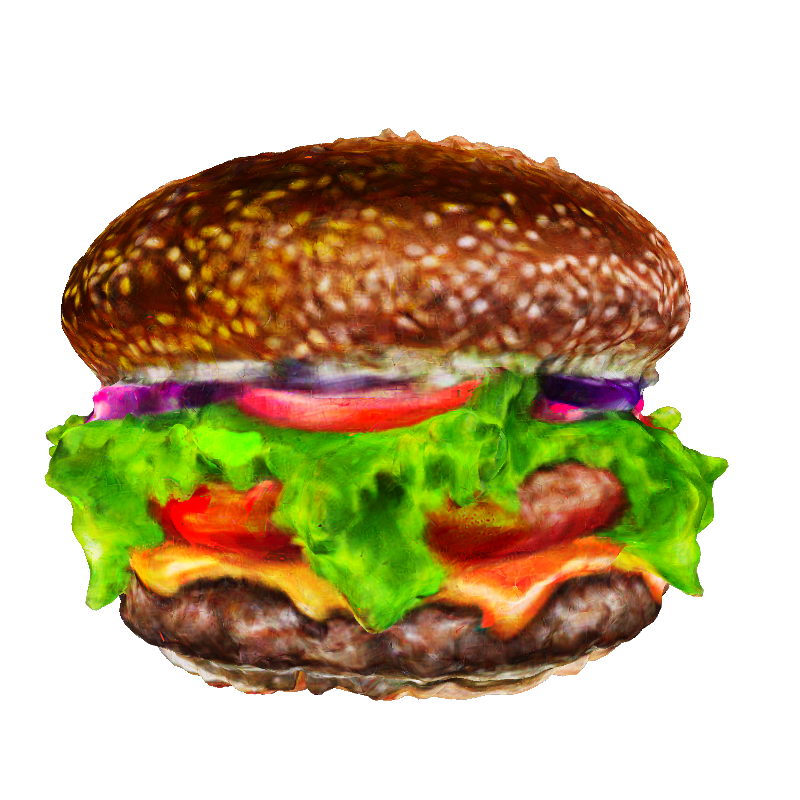} & 
 \includegraphics[width=0.22\textwidth]{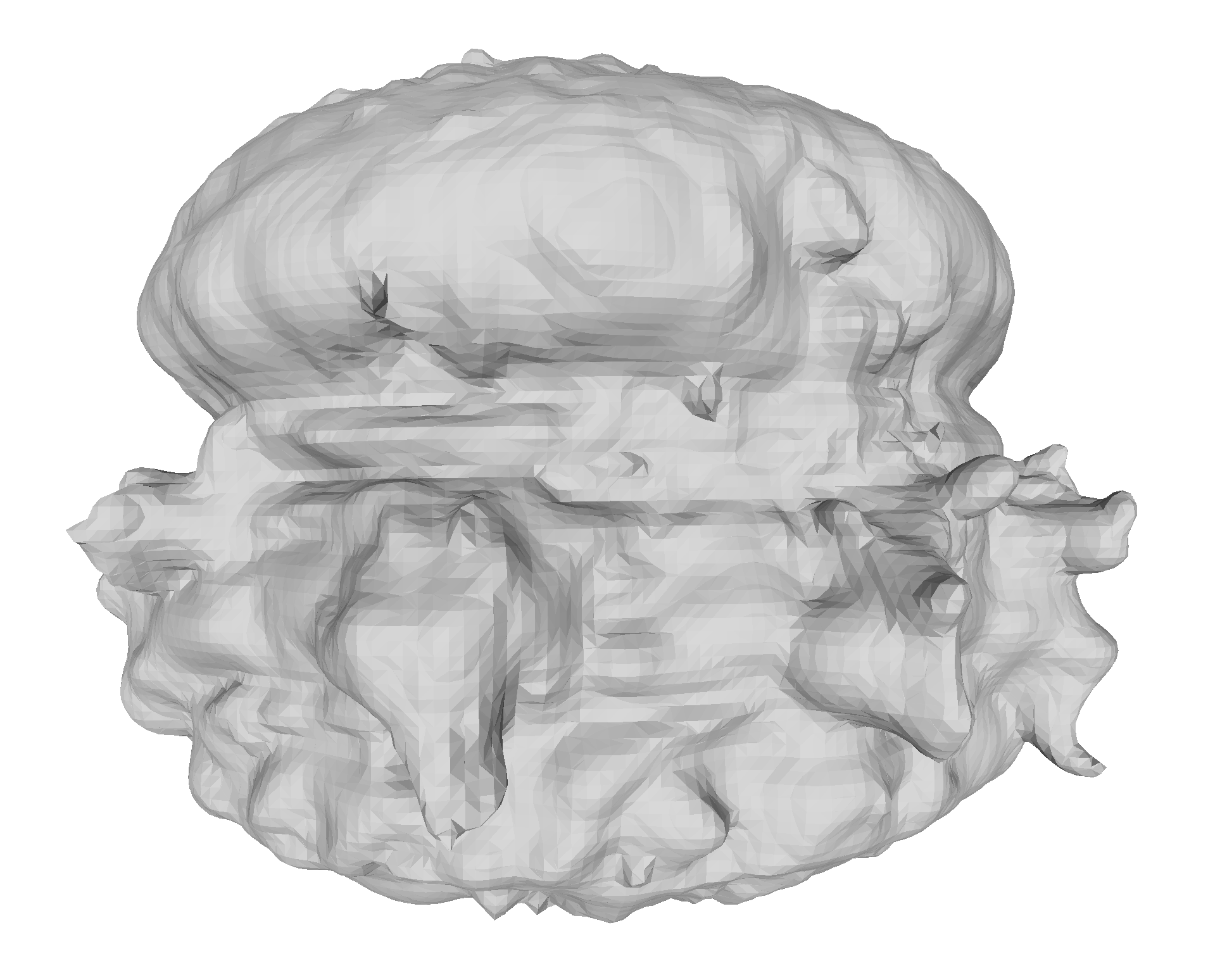} \\
\multicolumn{5}{c}{\textit{``A ripe strawberry."}}  \\
 \includegraphics[width=0.2\textwidth]{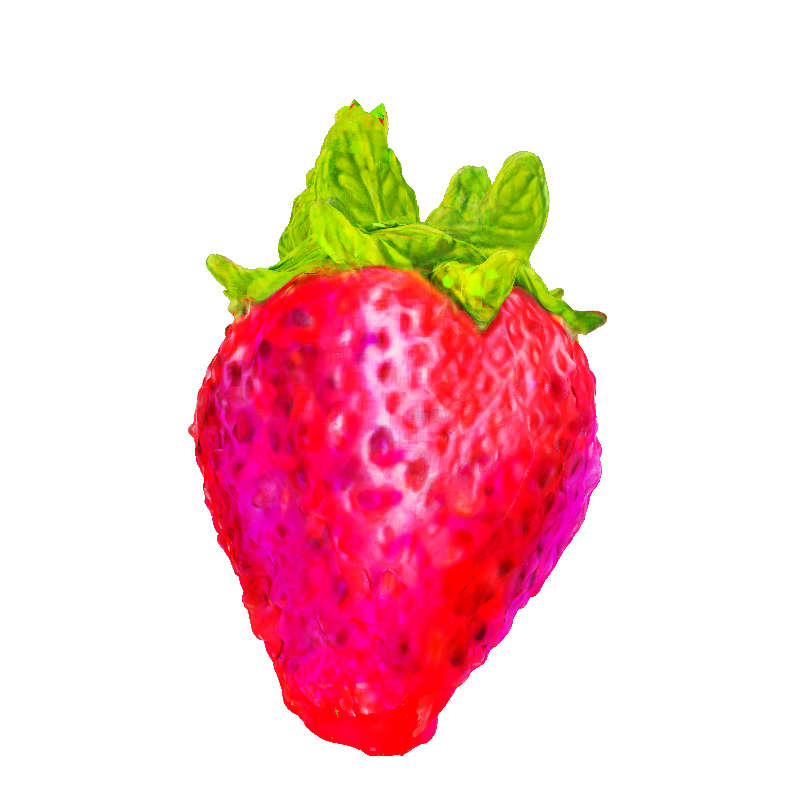} & 
 \includegraphics[width=0.2\textwidth]{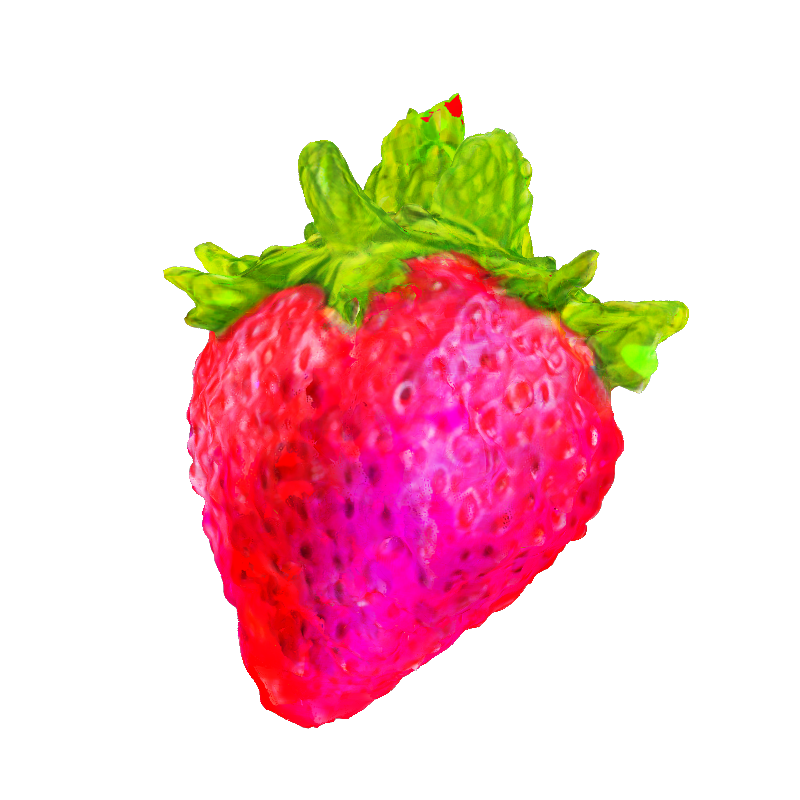} & 
 \includegraphics[width=0.2\textwidth]{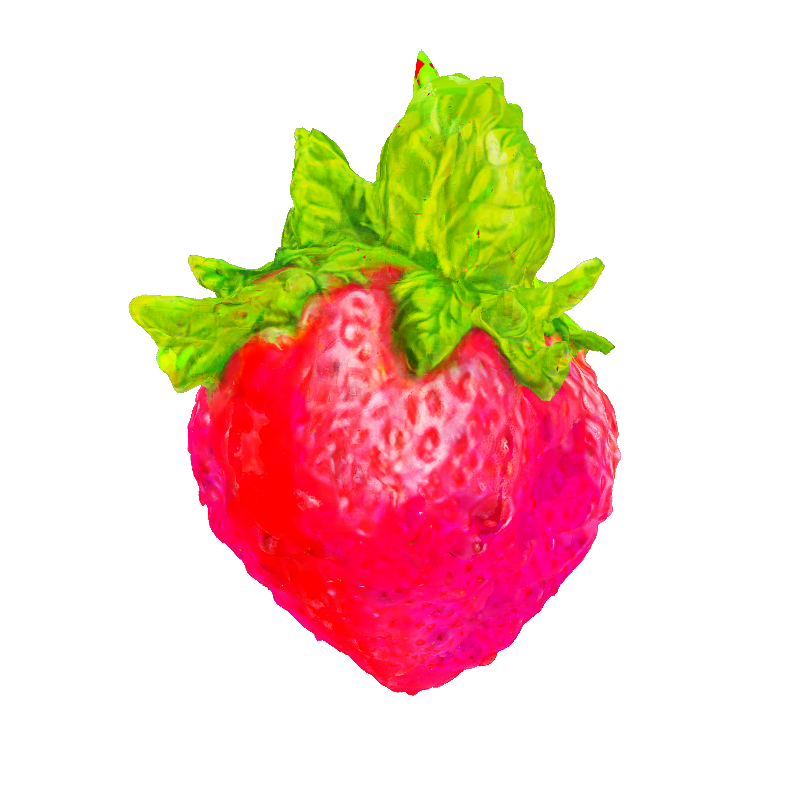} & 
 \includegraphics[width=0.2\textwidth]{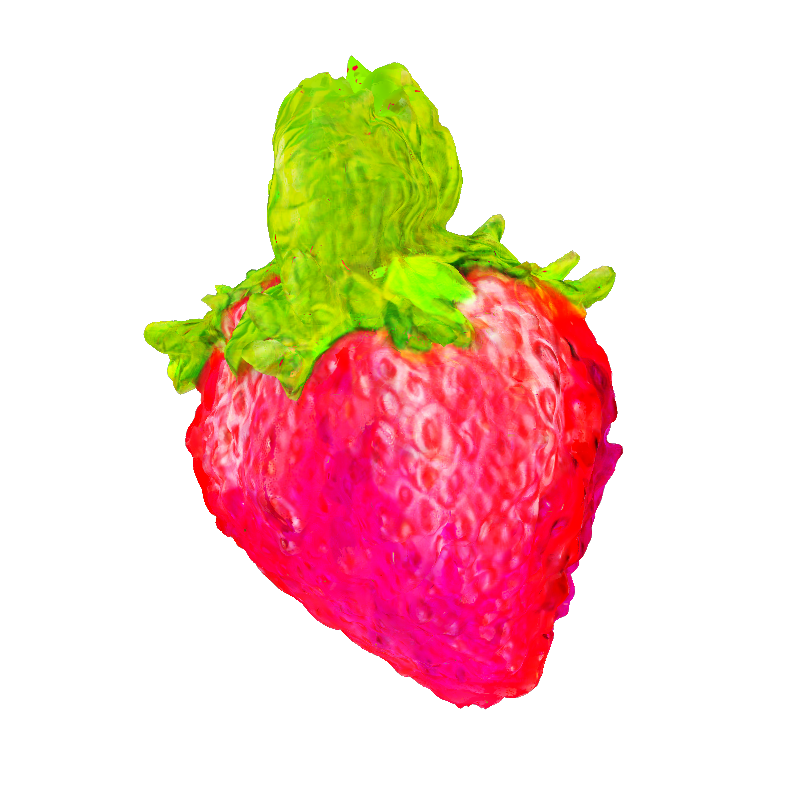} & 
 \includegraphics[width=0.22\textwidth]{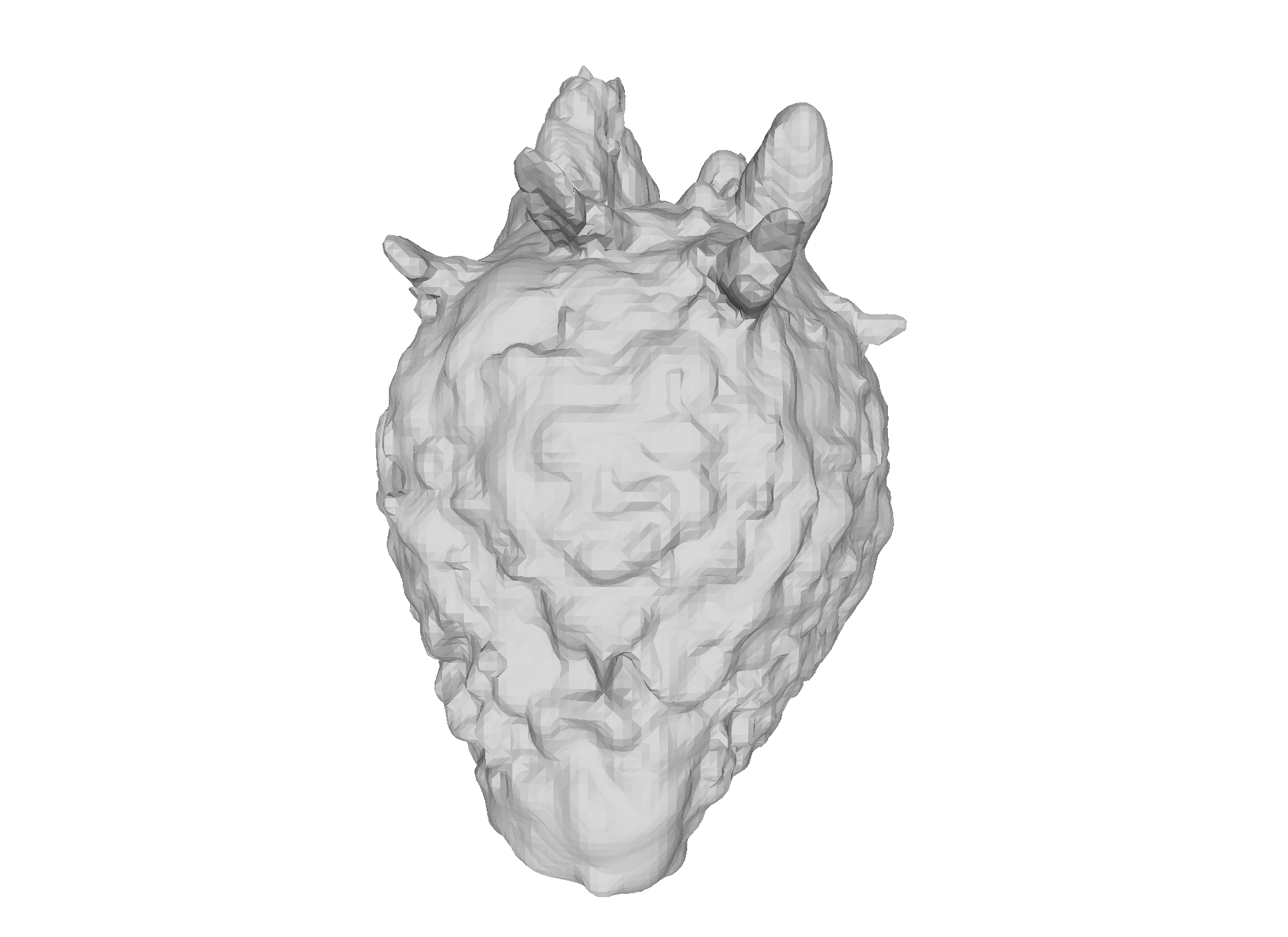} \\
\multicolumn{5}{c}{\textit{``A photo of fries and a hamburger."}}  \\
 \includegraphics[width=0.2\textwidth]{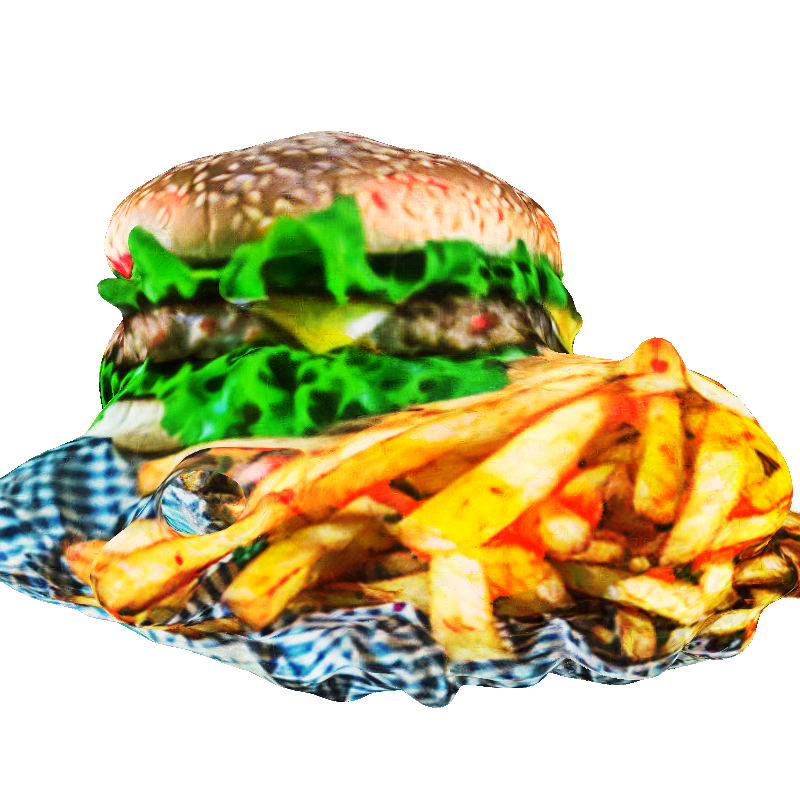} & 
 \includegraphics[width=0.2\textwidth]{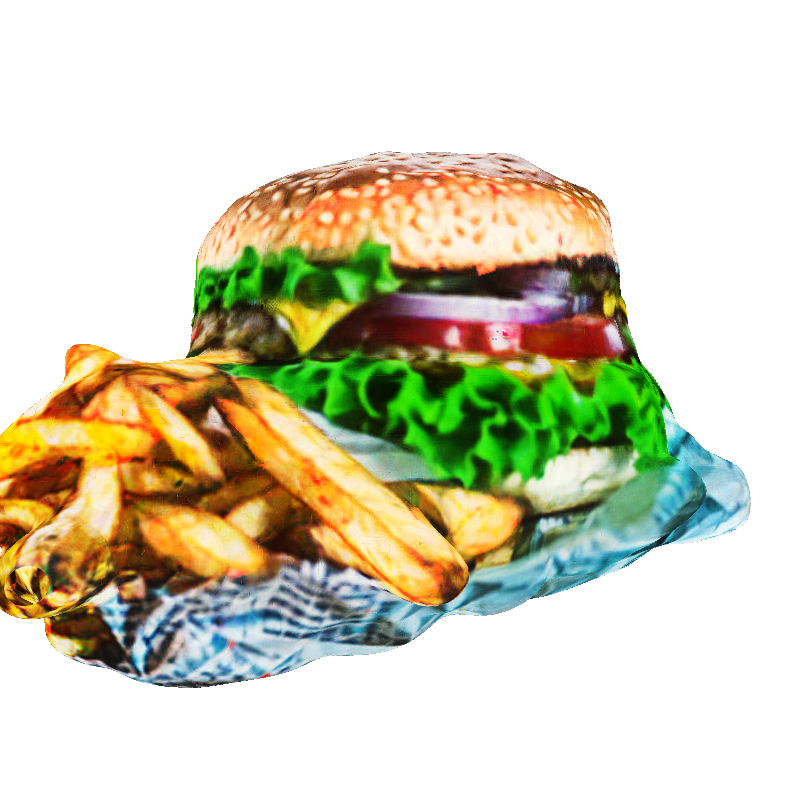} & 
 \includegraphics[width=0.2\textwidth]{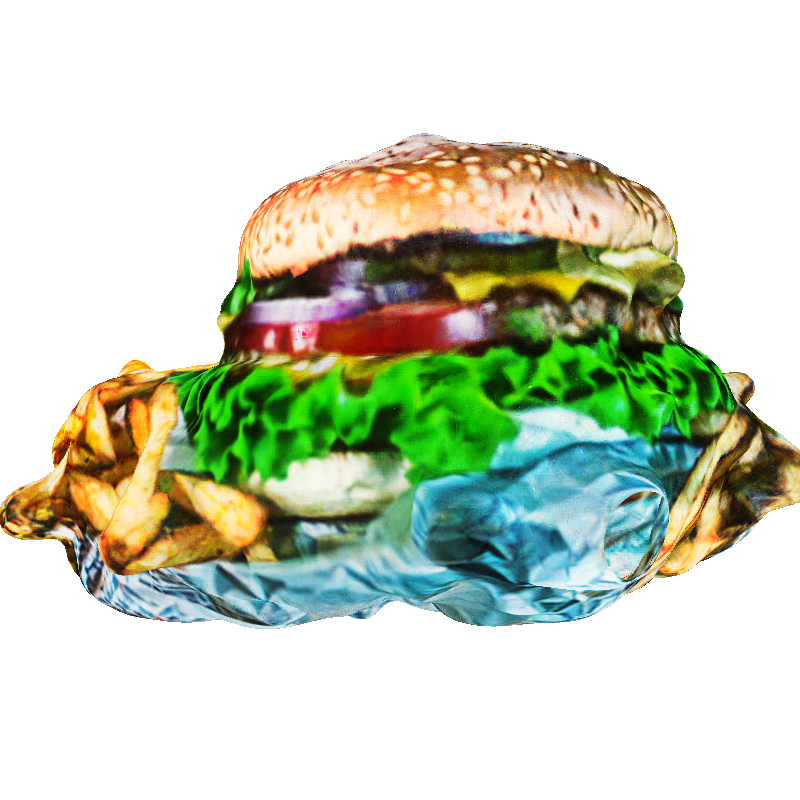} & 
 \includegraphics[width=0.2\textwidth]{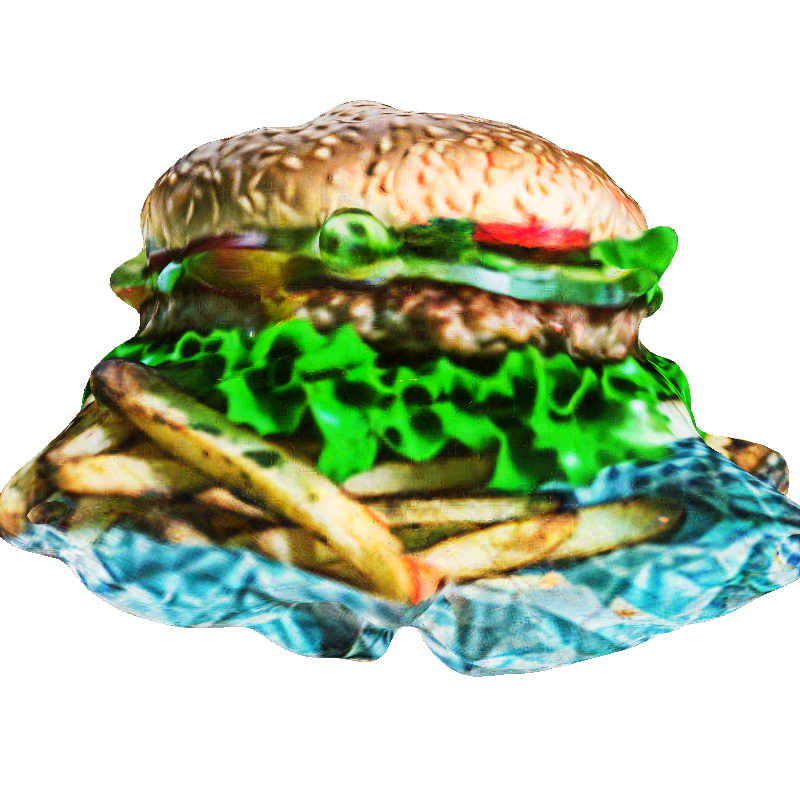} & 
 \includegraphics[width=0.22\textwidth]{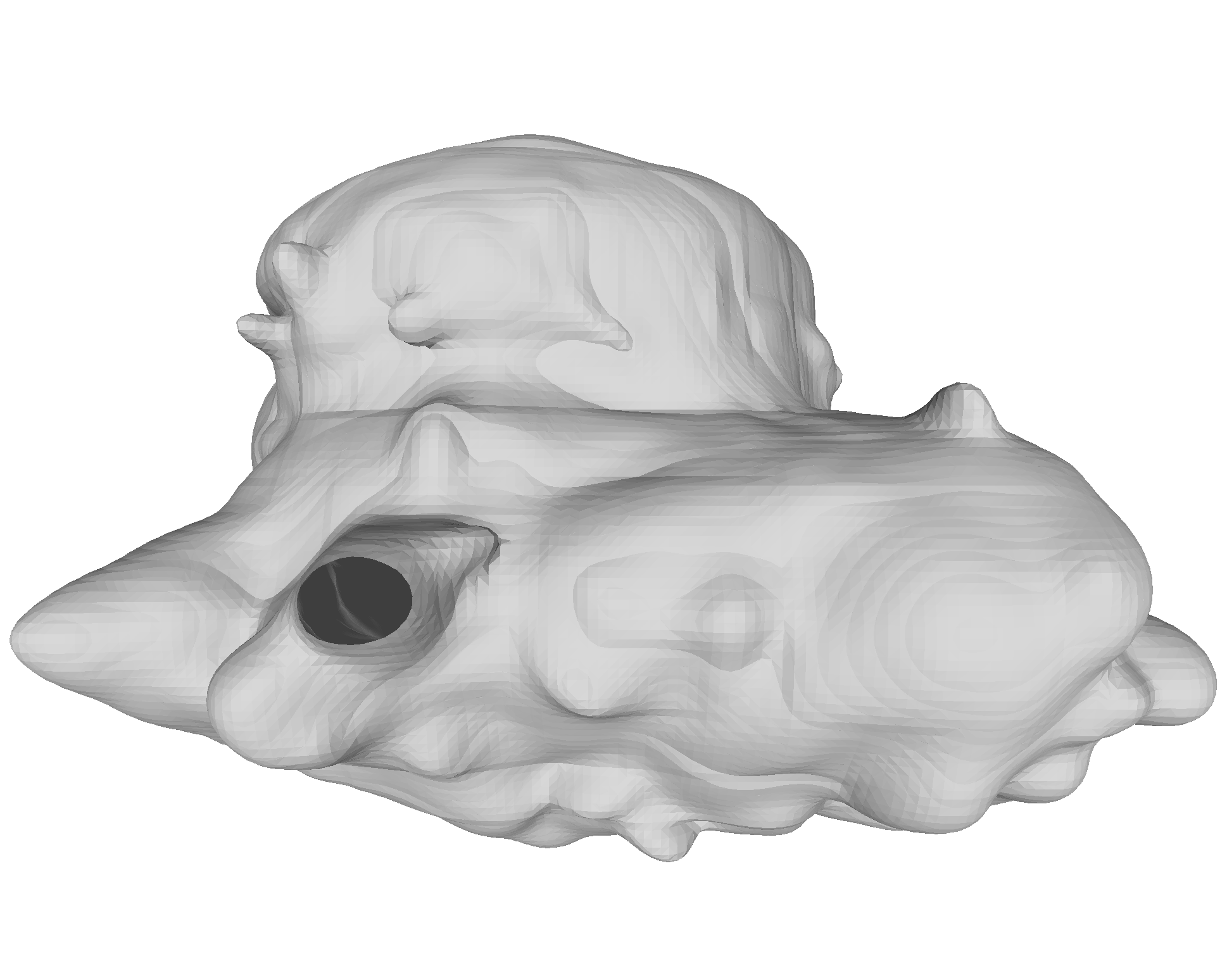} \\
\multicolumn{5}{c}{\textit{``A pineapple."}}  \\
 \includegraphics[width=0.2\textwidth]{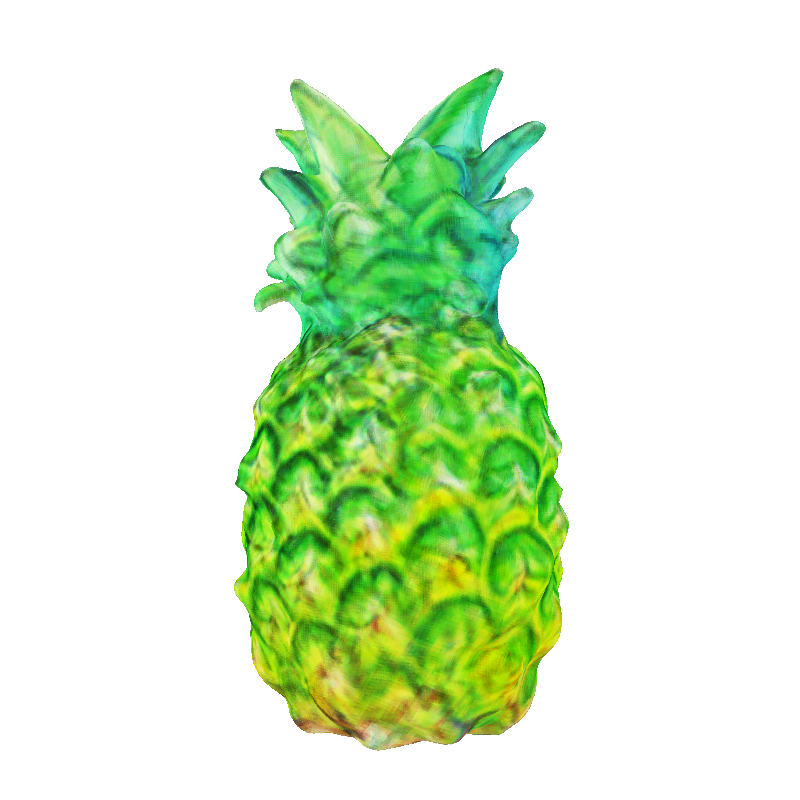} & 
 \includegraphics[width=0.2\textwidth]{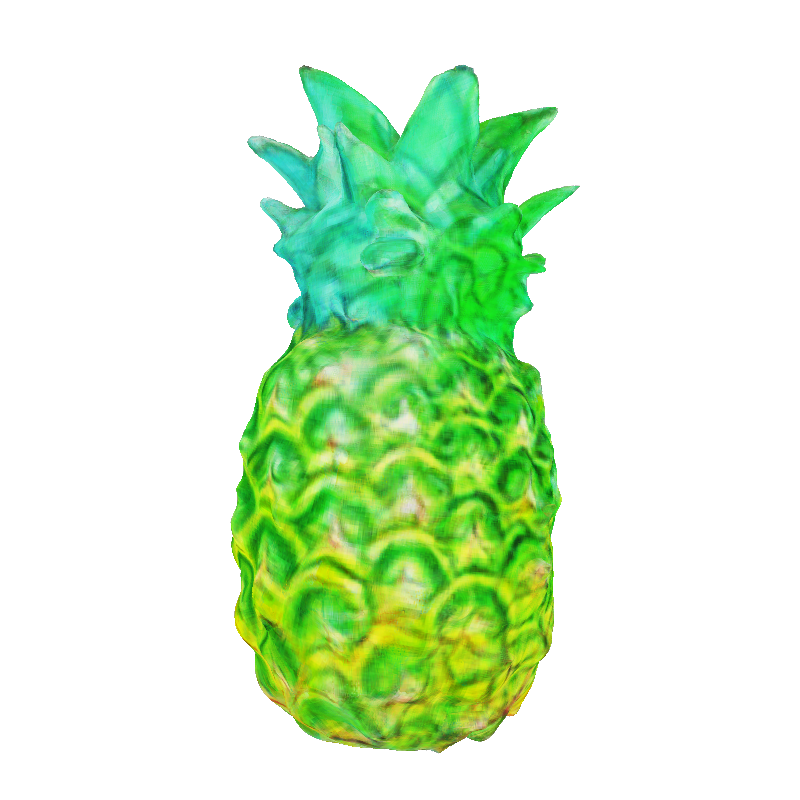} & 
 \includegraphics[width=0.2\textwidth]{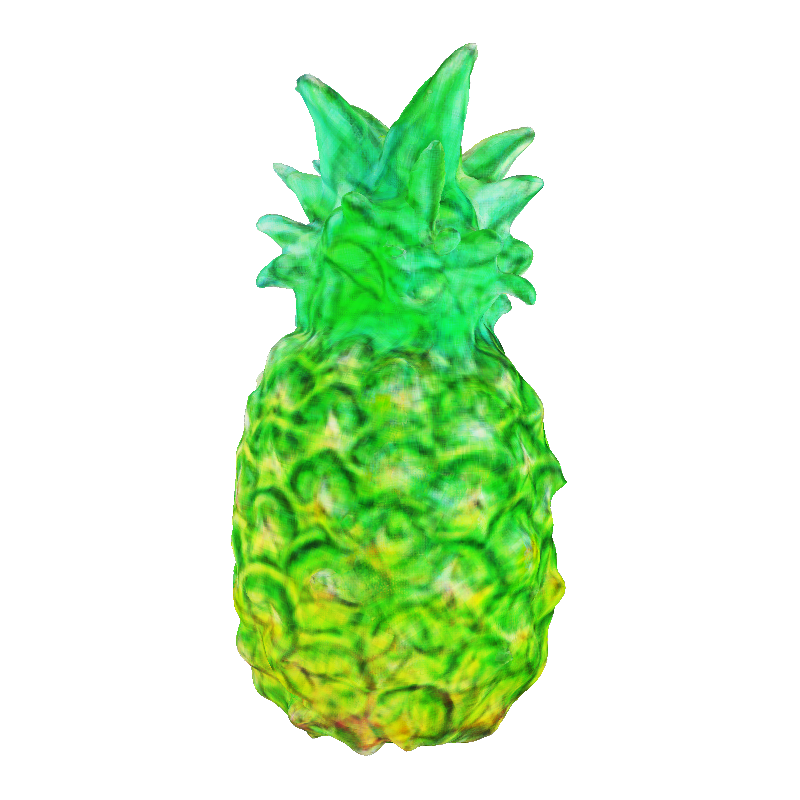} & 
 \includegraphics[width=0.2\textwidth]{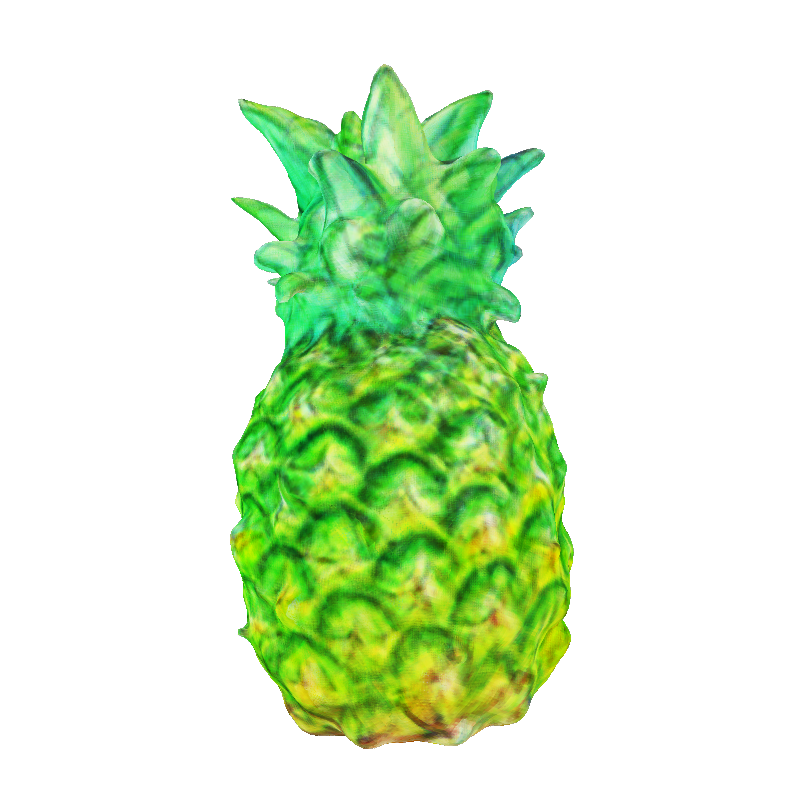} & 
 \includegraphics[width=0.26\textwidth]{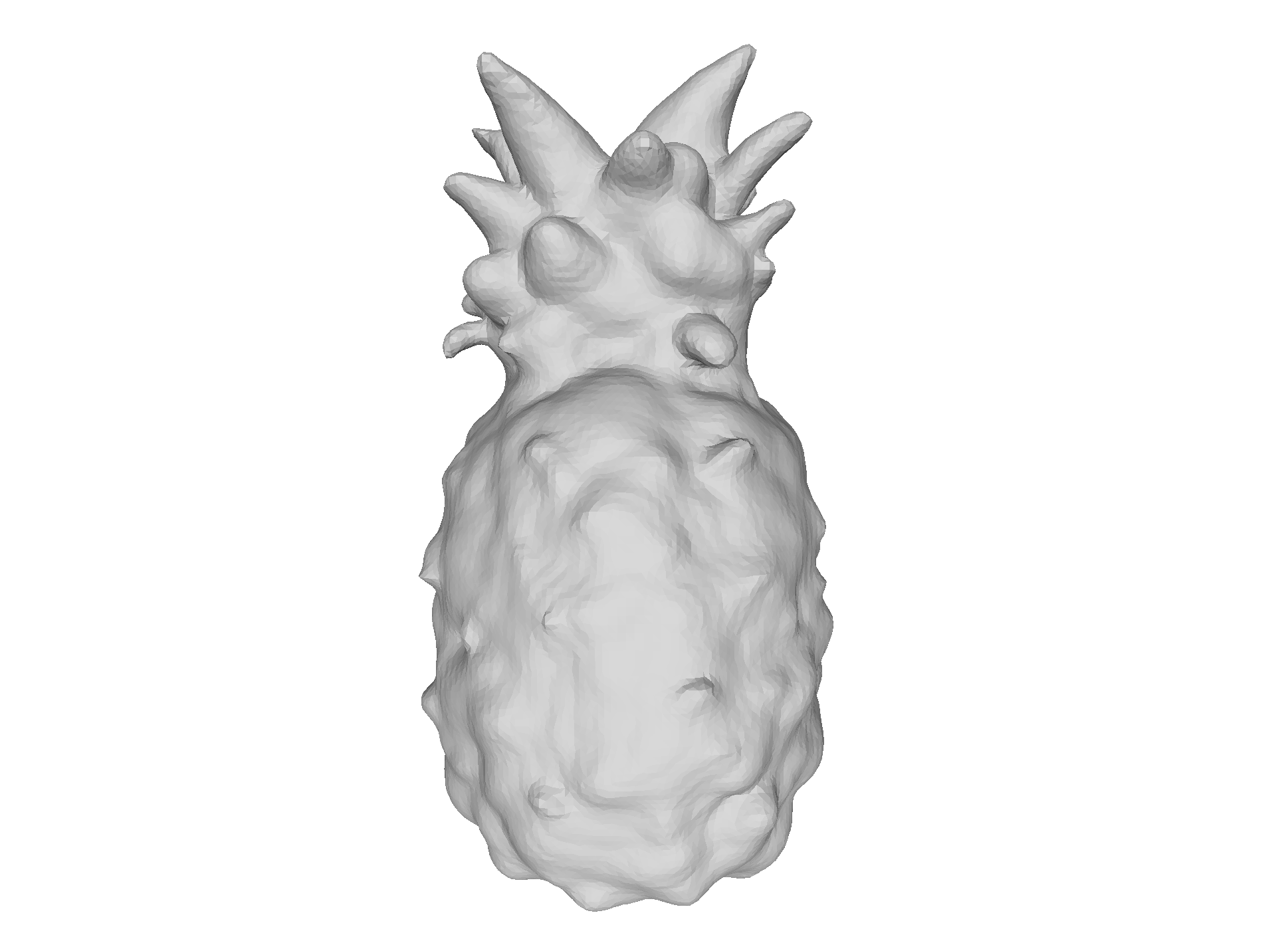} \\
\multicolumn{5}{c}{\textit{``A blue jay standing on a large basket of rainbow macarons."}}  \\
 \includegraphics[width=0.2\textwidth]{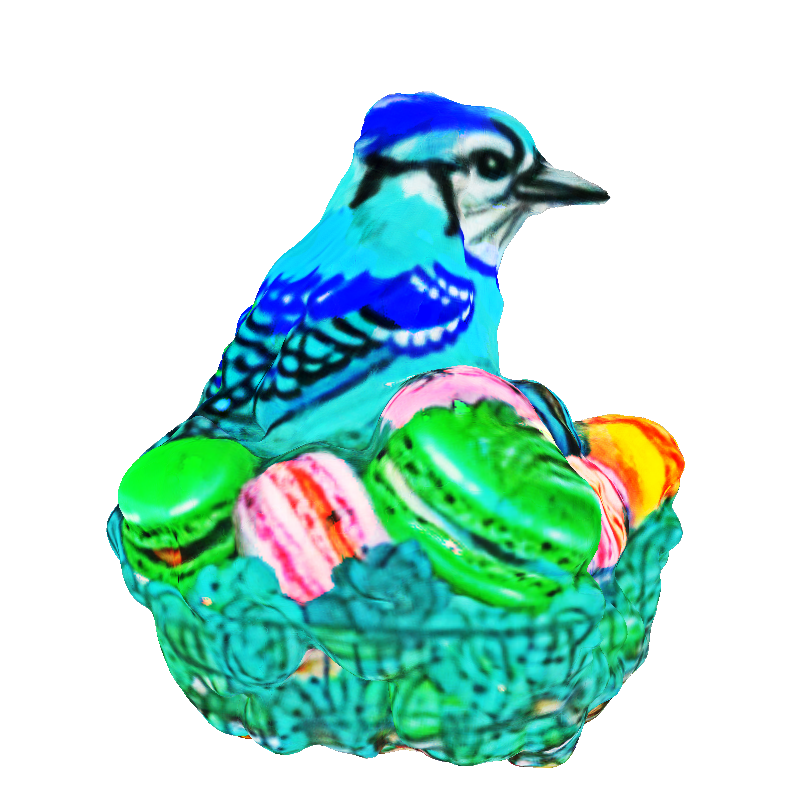} & 
 \includegraphics[width=0.2\textwidth]{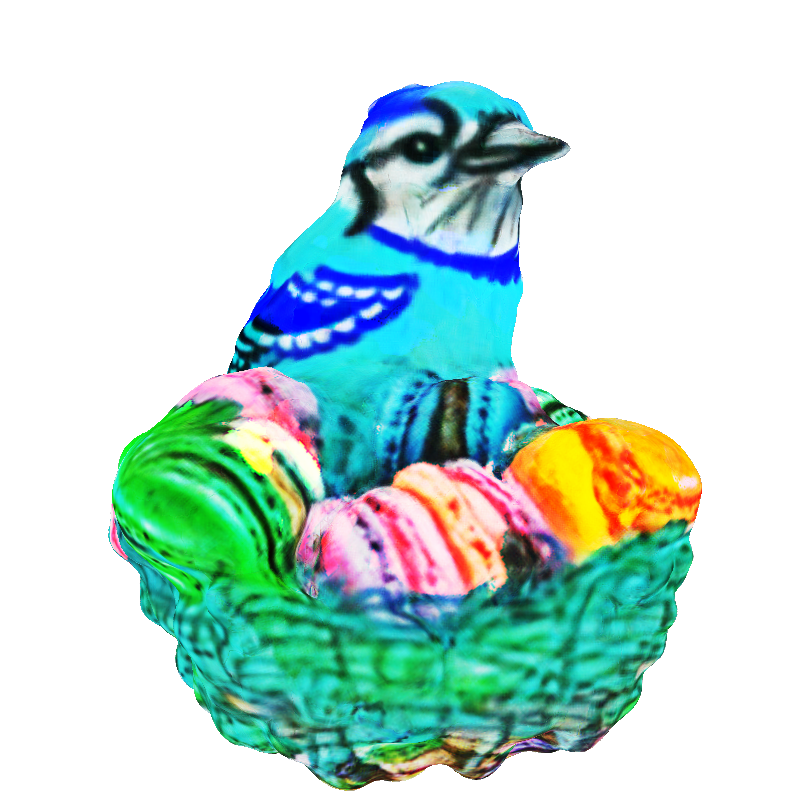} & 
 \includegraphics[width=0.2\textwidth]{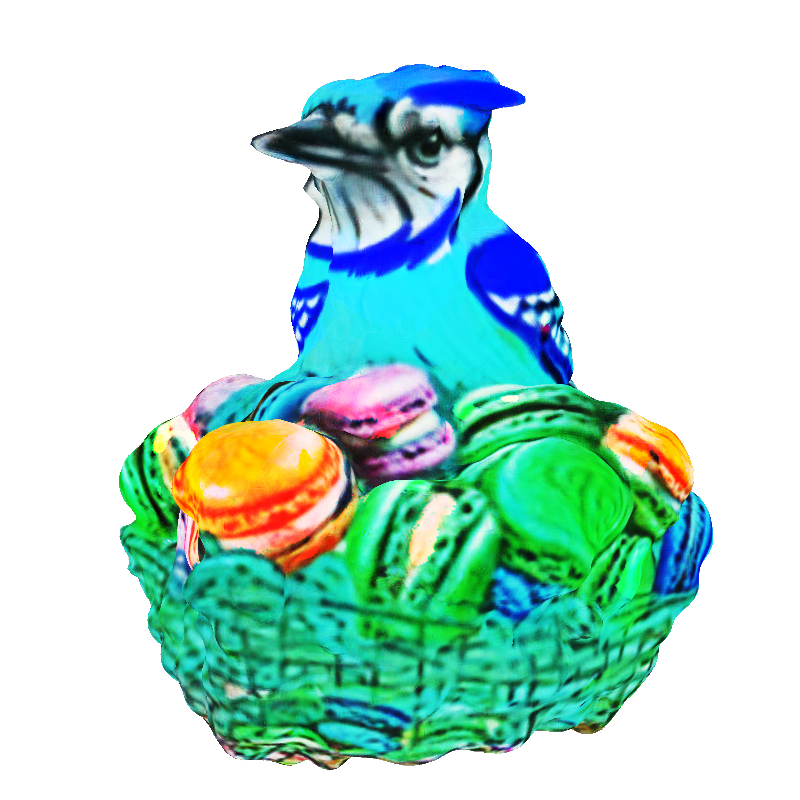} & 
 \includegraphics[width=0.2\textwidth]{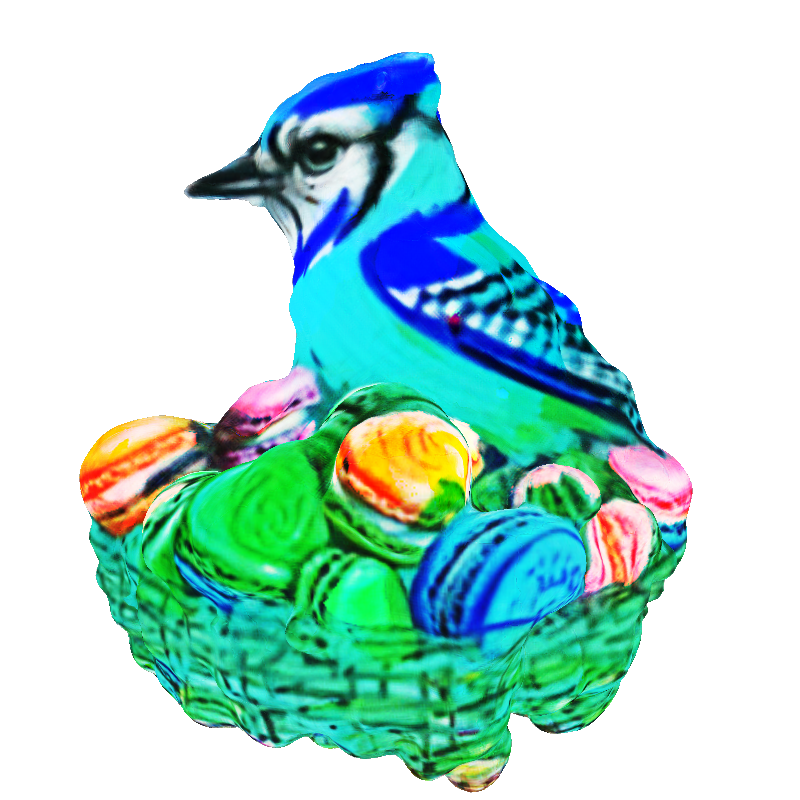} & 
 \includegraphics[width=0.26\textwidth]{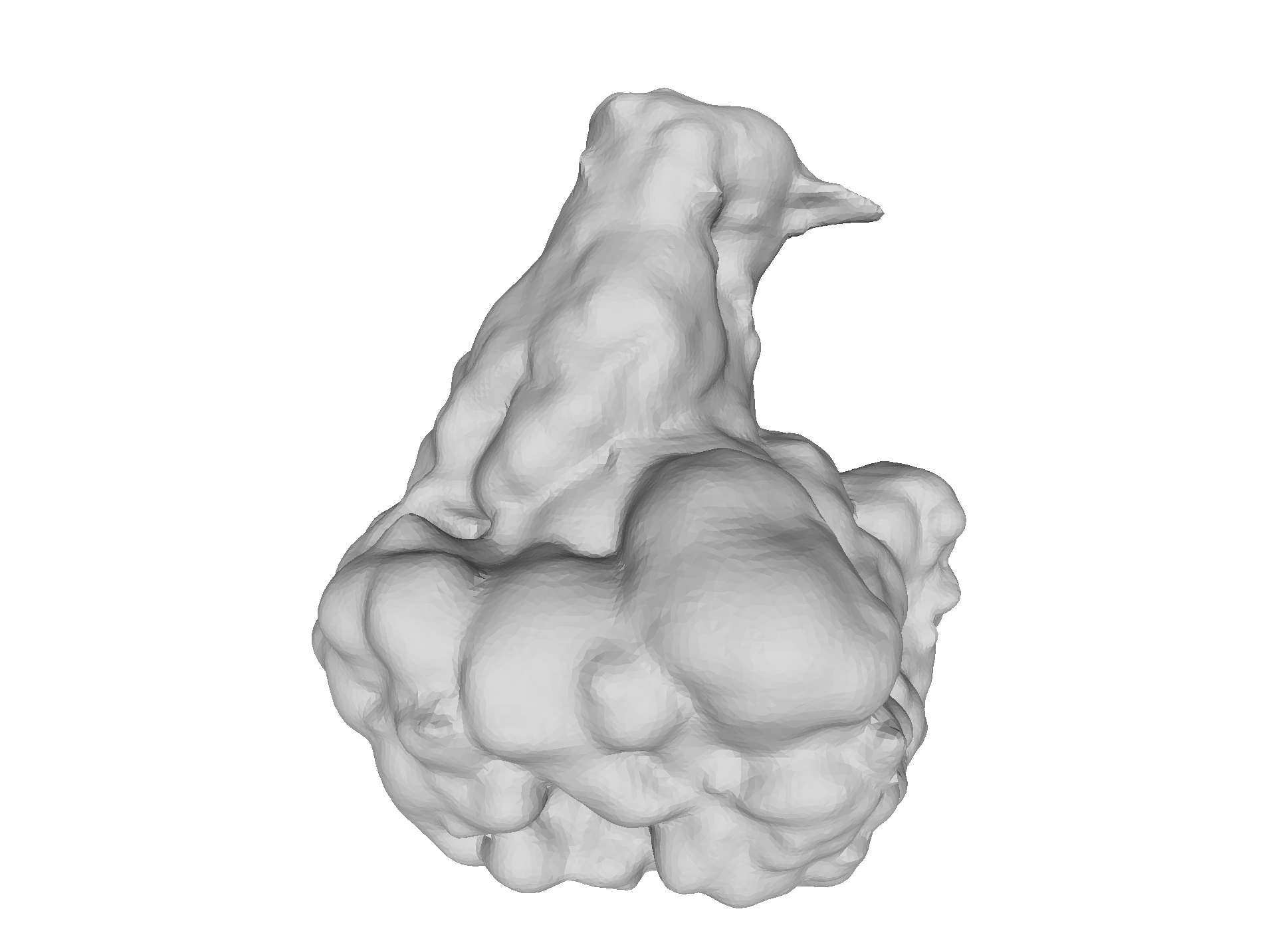} \\
\end{tabular}}
\caption{Supplementary results featuring DDPO3D. In each row, the initial four images showcase the multiview representations of the generated 3D models, aligning with the provided text prompts. The final image in each row depicts the textured mesh.}
\label{table:sup_results}
\end{table*}

\clearpage

We showcase supplementary outcomes for the text-to-3D challenge employing our DDPO3D methodology. Table \ref{table:sup_results} indicates that our approach excels in producing 3D assets with high fidelity, conditioned only on textual input.

We also explore various reward settings in our experiments to comprehend the influence of policy gradient. Further insights into these experiments will be provided in Sections \ref{sec:reward} and \ref{sec:compression}.

\subsection{Immediate vs Discounted Rewards}
\label{sec:reward}
We experiment with two settings primarily. We test our experiments with both immediate and discounted rewards. 

\begin{figure}[!htp]
    \small
    \setlength{\tabcolsep}{1pt}
    \scalebox{0.9}{
    
    \begin{tabular}{ccc}
    
         DreamGaussian \cite{tang2023dreamgaussian}& Ours  & 
         Ours 
         \\
         & 
         (immediate  & 
         (discounted  
         \\
         & 
         reward) &
    
         rewards)
    \\     
         \includegraphics[trim={2cm 2cm 2cm 0cm},clip,width=0.4\linewidth]{comparisons/dreamgaussian/cf1.jpg}&
         \includegraphics[trim={1cm 1cm 1cm 0cm},clip, width=0.4\linewidth]{comparisons/new_ddpo_multistep/cf_aes.jpg}
         & 
         \includegraphics[width=0.3\linewidth, trim={0cm -1cm 0cm -1cm},clip ]{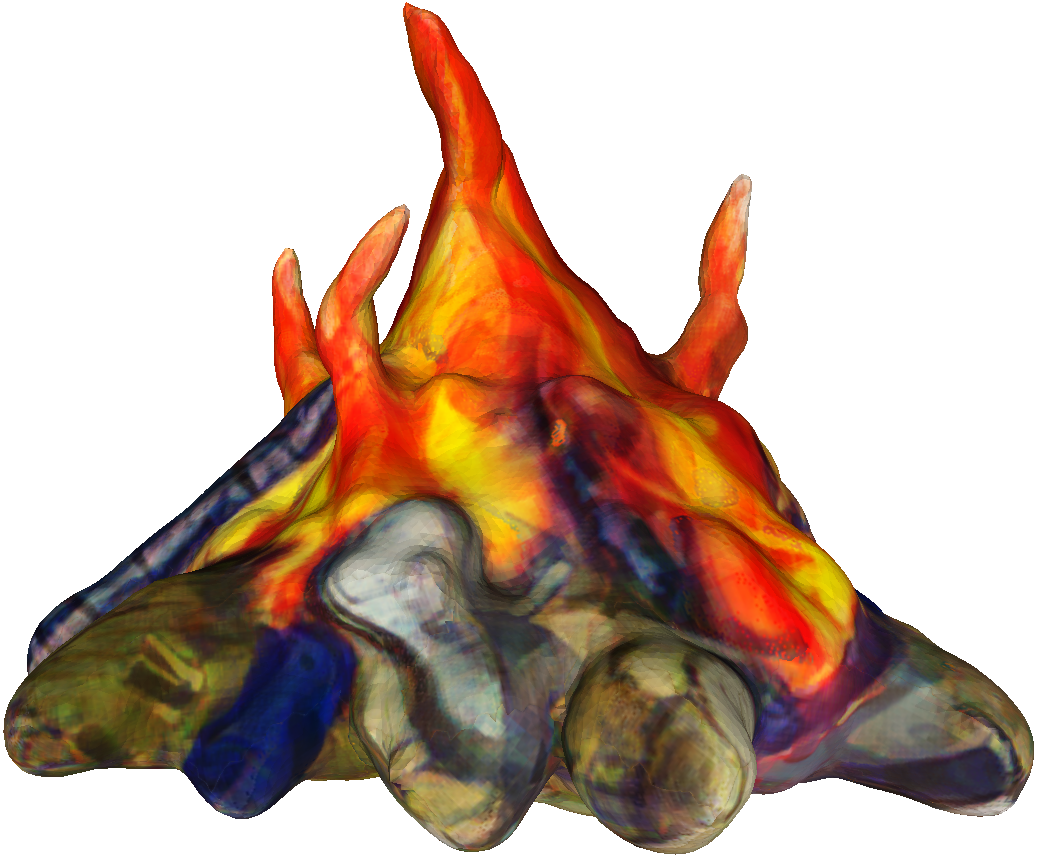}
    \\
    
    AES: 4.59 & AES: \textbf{5.16} & AES: 5.02 \\
    CLIP Score :  30.69 & CLIP Score : 31.18 & CLIP Score : \textbf{31.2}  
    \end{tabular}}
    
    \caption{We observe a similar performance with immediate and discounted rewards. However, we get greater execution speed with immediate rewards.}
    \label{fig:enter-label-sup}
\end{figure}
During the inference stage of the UNet in DDPO3D, we perform denoising only for half the time steps. In the original DDPO approach \cite{black2023training}, since the method performs updates based on the entire trajectory to finetune the Stable Diffusion model, they perform denoising on the entire diffusion trajectory. However, we do not need to do this since we do not update the Stable Diffusion model parameters but rather the Gaussian Splatting/NeRF parameters. After several denoising steps, we keep sufficient information in the images by starting the denoising process from the middle of the trajectory further away from pure Gaussian noise, thus closer to the rendered image and yielding a lower memory and temporal footprint. In the case of immediate rewards, we perform DDPO3D by calculating the log probability only on the middle time step for the immediate rewards. We use the complete trajectory after several denoising steps for discounted rewards. Whereby in discounted rewards, the last denoising time-step according to diffusion, i.e. $t=0$ receives the reward $r(x_0,c)$ similar to DDPO where $r$ is the reward function, and $x_0, c$ are the rendered image and context/prompt respectively. At the same time, the rest of the steps have a reward of $0$. In the case of immediate reward, we assign this reward to the $t=T/2$ step of denoising and ignore the rest of the denoising trajectory for calculating the discounted return. We do not notice much of a trade-off in performance between the two settings. While the discounted reward on the complete trajectory (here from $t=T/2$ to $t=0$) gives good results, there is a certain overhead due to the longer pass of the denoising pipeline but not much significant improvement over immediate rewards. In contrast, the immediate rewards balance the generated quality and time taken by the process, as discussed further in Section \ref{section:timing}.

        






        

\subsection{Compression Loss}
\label{sec:compression}
We also experiment with the compression-based reward presented in Black et al.\cite{black2023training}, where the size obtained after JPEG compression is used as a reward. We slightly modify this setting and compute this reward for the image rendered using Gaussian splatting at every time step. We examine that compression loss leads to more geometric structure reduction than the texture since we use DDPO3D in the first stage of DreamGaussian. We perceive a reduction in the geometric details of the flames in the case of "campfire" as the weight of the policy gradient term increases. Similarly, we notice a reduction in geometry details in the case of "tulip" as well. We see that the step of the tulip is no longer present, and similarly, the details of the petals also get reduced.


\section{Scheduling Impact \& Choice of HyperParams}

\begin{figure}[!htp]
    \centering
    \caption{Results with increasing weight of the policy gradient terms with \textbf{compression score (cs)}. We increase the weight of DDPO3D and perceive an increase in the reduction of geometric details. Furthermore, in the bottom row, we show the results obtained by stopping the SDS-based updates after $1K$ iterations and only resuming the policy gradient updates. The SDS terms prove to be highly assistive for maintaining structural information.}
    \scalebox{0.85}{
   
    \begin{tabular}{|c | c|}
        \midrule
         "a campfire" & "a tulip" \\
        \midrule    
        SDS post $1K$ iters & True \\

        \midrule
        
         DDPO3D weight=1
         
         & 
         DDPO3D weight=1 \\
       
         \includegraphics[width=0.5\linewidth]{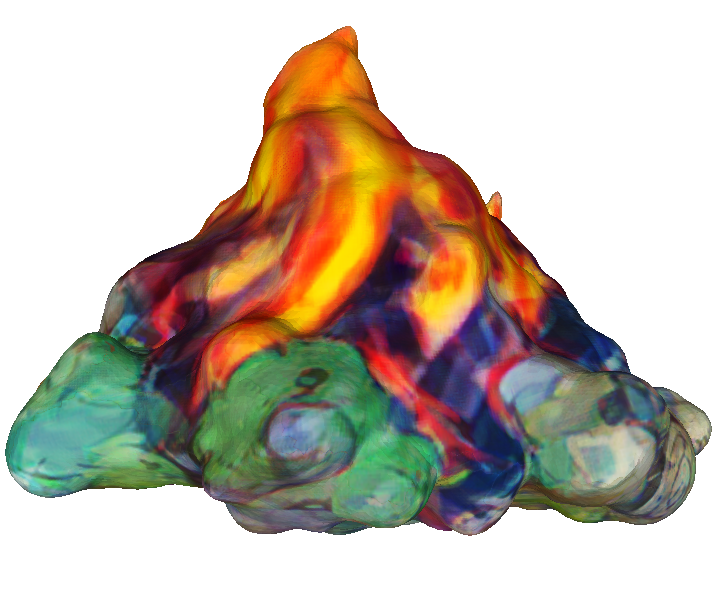} &
         
         \includegraphics[width=0.35\linewidth]{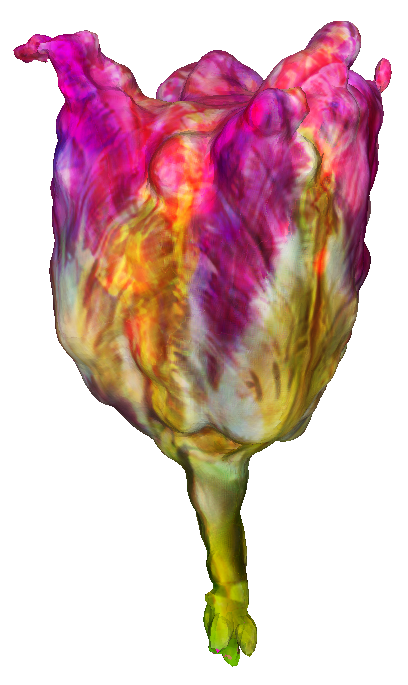}\\

         DDPO3D weight=2.5
         
         & 
         DDPO3D weight=2.5 \\
         
         \includegraphics[width=0.5\linewidth]{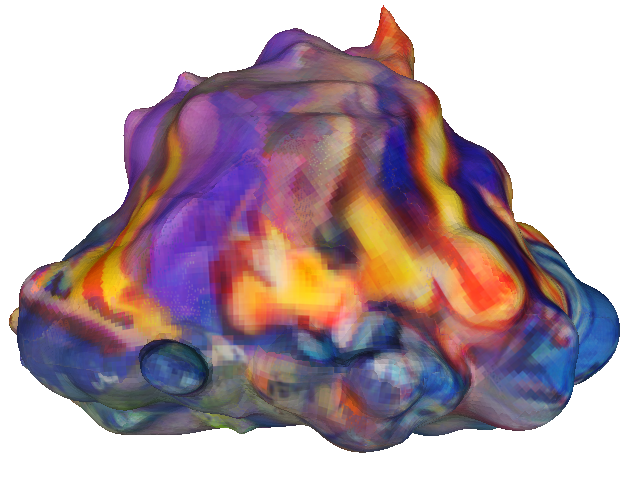} 
         
         & 
         \includegraphics[width=0.35\linewidth]{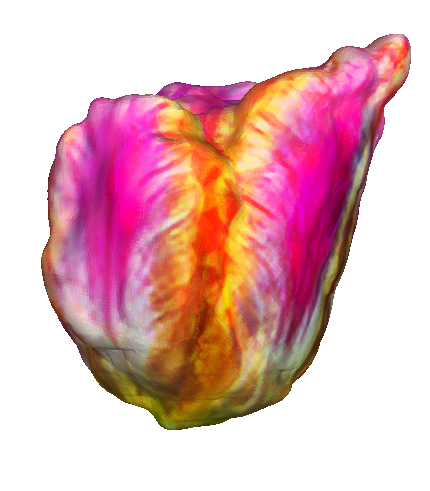}
          \\

         \midrule
         SDS post $1K$ iters & False \\
         \midrule
                 DDPO3D weight=2.5 & DDPO3D weight=2.5 \\ 
         \includegraphics[width=0.5\linewidth]{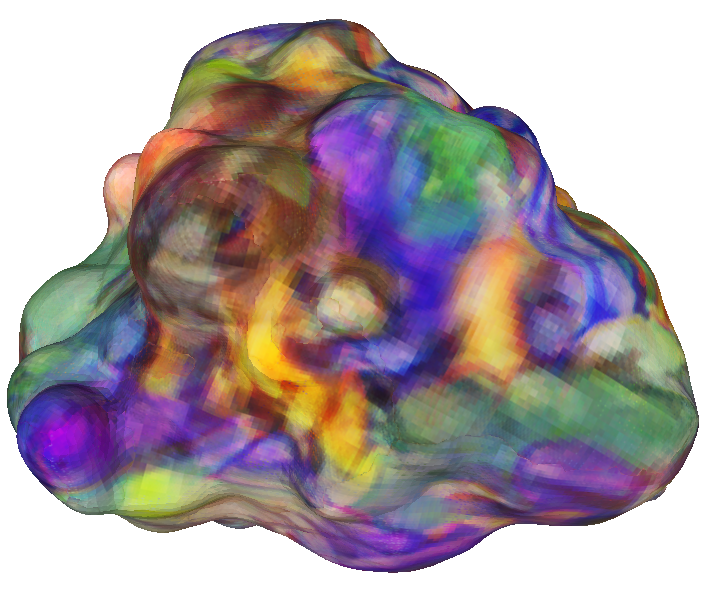} &  
         \includegraphics[width=0.35\linewidth]{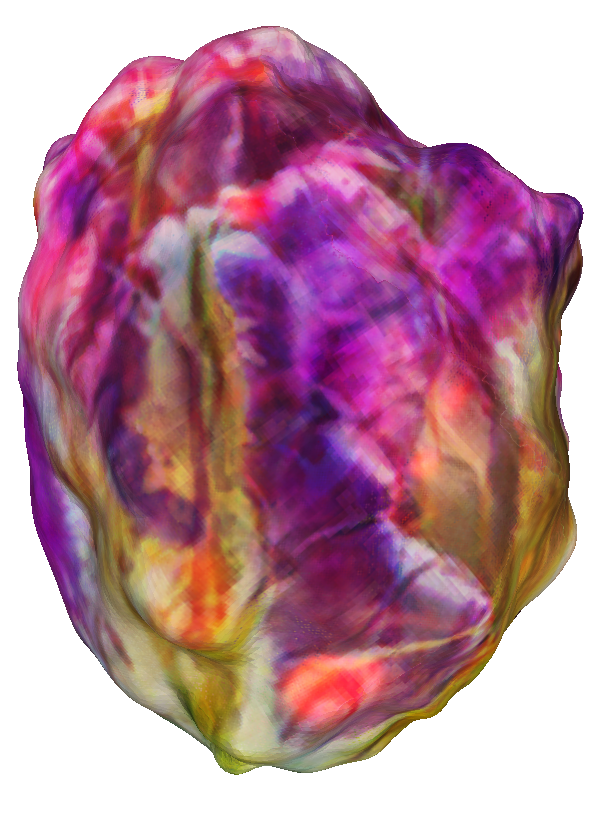} \\
         \midrule
         
    \end{tabular}}
    
    \label{fig:compress}
\end{figure}

We further test the impact of removing SDS loss during training and only use the Policy Gradient. We train the DreamGaussian for $1.5K$ steps, whereby we stop training using the SDS loss after $1K$ steps and only continue training with the policy gradient term. We witness additional color gradation, as we noticed in the case of failure and lack of structural information if the weight of the policy gradient is high. As can be seen from the last row in Figure \ref{fig:compress}, the removal of SDS from the latter iterations has a deteriorating impact on both geometry and texture as compared to when SDS is performed as in \ref{fig:compress}.

\section{Timing Information}
\label{section:timing}
While using the multistep discounted reward calculation in DDPO3D, we observe an increment of an additional $1.76$sec/iteration, while with the instantaneous reward, there is an additional time of $0.16$sec per iteration. To see the impact of DDPO3D, we generally run experiments for  1500 iterations.

%
%

\newpage